\documentclass[english]{article}
\usepackage[preprint,nonatbib]{nips_2018_wider_nonotice}

\usepackage[T1]{fontenc}
\usepackage[latin9]{inputenc}
\usepackage{color,colortbl}
\usepackage{babel}
\usepackage{verbatim}
\usepackage{url}
\usepackage{amsmath}
\usepackage{amssymb}
\usepackage{graphicx}
\usepackage{setspace}
\usepackage{hyperref}
\usepackage{cancel}
\hypersetup{unicode=true,pdfusetitle,breaklinks=false}
\usepackage[hyperpageref]{backref}
\usepackage{appendix}

\graphicspath{{figures/}}

\usepackage[labelfont=bf]{caption}
\usepackage{makecell}
\usepackage{multirow}

\makeatletter
\newcommand{\be}{\begin{eqnarray}}
\newcommand{\ee}{\end{eqnarray}}

\allowdisplaybreaks \numberwithin{equation}{section}
\makeatother

\def\<{\langle}

\newcommand\nn{\nonumber}

\usepackage[disable]{todonotes} % for    todos
\usepackage{tabularx} % used to handle word wrapping
\usepackage{booktabs} % for tables

\usepackage{caption} 
\title{Scaling Laws for Autoregressive Generative Modeling}

\author{
Tom Henighan\thanks{equal contribution} \And
Jared Kaplan\footnotemark[1] \thanks{Johns Hopkins University and OpenAI 
\newline \newline
Correspondence to: \href{mailto:henighan@openai.com,jared@openai.com,sam@openai.com}{\texttt{[henighan,jared,sam]@openai.com}}
\newline\newline Author contributions \hyperref[sec:contributions]{listed at end of paper}.
} \And
Mor Katz\footnotemark[1] \AND 

Mark Chen \And
Christopher Hesse \And
Jacob Jackson \And
Heewoo Jun \AND

Tom B. Brown \And
Prafulla Dhariwal \And
Scott Gray \And
Chris Hallacy \And
Benjamin Mann \AND
Alec Radford \And
Aditya Ramesh \And
Nick Ryder \And
Daniel M. Ziegler \AND

John Schulman \And
Dario Amodei \And
Sam McCandlish \AND

{\large OpenAI}
}

\begin{document}
\maketitle

\begin{abstract}
We identify  empirical scaling laws for the cross-entropy loss in four  domains:  generative image modeling, video modeling, multimodal image$\leftrightarrow$text models, and mathematical problem solving.  In all cases autoregressive Transformers smoothly improve in performance as model size and compute budgets increase, following a power-law plus constant scaling law.  The optimal model size also depends on the compute budget through a power-law, with exponents that are nearly universal across  all data domains.

The cross-entropy loss has an information theoretic interpretation as $S($True$) + D_{\mathrm{KL}}($True$||$Model$)$, and  the empirical scaling laws suggest a prediction for both the true data distribution's entropy and the KL divergence between the true and model distributions.  With this interpretation, billion-parameter Transformers are nearly perfect models of the YFCC100M image distribution downsampled to an $8\times 8$ resolution, and we can forecast the model size needed to achieve any  given reducible loss (ie $D_{\mathrm{KL}}$) in nats/image for other resolutions.  

We find a number of additional scaling laws in specific domains:  (a) we identify a scaling relation for the mutual information between captions and images in multimodal models, and show how to answer the question ``Is a picture worth a thousand words?'';  (b) in the case of mathematical problem solving, we identify scaling laws for model performance when extrapolating beyond the training distribution; (c) we  finetune generative image models for ImageNet classification and find smooth scaling of the classification loss and error rate, even as the generative loss levels off.   Taken together, these results strengthen the case that scaling laws have important implications for neural network performance, including on downstream tasks.

\end{abstract}

\newpage\tableofcontents{}

\newpage

\section{Introduction}

Large scale models, datasets, and compute budgets have driven rapid progress in  machine learning.  Recent work \cite{1712.00409, rosenfeld2019constructive, li2020train, roller2020recipes, kaplan2020scaling, sharma2020neural, brown2020language} suggests that the benefits of scale are also highly predictable.
When the cross-entropy loss $L$ of a language model is bottlenecked by either the compute budget $C$, dataset size $D$, or model size $N$, 
the loss scales with each of these quantities as a simple power-law.  Sample efficiency also improves  with model size.  

These results raise a number of questions.  Do they apply to all data modalities?  How do improvements on the loss translate to improvements in representation quality and performance on downstream tasks?  Is there any way to determine when and why the performance of a model might be maxed out, so that  further scaling will be met with diminishing returns?  What explains the precision and universality of these trends, and what else can we learn from them?

\begin{figure}
\noindent \centering{} 
\includegraphics[width=\textwidth]{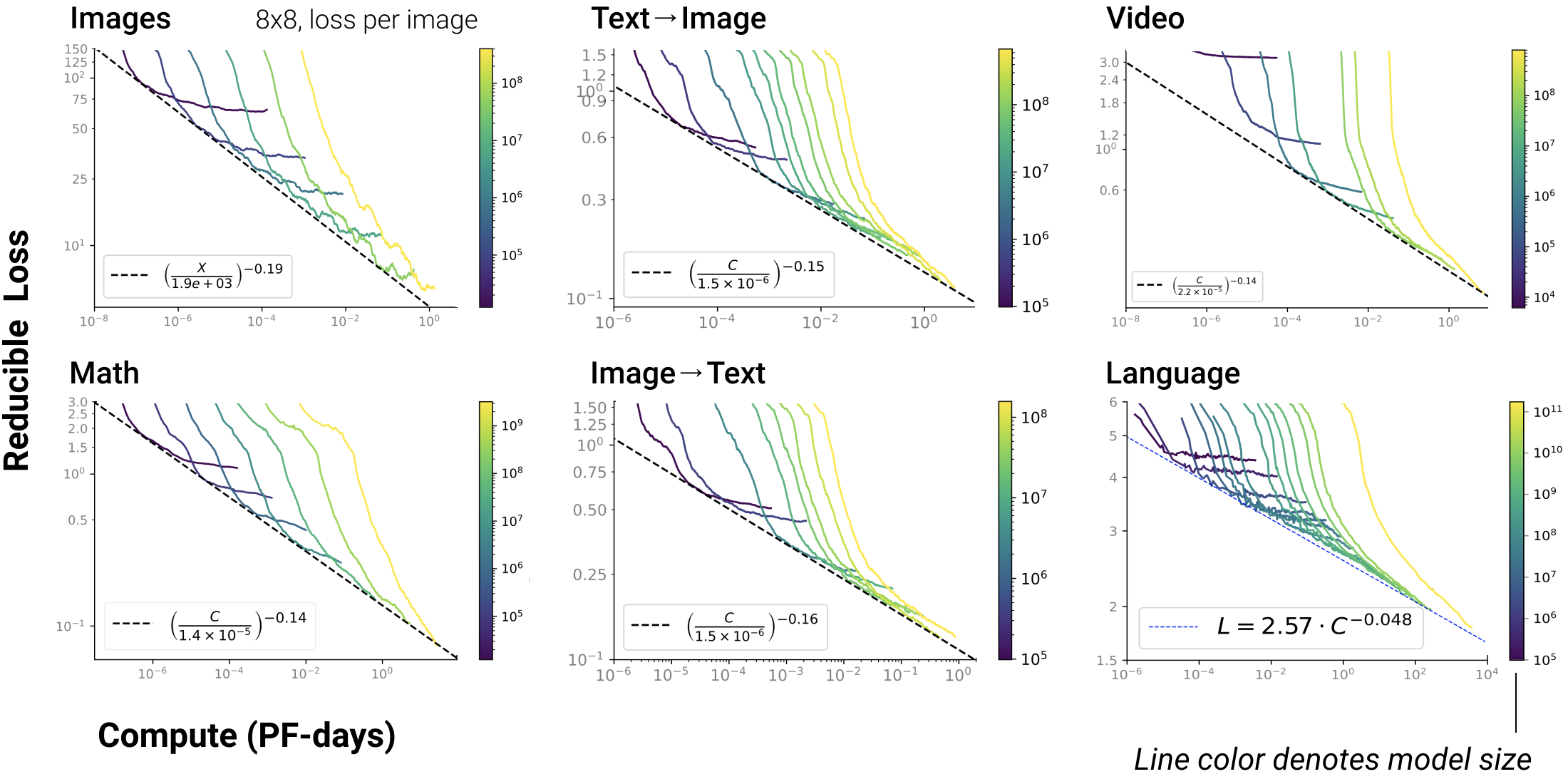}

\caption[Compute Scaling]{\textbf{Smooth scaling of \emph{reducible} loss across domains---} We show power-law scaling laws for the \emph{reducible} loss $L - L_{\infty}$ as a function of compute, where the irreducible loss $L_\infty$ is a fitted domain-dependent constant.   Under plausible assumptions concerning the infinite data and compute limits, the irreducible loss estimates the entropy of the underlying data distribution, while the reducible loss approximates the KL divergence between the data and model distributions.  In the case of language we use results from \cite{brown2020language}, and only show the full loss $L$. 
\label{fig:ComputeScaling}}
\end{figure}

We will demonstrate that scaling laws apply  to generative modeling  across a wide variety of data modalities, including generative language \cite{kaplan2020scaling, brown2020language}, image \cite{DBLP:journals/corr/ThomeeSFENPBL15, icml2020_6022}, and video modeling \cite{weissenborn2019scaling}, multimodal modeling \cite{tsai2019multimodal} of text-image correlations, and even mathematical problem solving \cite{DBLP:journals/corr/abs-1904-01557}, a task requiring a degree of reasoning ability.  Moreover, we demonstrate that a single architecture -- the Transformer \cite{OriginalTransformer, liu2018generating}, with an autoregressive cross-entropy loss -- scales smoothly in all of these domains, with only minimal changes to hyperparameters such as width, depth, or learning rate.  We also observe that larger models consistently learn faster, achieving any given value of the loss in fewer steps.  

By studying many different model sizes $N$, compute budgets $C$, or dataset sizes $D$, we demonstrate that the  scaling relation for the loss 
\be
\label{eq:PowerLawPlusConstant}
L(x) = L_\infty + \left( \frac{x_0}{x}  \right)^{\alpha_x} 
\ee
applies to each data modality, where $\alpha_x$ is a modality-dependent scaling exponent, and  we primarily study $x = N, C,$ and occasionally $D$.  We will refer to $L_\infty$ as the irreducible loss and the power-law scaling term as the reducible loss. These scaling relations often hold to high precision, even when the reducible loss is much smaller than the irreducible loss; we display trends in $L(C)$ for the reducible loss in figure \ref{fig:ComputeScaling}.  Note that small deviations are visually amplified on the log-plot, but nevertheless the trends fit remarkably well.  

These observations suggest the information theoretic interpretation 
\begin{align}
\label{eq:InfoTheoryInterpretationofLoss}
L_{\infty} & \approx S({\rm True}) & \text{``Irreducible Loss''}\nn\\
\left(\frac{x_{0}}{x}\right)^{\alpha_{x}} & \approx D_{\mathrm{KL}}({\rm True}||{\rm Model}) & \text{``Reducible Loss''}
\end{align}
In other words, the irreducible loss estimates the entropy of the true data distribution, while the reducible loss is an estimate of the KL divergence between the true and model distributions.  One might have guessed that as the $L(x)$ curve bends and the loss approaches $L_\infty$, returns to increasing $N, C, D$ are diminishing.  But the identification of the reducible loss with $D_{\mathrm{KL}}$ suggests this is not necessarily the case, and further increases in scale may still provide important additional semantic information.  
To  justify equation (\ref{eq:InfoTheoryInterpretationofLoss}), we must assume that in the limit $D \to \infty$ followed\footnote{We specify this order of limits to make it clear that regularization will not be required.} by $N, C \to \infty$, an infinitely large transformer could  model the data distribution exactly.

\begin{figure}
\noindent \centering{} 
\includegraphics[width=0.7\textwidth]{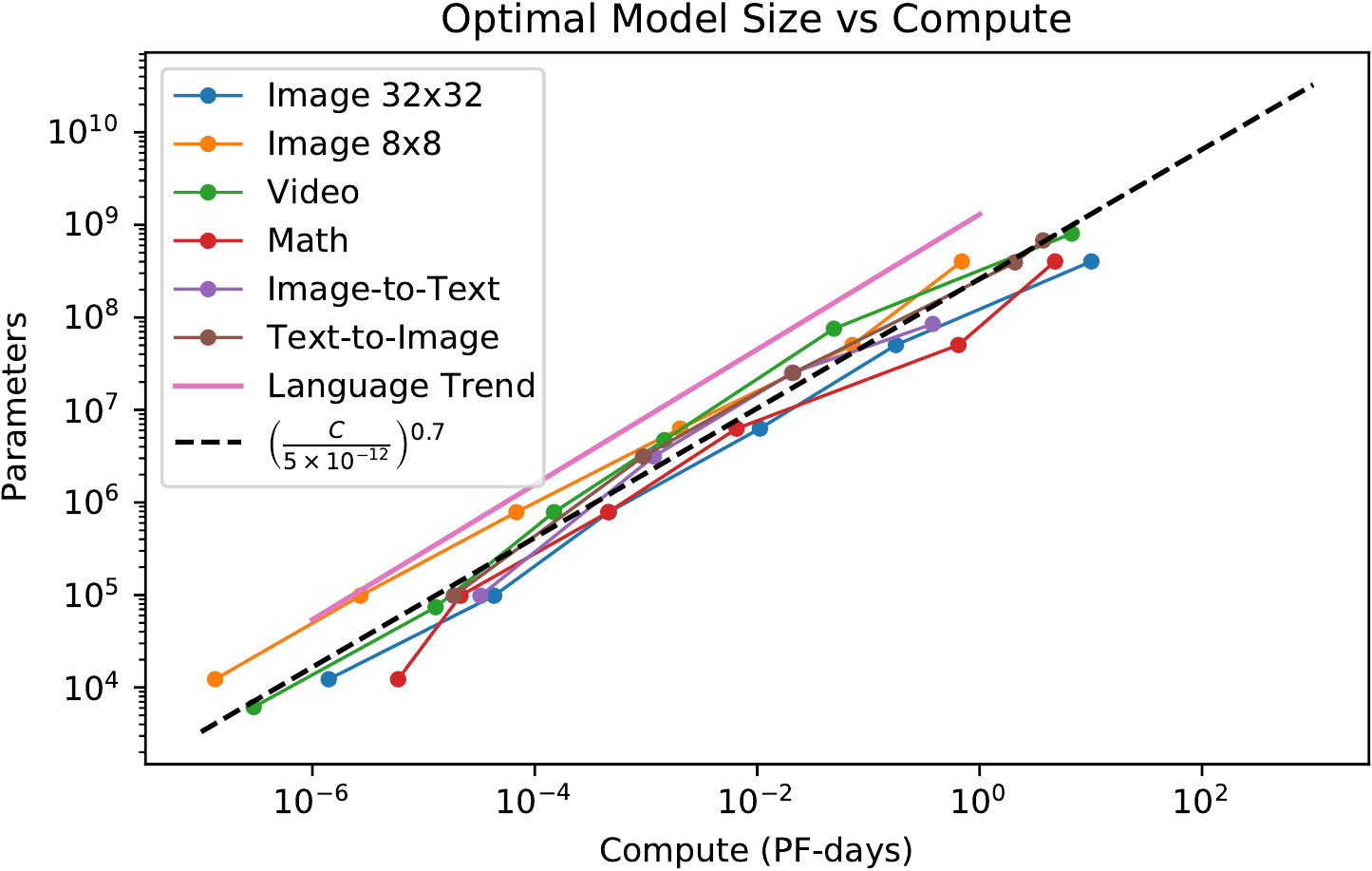}
\caption[Optimal Model Size ]{\textbf{Optimal model size is consistent across domains---} We display the optimal model size $N_{\rm opt}$ as a function of the training compute budget $C$.  Not only does $N_{\rm opt}(C)$ behave as a power-law, but the behavior is remarkably similar for all data modalities.
\label{fig:OptimalModelSizeAllDomains}}
\end{figure}

The scaling relations provide insights into the complexity of the data and clarify the value of increasing $N, D,$ and $C$.  
By evaluating the reducible loss for a full image or video, we are actually estimating  the number of bits of information that `remain to be understood' by a given model.  Equivalently, the reducible loss approximates the degree to which the data could be further compressed.  We find that billion-parameter models can extract all but a few nats/image concerning YFCC100M images \cite{DBLP:journals/corr/ThomeeSFENPBL15} downsampled to an 8x8 resolution, so they may be nearly perfect models of this data distribution.  For larger, more practically relevant images we would need far larger models to achieve this feat, but the scaling laws make it possible to forecast this precisely.  These trends are closely tied to the scaling exponents $\alpha_x$: smaller exponents imply slower improvement with increasing scale, meaning that the data can only be compressed further with much larger models.   

The scaling of loss with compute makes it possible to estimate the optimal model size for a given compute budget.  We find that just as in \cite{kaplan2020scaling} this relation is very nearly a pure power-law  $N_{\rm opt}(C) \propto C^\beta$.  Surprisingly, the exponent $\beta \sim 0.7$ is very similar for all domains, as shown in figure \ref{fig:OptimalModelSizeAllDomains}.  This has important implications for the scaling of dataset size with model size for compute-optimal training, suggesting that $D \propto N^{0.4}$ if we only train on each data element once.  Even allowing for significant errors or deviations, this strongly suggests sub-linear scaling of dataset size with model size.

We can learn more if we focus on questions specific to each data modality.  
Generative image models can be finetuned for classification.  We will show that ImageNet \cite{DBLP:journals/corr/ChrabaszczLH17} classification performance improves smoothly with pre-trained model size,  following another power law.  This trend continues even into the large-model regime where the generative loss trend ``bends'' and becomes dominated by the irreducible component. This strongly suggests that there are  benefits to squeezing as much performance as possible out of large generative image models, as significant semantic information may lie in the `last few bits'.   
The smooth trends for finetuned performance on image classification suggest a more general lesson:  that the scaling laws for unsupervised learning imply that downstream performance also improves with model size and compute.

Information theory  provides a useful lens for examining model performance in other contexts.  A striking case is provided by multimodal models, such as those that model the joint distribution between text captions and images.  Typically the entropy of the caption is much smaller than that of the image, so  the ratio between the (empirical) mutual information\footnote{By the empirical mutual information we are referring to $\mathbb{E}_{x,y \sim q }  \left[ \log \frac{p(x,y)}{p(x) p(y)} \right]$ where $p$ is the model distribution and $q$ is the true distribution of the data.  This must be smaller than the cross-entropy loss of the model on both $X$ and $Y$. } and the model's loss on the text, which we refer to as the
\be
\label{eq:InfogainDefinition}
{\rm Infogain} \equiv \frac{I({\rm text}, {\rm image})}{ L({\rm text})} 
\ee
provides an interesting metric for model performance.  The mutual information shared between distributions must be smaller than the amount of information in either distribution, so this ratio must be less than $1$.  Furthermore,  it  appears that the Infogain increases smoothly with model size, so that the bound ${\rm Infogain} < 1$ can suggest a target model size for maximum performance.  Typically this is far beyond current capabilities.

These smooth scaling results on a wide variety of datasets also demonstrate the remarkable versatility of the Transformer architecture.

\begin{figure}
\noindent \centering{} 
\includegraphics[width=0.31\textwidth]{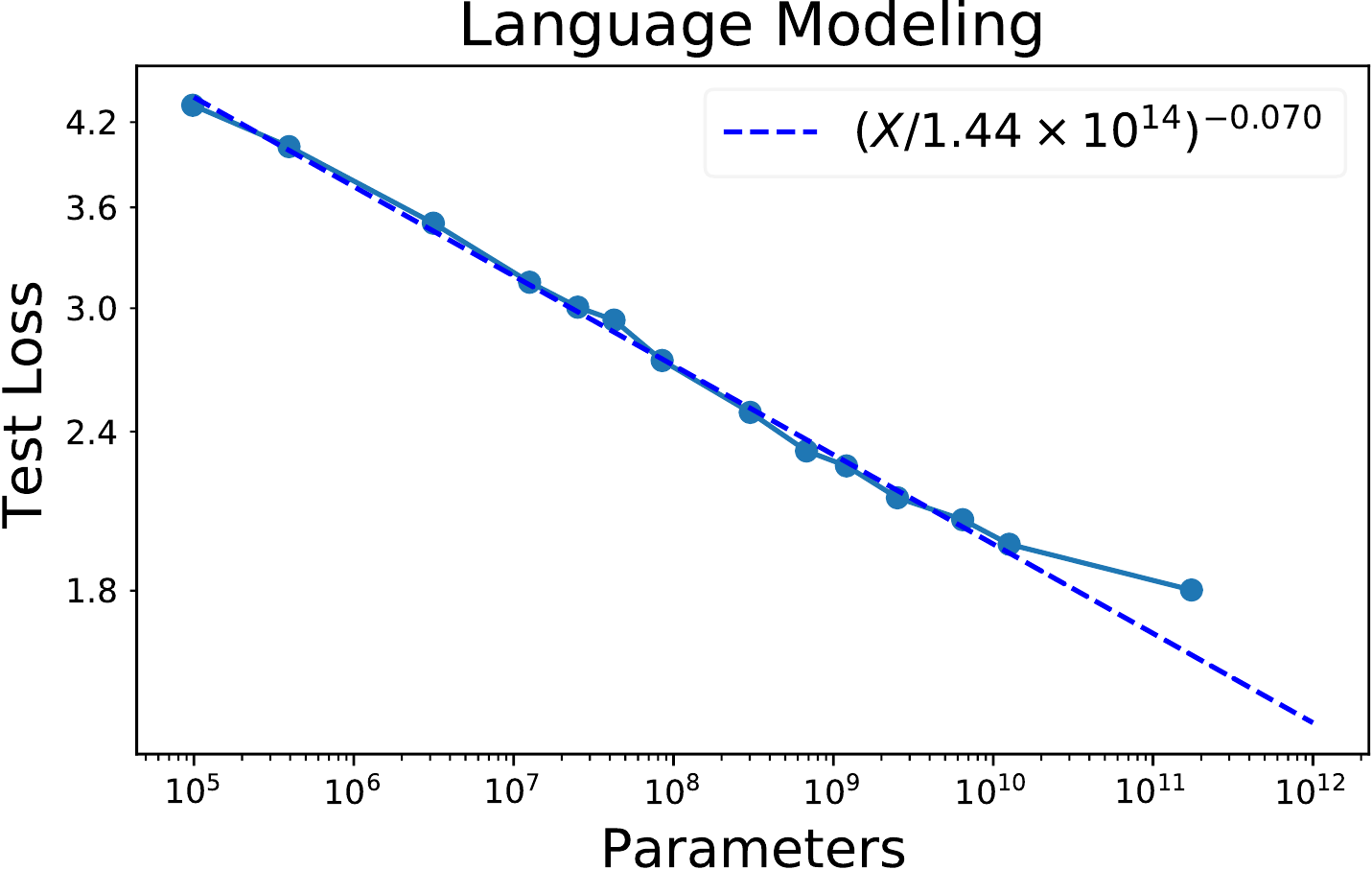}\hfill
\includegraphics[width=0.31\textwidth]{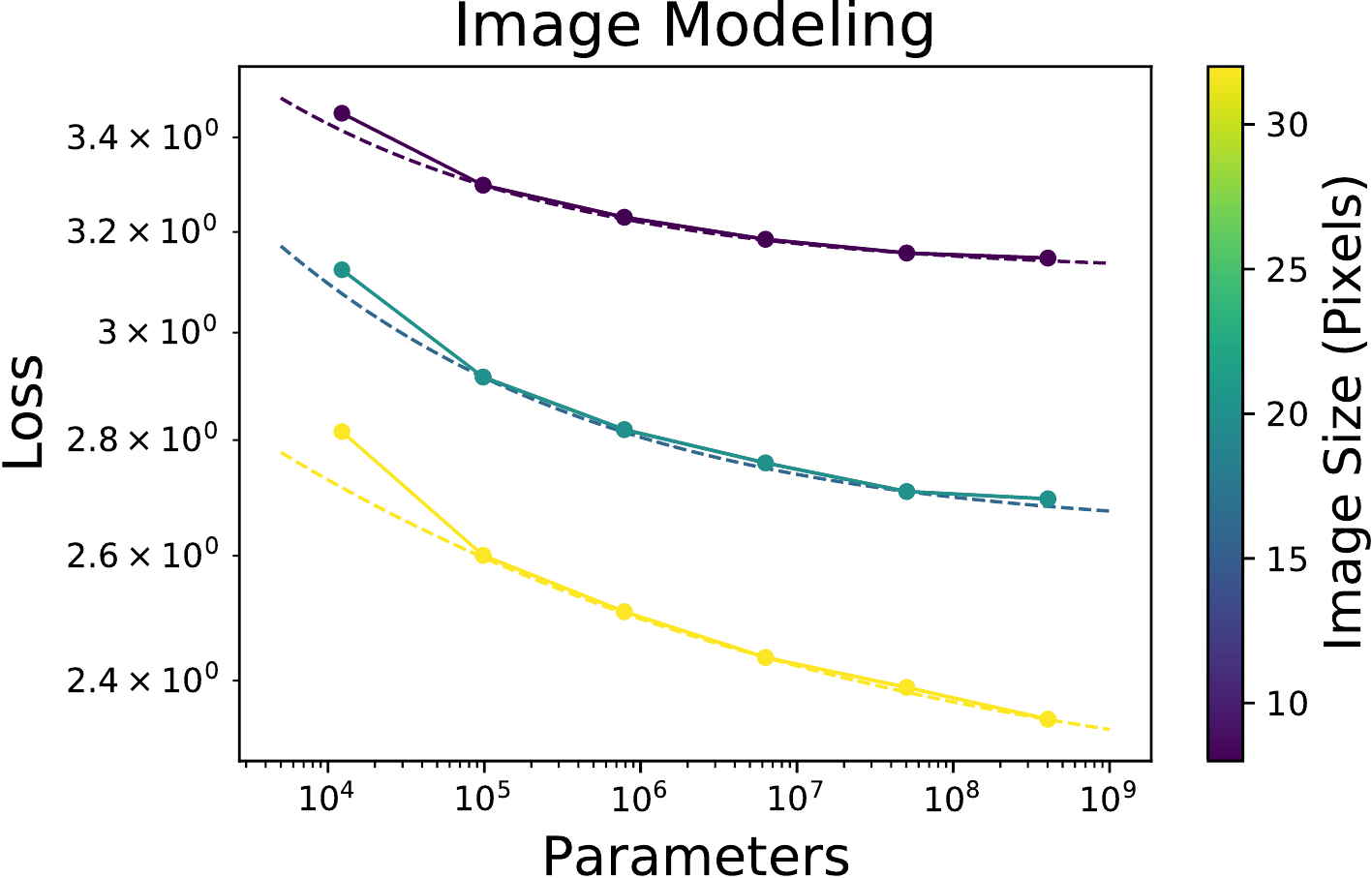}\hfill
\includegraphics[width=0.31\textwidth]{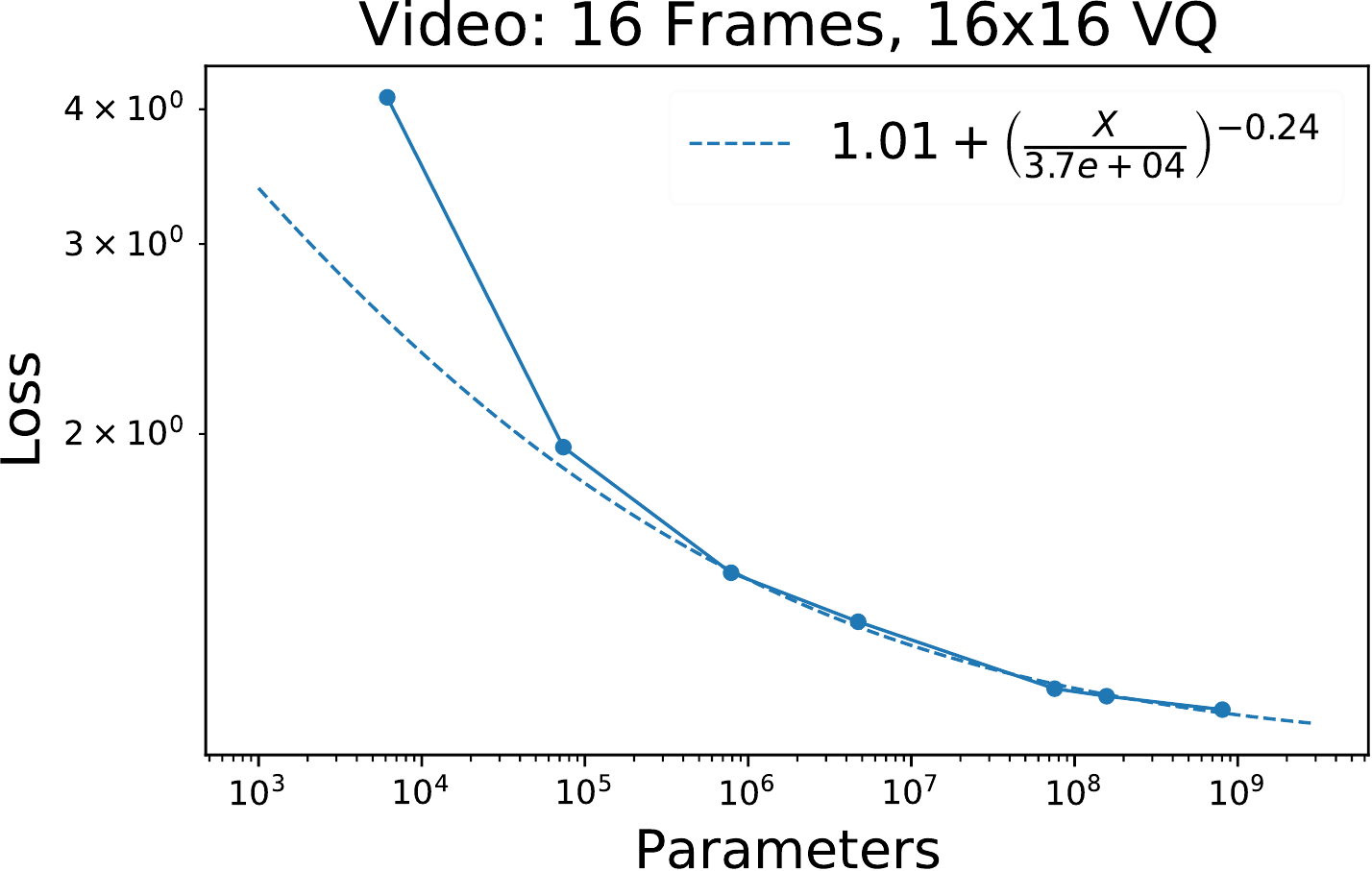}
\\
\vspace{1em}\includegraphics[width=0.31\textwidth]{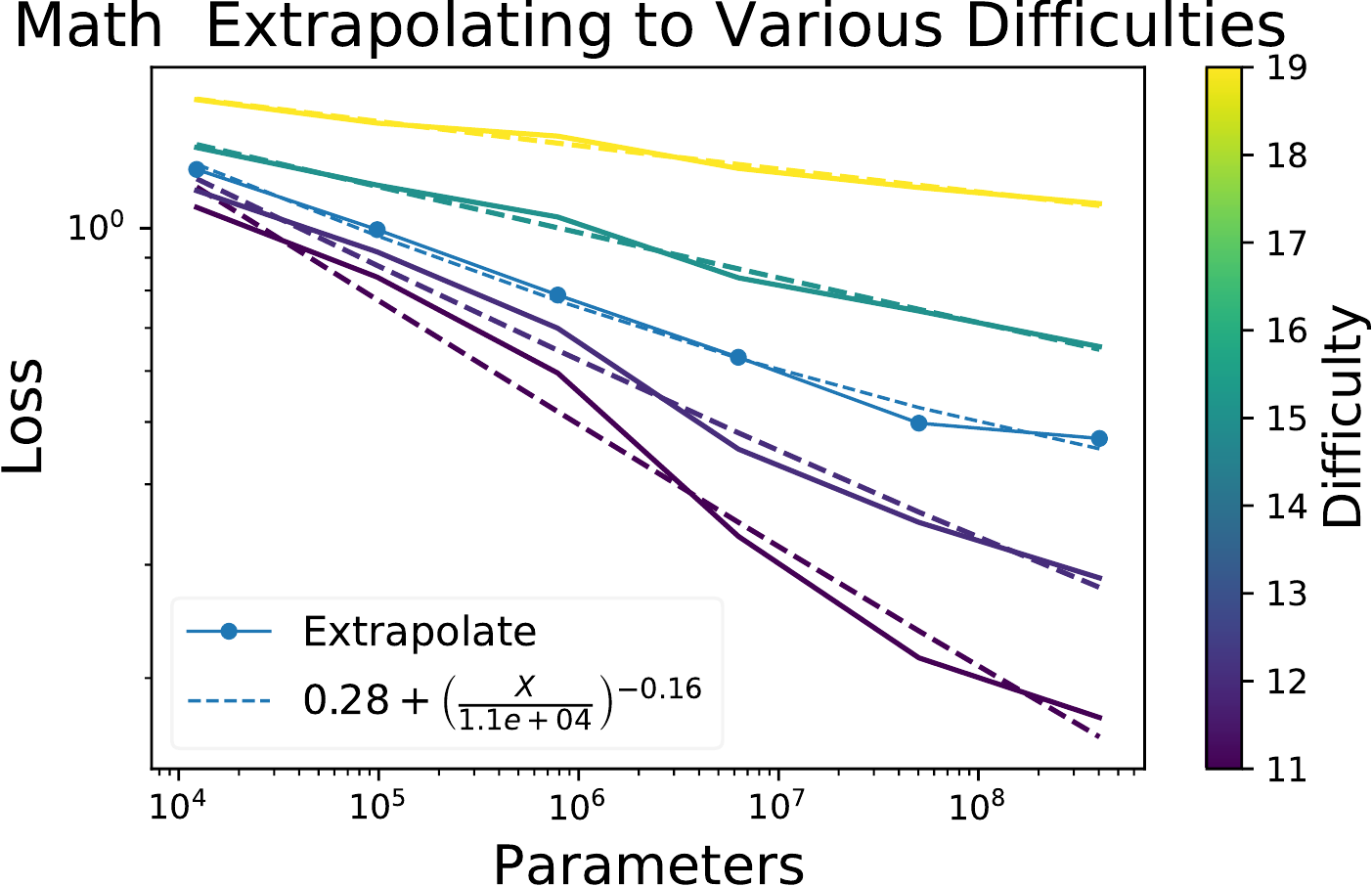}\hfill
\includegraphics[width=0.31\textwidth]{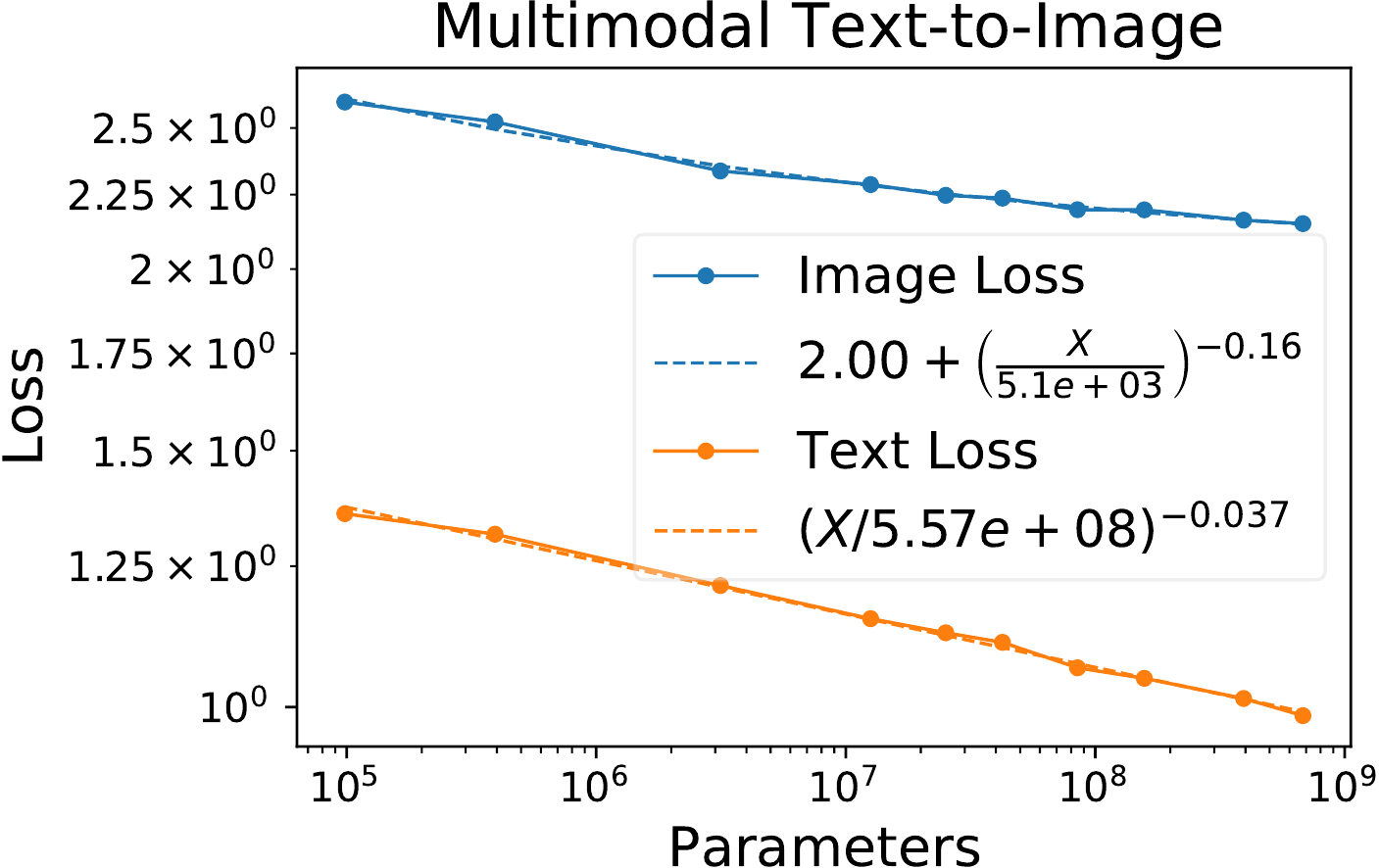}\hfill
\includegraphics[width=0.31\textwidth]{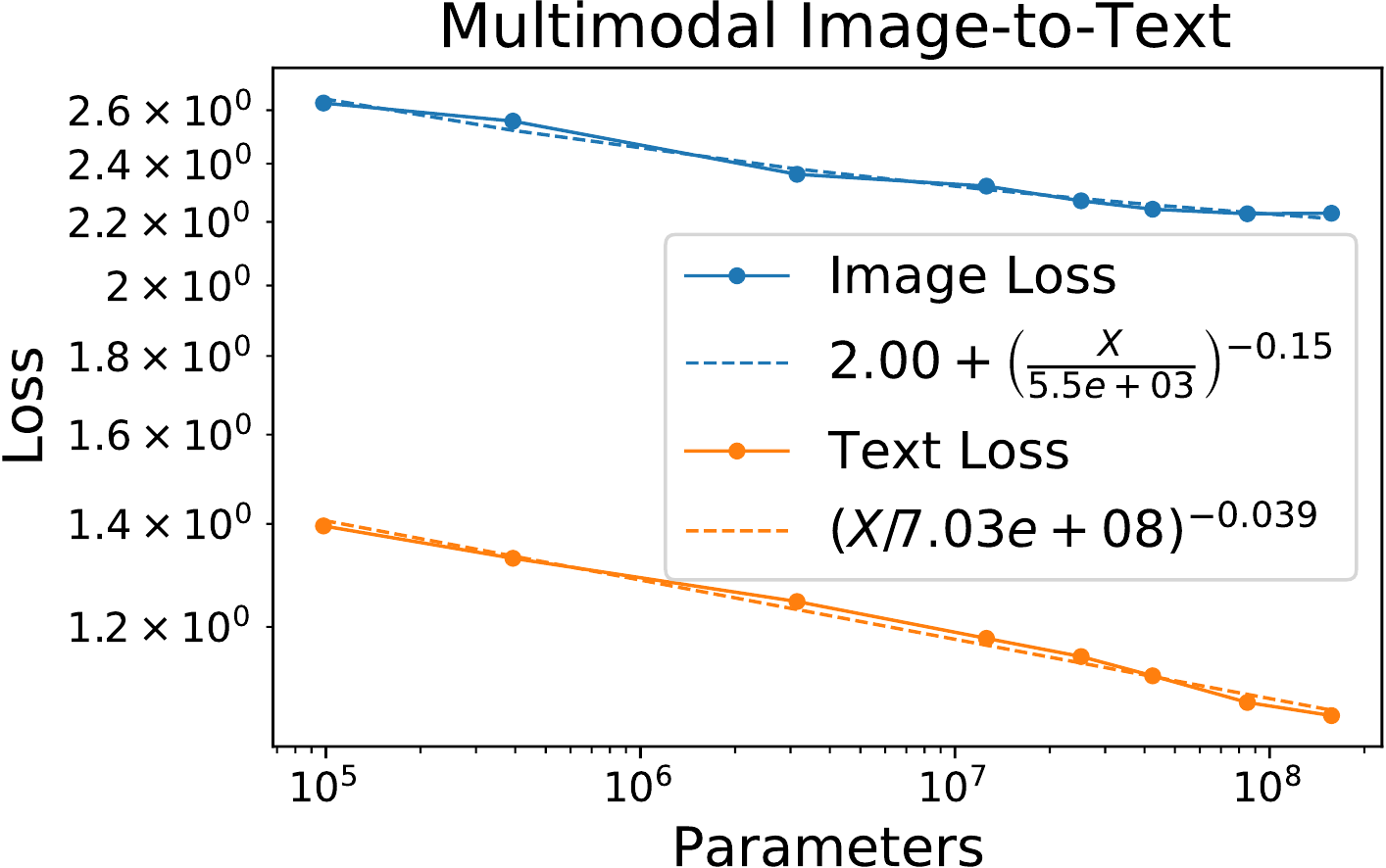}
\caption[Model Size Scaling]{\textbf{Scaling with model size---} We show scaling laws with model size for various domains, along with  fits (dashed) to equation (\ref{eq:PowerLawPlusConstant}).  Note that the largest language models \cite{brown2020language} in the top-left figure were not trained to convergence, so deviations from the trend are not necessarily meaningful. Very small models for video and higher-resolution images are off-trend; we speculate this is due to these models attempting to attend to a context with length comparable to their non-embedding parameter count. \label{fig:ModelSizeScaling}}
\end{figure}

\subsection{Summary of Results}

We apply autoregressive decoder-only Transformer models to all data modalities, which include web-scraped YFCC100M images \cite{DBLP:journals/corr/ThomeeSFENPBL15} of various resolutions, video data from various sources, multimodal image+language data, and procedurally generated math problems.  We also reference prior results on language \cite{kaplan2020scaling, brown2020language}. {\bf Across all domains} we find:

\begin{itemize}
\item The  scaling laws of equation (\ref{eq:PowerLawPlusConstant}) apply consistently, including for very small values of the reducible loss.  Since the $L(C)$ trends can be extended to arbitrarily large data distributions, model sizes, and training steps, we argue that this supports the interpretation of equation (\ref{eq:InfoTheoryInterpretationofLoss}).
\item We identify the optimal model size $N_{\rm opt}(C)$ for a given compute budget, and find that it can be accurately modeled as a pure power law \cite{kaplan2020scaling}
\be
N_{\rm opt} \propto C^{\beta}
\ee
with a power $\beta \sim 0.7$ for all modalities, as shown in figure \ref{fig:OptimalModelSizeAllDomains}.  As  compute budgets grow, it's best to devote a majority of resources towards training larger models.  This strongly suggests sub-linear scaling of $D \propto N^{0.4}$ for dataset size with model size during compute-optimal training.
\item For each domain, there is an optimal aspect ratio $d_{\rm model}/n_{\rm layer}$  for the Transformer.  Most data modalities require smaller aspect ratios (i.e. deeper networks) as compared to language \cite{kaplan2020scaling}.  
\item We study an apparent inconsistency between $L(D)$ and $L(C)$ trends in section \ref{sec:Paradox}.
\end{itemize}

We also find a number of results specific to certain domains, though we expect that many of the lessons are more general.   For {\bf image and video modeling}  (see section \ref{sec:ImageandVideo}):
\begin{itemize}
\item When generative image models are finetuned for ImageNet classification, we find a power-law for  classification loss vs model size (see figure \ref{fig:ImageNetClassificationTrends}), even beyond the model size where we approach the irreducible loss for generative modeling.  We conclude that \textbf{the approach to the irreducible loss does not necessarily indicate diminishing returns for representation quality or semantic content}.
\item We explore scaling trends for individual images and for percentiles of the image loss distribution (see figures \ref{fig:TrendsIndividualImages}, \ref{fig:ImagePercentileTrends}, \ref{fig:MostLeastImprovedImages}, \ref{fig:CompletionsBySize}).  We find that the loss on individual images scales with model size in the same way as the mean over all images in the data distribution.  We expect similar behavior in other data modalities.
\item We test a variety of image resolutions (see figure \ref{fig:ComputeScalingReducibelLossImageResolution}), and find distinct scaling exponents and irreducible losses for each.  We also test two VQVAE \cite{oord2018neural} based models.
\item We examine scaling of the loss with video frame index (see figures \ref{fig:LossvsContext} and \ref{fig:PerFrameVideoScaling}).
\end{itemize}
For {\bf multimodal models} (see section \ref{sec:Multimodal}):
\begin{itemize}
\item We explore the mutual information between captions and images (see figure \ref{fig:InfoGain}), and the information gain defined in equation (\ref{eq:InfogainDefinition}). 
We find a smooth scaling for both the mutual info and information gain with model size $N$.
\item We revisit the question ``Is a picture worth a thousand words?'' by comparing the information-content of textual captions to the image/text mutual information.
\end{itemize}
For {\bf mathematical problem solving} (see section \ref{sec:Math} and appendix \ref{app:MathDetails}):
\begin{itemize}
\item We explore the ability of models to extrapolate from the training distribution to increasingly more challenging problems.  We find that extrapolation performance depends predominantly on performance on the training distribution (figure \ref{fig:MathPerformancebyLevel}), and is otherwise independent of model size.  So while larger models perform better,  model size does not provide  benefits to `strong generalization'.
\item We provide a detailed breakdown of performance by math problem type (see appendix \ref{app:MathDetails}).
\end{itemize}

\section{Central Empirical Scaling Laws in Each Domain}

In this section we will describe our common experiments in each domain and our results establishing equation (\ref{eq:PowerLawPlusConstant}) for compute, model size, and (in a few cases) dataset size scaling.

\begin{table}
    \begin{center}
        \begin{tabular}{ | c | c| c |c|}
        \hline
        Domain &   $L(N)$ (model size) &  $L(C)$ (compute) & $N_{\rm opt}(C)$ \\ 
        \hline
        Language  & $\left( \frac{N}{ 1.47 \times 10^{14} } \right)^{-0.070}$ & 
        $\left( \frac{C}{ 3.47 \times 10^8 } \right)^{-0.048}$ & 
        $\left( \frac{C}{3.3 \times 10^{-13}}  \right)^{0.73} $ \\ 
                \hline
        Image 8x8 &  $3.12 + \left(\frac{N}{ 8.0 \times 10^1}\right)^{-0.24}$ & 
        $3.13 +  \left(\frac{C}{1.8 \times 10^{-8}}\right)^{-0.19}$ &
        $\left( \frac{C}{5.3 \times 10^{-14}}  \right)^{0.64}$ \\ 
	Image 16x16 &  $2.64 + \left(\frac{N}{ 2.8 \times 10^2}\right)^{-0.22}$ &  
	$2.64 + \left(\frac{C}{1.6 \times 10^{-8}}\right)^{-0.16}$ &
	$\left( \frac{C}{4.8 \times 10^{-12}}  \right)^{0.75}$ \\ 
        Image 32x32 & $2.20 + \left(\frac{N}{ 6.3 \times 10^1}\right)^{-0.13}$ & 
        $2.21 + \left(\frac{C}{3.6 \times 10^{-9}}\right)^{-0.1}$ & 
        $\left( \frac{C}{1.6 \times 10^{-13}}  \right)^{0.65}$ \\ 
       % Image 64x64 & $1.93 + \left(\frac{N}{ 2.3 \times 10^3}\right)^{-0.19}$  &  &  \\ 
                \hline  
        Image VQ 16x16 &  $3.99 + \left(\frac{N}{ 2.7 \times 10^4}\right)^{-0.13}$ &
    $4.09+\left( \frac{C}{ 6.1 \times 10^{-7}} \right)^{-0.11}$ & 
    $\left( \frac{C}{6.2 \times 10^{-14}}  \right)^{0.64}$ \\ 
        Image VQ 32x32 &   $3.07 + \left(\frac{N}{ 1.9 \times 10^4}\right)^{-0.14}$  & $3.17+\left( \frac{C}{ 2.6 \times 10^{-6}} \right)^{-0.12}$ & 
        $\left( \frac{C}{9.4 \times 10^{-13}}  \right)^{0.7}$ \\ 
                \hline
        Text-to-Im (Text)  & $\left(\frac{N}{ 5.6 \times 10^8}\right)^{-0.037}$  & 
         (combined text/image loss) &  \\ 
        Text-to-Im (Image)  &  $2.0 + \left(\frac{N}{ 5.1 \times 10^3}\right)^{-0.16}$ & $1.93 + \left(\frac{C}{ 1.5 \times 10^{-6}}\right)^{-0.15}$ & 
        $\left( \frac{C}{9.4 \times 10^{-13}}  \right)^{0.7}$ \\ 
        Im-to-Text (Text) &  $\left(\frac{N}{ 7.0 \times 10^8}\right)^{-0.039}$ &  
         (combined text/image loss) &  \\ 
        Im-to-Text (Image) &  $2.0 + \left(\frac{N}{ 5.5 \times 10^3}\right)^{-0.15}$  & $1.97 + \left(\frac{C}{ 1.5 \times 10^{-6}}\right)^{-0.16}$  &  
        $\left( \frac{C}{3.3 \times 10^{-12}}  \right)^{0.72}$ \\ 
                        \hline
        Video  VQ 16x16x16 &  $1.01 +  \left(\frac{N}{ 3.7 \times 10^4}\right)^{-0.24}$ & $ 0.95 +  \left(\frac{C}{ 2.2 \times 10^{-5}}\right)^{-0.14}$ & 
        $\left( \frac{C}{1.13 \times 10^{-12}}  \right)^{0.71}$ \\ 
                \hline
        Math (Extrapolate)  &  $0.28 +  \left(\frac{N}{ 1.1 \times 10^4}\right)^{-0.16}$ & 
        $0.14 +  \left(\frac{C}{ 1.4 \times 10^{-5}}\right)^{-0.17}$ &
        $\left( \frac{C}{2.3 \times 10^{-12}}  \right)^{0.69}$ \\ 
        \hline
        \end{tabular}
    \end{center}
    \caption{\textbf{Summary of scaling laws---} In this table we summarize the model size and compute scaling fits to equation (\ref{eq:PowerLawPlusConstant}) along with $N_{\rm opt}(C)$, with the loss in nats/token, and compute measured in petaflop-days.    In most cases the irreducible losses match quite well between model size and compute scaling laws.  The math compute scaling law may be affected by the use of weight decay, which typically hurts performance early in training and improves performance late in training.  The compute scaling results and data for language are from \cite{brown2020language}, while $N_{\rm opt}(C)$ comes from \cite{kaplan2020scaling}. Unfortunately, even with data from the largest language models  we cannot yet obtain a meaningful estimate for the entropy of natural language.  
    \label{table:AllScalingExponents}}
\end{table}

\subsection{Domain Descriptions and Training Setups}

In every domain we use decoder-only transformer models trained using an autoregressive cross-entropy loss.  For many models we use a sparse attention pattern \cite{DBLP:journals/corr/abs-1904-10509}, though we use  dense attention when solving math problems. 

The transformers used for language and multimodal modeling  have fully connected layers of size $4 d_{\rm model}$ and attention layers of size $d_{\rm model}$, in the notation of \cite{kaplan2020scaling, brown2020language}.  For  math, image, and video modeling we scale the FC layers  to $d_{\rm model}$ and the attention layers to $d_{\rm model}/4$.  We use an aspect ratio $d_{\rm model}/n_{\rm layer} \approx 10$ for math, images, and videos as we find that this is approximately optimal, meaning that these domains prefer much deeper models as compared to language \cite{kaplan2020scaling}, where the optimal aspect ratio $\sim 100$.  Thus our math, image, and video models are essentially identical, differing only in context length.  For math alone we used a weight decay \cite{DBLP:journals/corr/abs-1711-05101} of $0.05$.  We provide more detailed hyperparameter settings in appendix \ref{app:hyperparameters}.

\subsubsection{Language}

We show results from GPT-3 \cite{brown2020language} for comparison, including the performance of much larger models than we train in other domains.  In figure \ref{fig:OptimalModelSizeAllDomains} we use the optimal model size trend from \cite{kaplan2020scaling}.  In appendix \ref{app:MoreLanguage} we show some experiments on the scaling of arithmetic and factual question answering abilities, and make some additional qualitative observations about the progression of language understanding with scale.

\subsubsection{Images}
\label{desc:images}

We study a dataset of approximately $10^8$ web images \cite{DBLP:journals/corr/ThomeeSFENPBL15} scaled to pixel resolutions $R\times R$ = 8x8, 16x16, and 32x32 represented in raster order using RGB colors, each in the range $[0, 255]$, giving a total of $3R^2$ tokens per image.  We also study the same  images at 64x64 resolution but VQ \cite{oord2018neural} encoded  with either a 16x16 or 32x32 VQ encoding pattern, for a total of either 256 or 1024 tokens per image. To reduce compute, we use sparse attention patterns \cite{DBLP:journals/corr/abs-1904-10509}, alternating between locally-banded attention and fixed-stride attention in sequential layers, where both the local context length and fixed-stride length are given by the side-length in tokens of the square images.

\subsubsection{Video}
\label{desc:video}

We study a dataset of approximately $7 \times 10^5$ videos totaling about $100$ hours scraped from the web, where each frame is scaled to a pixel resolution of 64x64.  Each individual frame is encoded with the same 16x16 VQVAE \cite{oord2018neural} used for images, resulting in 256 tokens per frame. We train on sequences of 16 sequential frames, resulting in a total of 4096 tokens per video. As with images, we reduce compute by using a sparse attention pattern \cite{DBLP:journals/corr/abs-1904-10509} alternating between locally-banded and fixed-stride attention, where both the local context length and fixed-stride length are given by the side length in tokens of the square frames.  

\subsubsection{VQ Encoding}

The VQVAE models mentioned in \ref{desc:images} and \ref{desc:video} were trained on frames of the web-scraped videos described in \ref{desc:video}, using the VQ-VAE architecture \cite{oord2018neural} with modifications described in \cite{dhariwal2020jukebox}, including dead code revival. More details can be found in table \ref{table:VQs}.

\begin{table}
    \begin{center}
        \begin{tabular}{ | c | c| c |}
        \hline
        Input Resolution (pixels) & Output Resolution (VQ Codes) &  Codebook Size \\ 
        \hline
        64x64 & 16x16 & 4096 \\
        \hline
        64x64 & 32x32 & 1024 \\
        \hline
        \end{tabular}
    \end{center}
    \caption{Details of VQVAEs used to encode images and frames of video. 
    \label{table:VQs}}
\end{table}

\subsubsection{Multimodal Text and Images}

Multimodal models are trained to autoregressively predict both image tokens and language tokens in series.  We simply concatenate together the token lists for BPE encoding of text (using the tokenization of \cite{brown2020language}) and the $[0,255]$ colorscale of each of the RGB pixels in the images, and let the model learn the necessary embedding matrix.   We separately study models for text-to-image and image-to-text mappings, as we found poor performance for bidirectional models in preliminary experiments.  For both image-to-text and text-to-image models we compute the mean pixel and mean text token loss, and then weight them to form the total loss $L = 9 L_{\rm image} + L_{\rm text}$, as we found this weighting produced good results in a scan.  We use 32x32 images together with a 128-token captions (padded or trimmed as needed), for a total context length of $3200$ tokens per image/caption pair.  For the multimodal dataset we used a wide variety of image/text pairs curated through web search.

\subsubsection{Mathematical Problem Solving}

Mathematical problem solving would seem to be a rather different domain from generative language, image, video, and multimodal modeling.  To solve math problems, a model needs to learn to execute an algorithm to arrive at a deterministic answer.  In contrast, the other distributions we have studied are typically genuinely probabilistic, and at least at an intuitive level, seem to require something a bit different from the simple algorithms that perform arithmetic or solve equations.  We have included math problems to probe the generality of scaling laws and transformer performance.

We train and test models using the math problem generator \cite{DBLP:journals/corr/abs-1904-01557}, which generates a variety of problems in  algebra, arithmetic, calculus, comparisons, numbers (integer properties), measurement, polynomials, and probability.  When studying model and compute-budget scaling we procedurally generate the training problems in an online setting.  We sample the default mixture of easy, medium, and hard problems, without a progressive curriculum.   When studying dataset size scaling we use static training data sampled from the same distribution.  As discussed further in appendix \ref{app:MathDetails}, the data distribution has some unusual features, as easier problems will naturally appear  more often than more difficult problems.

A few problem types require interpreting both numbers and strings as sequences of individual characters, so for simplicity we model all questions and responses at the character (byte) level.  
The  model receives the problems as plain text, and we fill a transformer's $512$-token context window with concatenated problems, using a mask so that only the tokens corresponding to answers contribute to the loss. 

%For evaluation, the dataset includes an `interpolate' and `extrapolate' setting.  
The problem generator\footnote{The generator settings vary somewhat among problem types, with some depending on more parameters.} \cite{DBLP:journals/corr/abs-1904-01557} can be provided with an `entropy'  $s$.  The training distribution samples from  $s \in [3,10]$, while interpolate-testing corresponds to $s=8$, and the extrapolate test involves $s=12$, along with some other extensions to increase compositionality.  In the online setting, we cannot be sure the interpolate tests are deduplicated from the training data, but the extrapolate test must be.  To supplement the test data and further study extrapolation, we generated new test sets with $s \in [1,19]$, with larger $s$ posing a greater challenge to the model, as $s > 10$ is literally out of the training distribution, and requires extrapolation.  

We found consistently poor performance on the two  extrapolation  generators \texttt{probability\_\_swr\_p\_level\_set\_more\_samples} and \texttt{probability\_\_swr\_p\_sequence\_more\_samples} from \cite{DBLP:journals/corr/abs-1904-01557}, with larger models overfitting against them and achieving worse loss (but higher accuracy) than some smaller models.  So \emph{we have not included their contribution} in figures \ref{fig:ComputeScaling} and \ref{fig:ComputeScalingFullLoss}, as the poor loss on these modules would dominate the trends.

We provide more details and many additional results on math in appendix \ref{app:MathDetails}, including results per module, dataset size\footnote{The math models in figure \ref{fig:OptimalAspectRatio} used $m_{mlp} = 4, m_{attn}=1$ like language models, unlike the math models used to study model and compute scaling, as these aspect ratio tests were performed earlier. } scaling, and further analysis of performance vs difficulty level.  There we also show trends for the training loss, which do not adhere as well to a power-law form, perhaps because of the implicit curriculum in the frequency distribution of easy and hard problems.

\subsection{Model Size Scaling and Aspect Ratios}

Arguably the simplest scaling relation compares the loss achieved by models of various sizes $N$ once they are trained to convergence with a dataset large enough to obviate overfitting.   Throughout this paper we report $N$ as the number of non-embedding parameters in a transformer model, motivated by prior results on language \cite{kaplan2020scaling}.  Results for the scaling of $L(N)$ are depicted in figure \ref{fig:ModelSizeScaling}, along with fits to equation (\ref{eq:PowerLawPlusConstant}).  

We define $L(N)$ using the loss at convergence (practically, this means as close to convergence as is feasible), but the largest models we study will not have fully converged.  Thus caution is warranted when interpreting $L(N)$ trends according to equation (\ref{eq:InfoTheoryInterpretationofLoss}) and identifying the irreducible loss as an entropy, and the reducible loss as a KL divergence.  Nevertheless, the reducible losses typically fit very well to a pure power-law trend.  As an aside, we often find intriguingly good power-law plus constant trends when recording the loss after training all models for a fixed number of training steps.

\begin{figure}
\noindent \centering{} 
\includegraphics[width=0.31\textwidth]{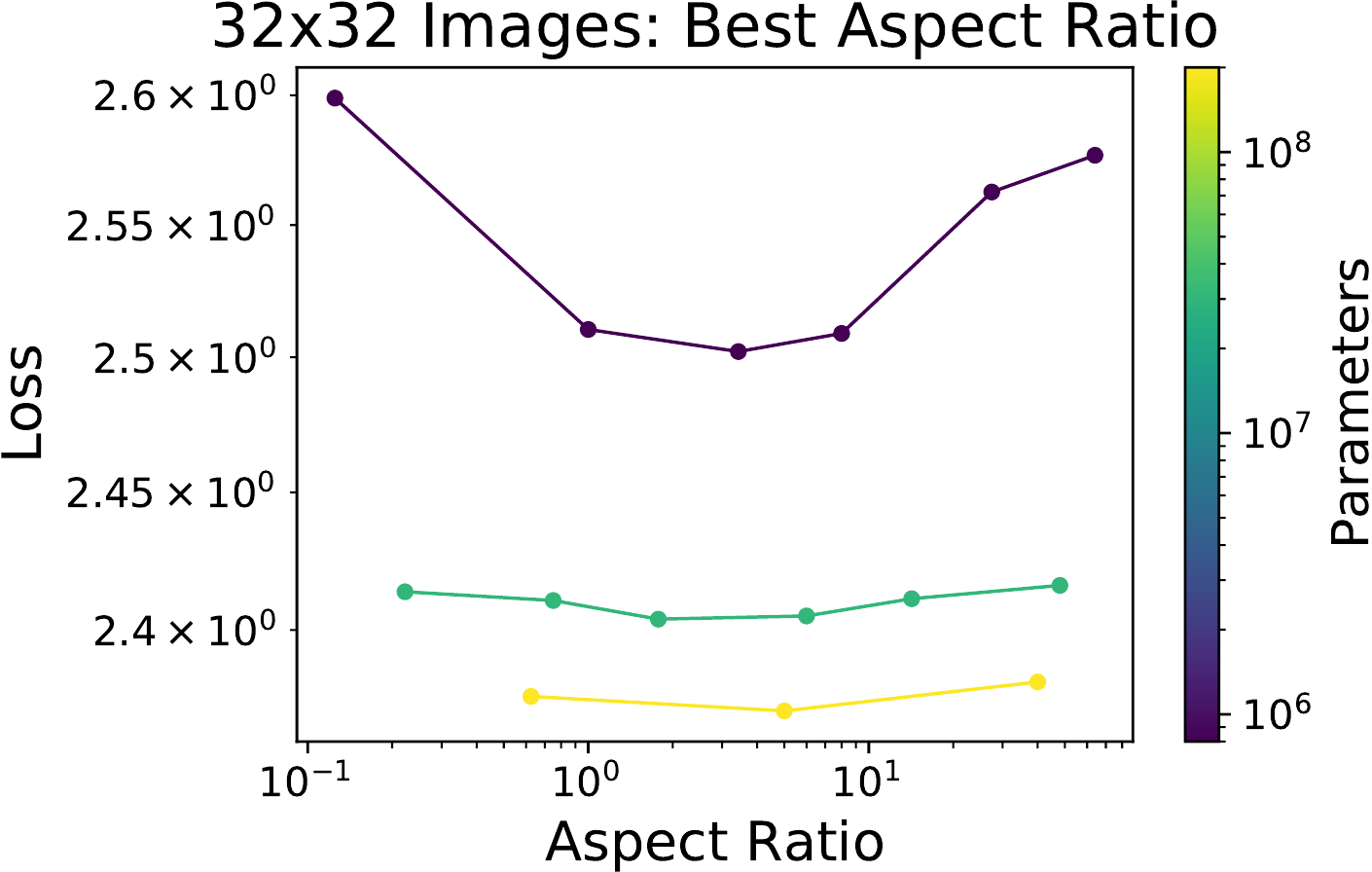}\hfill
\includegraphics[width=0.31\textwidth]{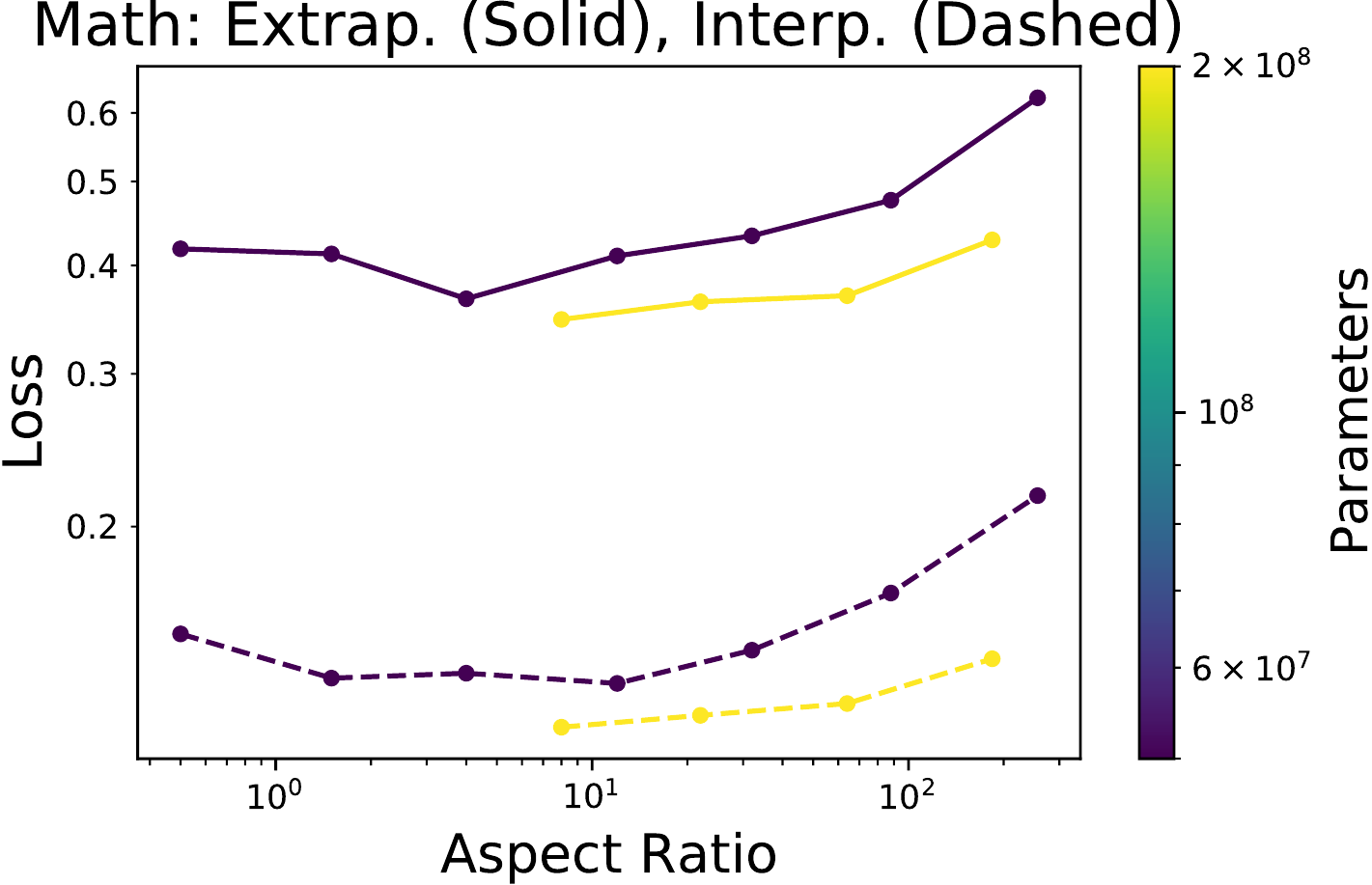}\hfill
\includegraphics[width=0.31\textwidth]{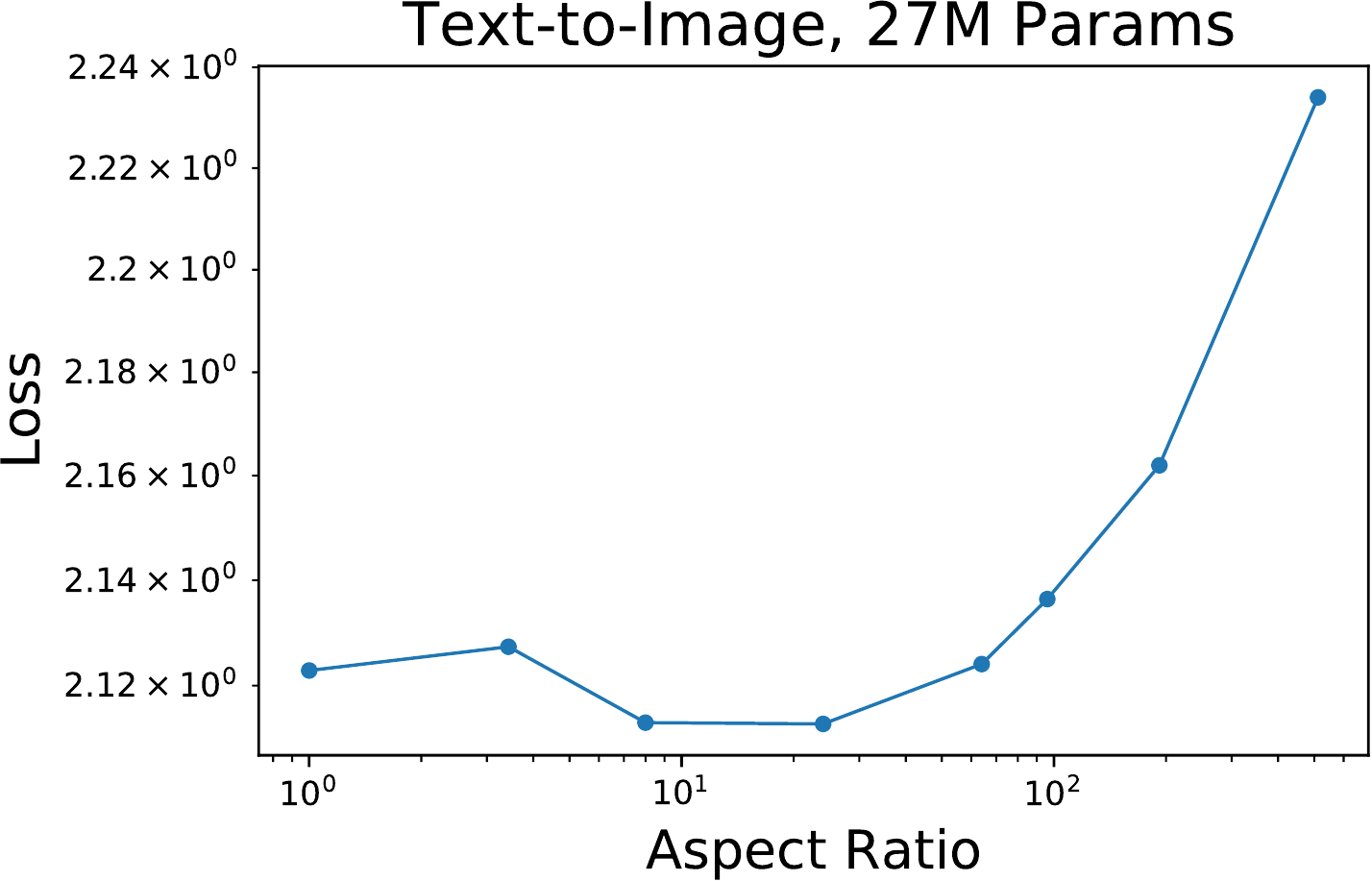}
\caption[Optimal Aspect Ratio]{\textbf{Optimal aspect ratio---} We show trained performance as a function of the aspect ratio, defined as width / depth, or more precisely $\equiv d_{\rm model} / n_{\rm layer}$.  The optimal aspect ratio for language \cite{kaplan2020scaling} was about 10x larger. \label{fig:OptimalAspectRatio}}
\end{figure}

We have found that for any given data modality, transformer models typically have an ideal aspect ratio $d_{\rm model} / n_{\rm layer}$ that maximizes performance while holding model size $N$ fixed.  In figure \ref{fig:OptimalAspectRatio} we display converged performance as a function of aspect ratio for a few model sizes in several domains.  We see that image and math models perform optimally with an aspect ratio $\approx 5$, which suggests that on these domains we should aim for deeper and thinner models, with at least a 10x smaller aspect ratio compared to optimized language models.  The difference may be even greater due variations in $m_{\rm attn}$ and $m_{\rm mlp}$ settings.

Finally, note that  image and video models with roughly $10^4$ parameters under-perform the trends, with worse performance evident for higher resolution images.  The video models must attend to a $4096$-token context, while 32x32 images have a $3072$-token context, so we speculate that tiny models under-perform because they have difficulty attending to contexts comparable in length to their non-embedding parameter count.  

\subsection{Compute Scaling and Optimal Model Sizes}

\begin{figure}
\noindent \centering{} 
\includegraphics[width=0.31\textwidth]{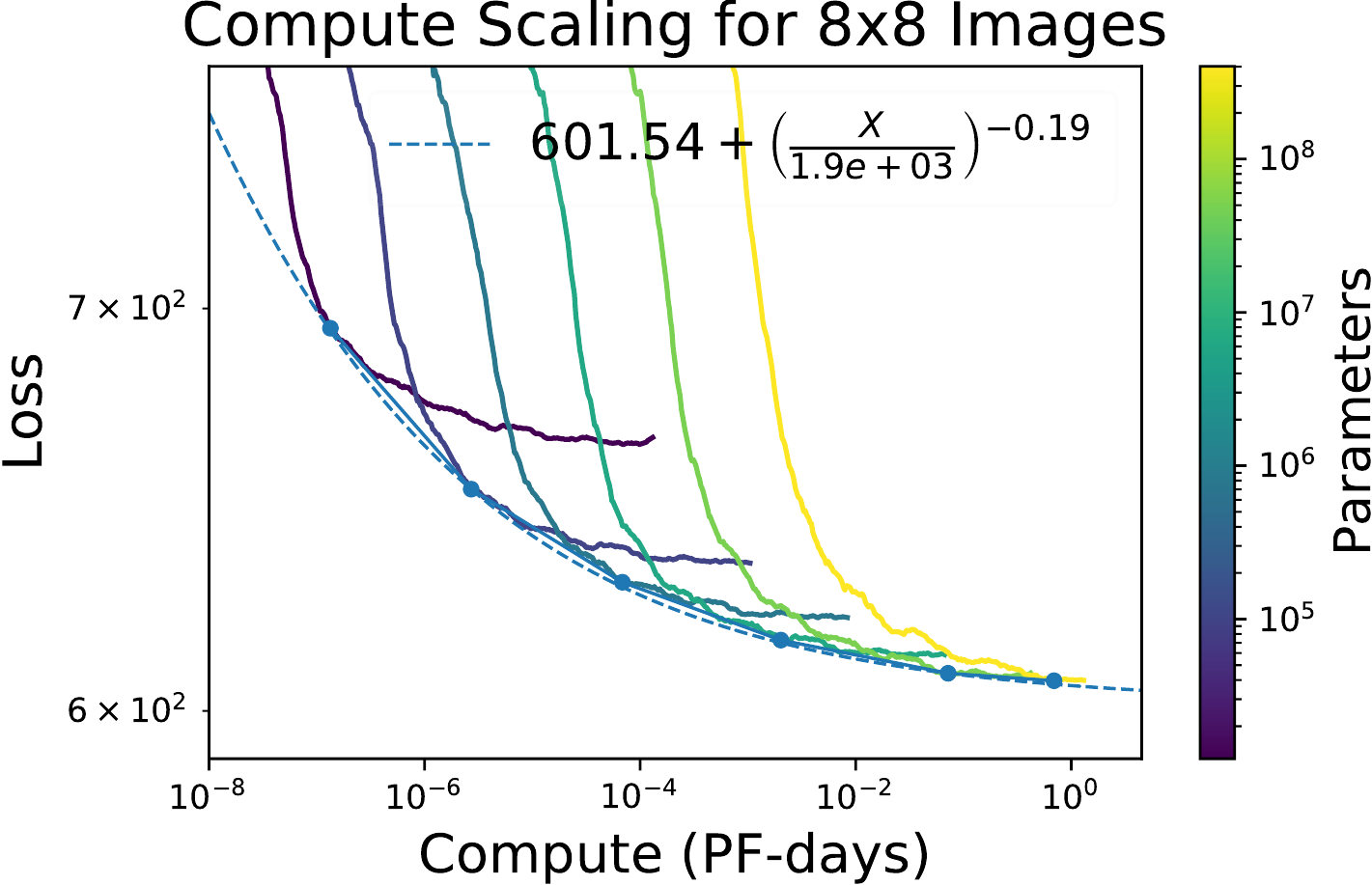}\hfill
\includegraphics[width=0.31\textwidth]{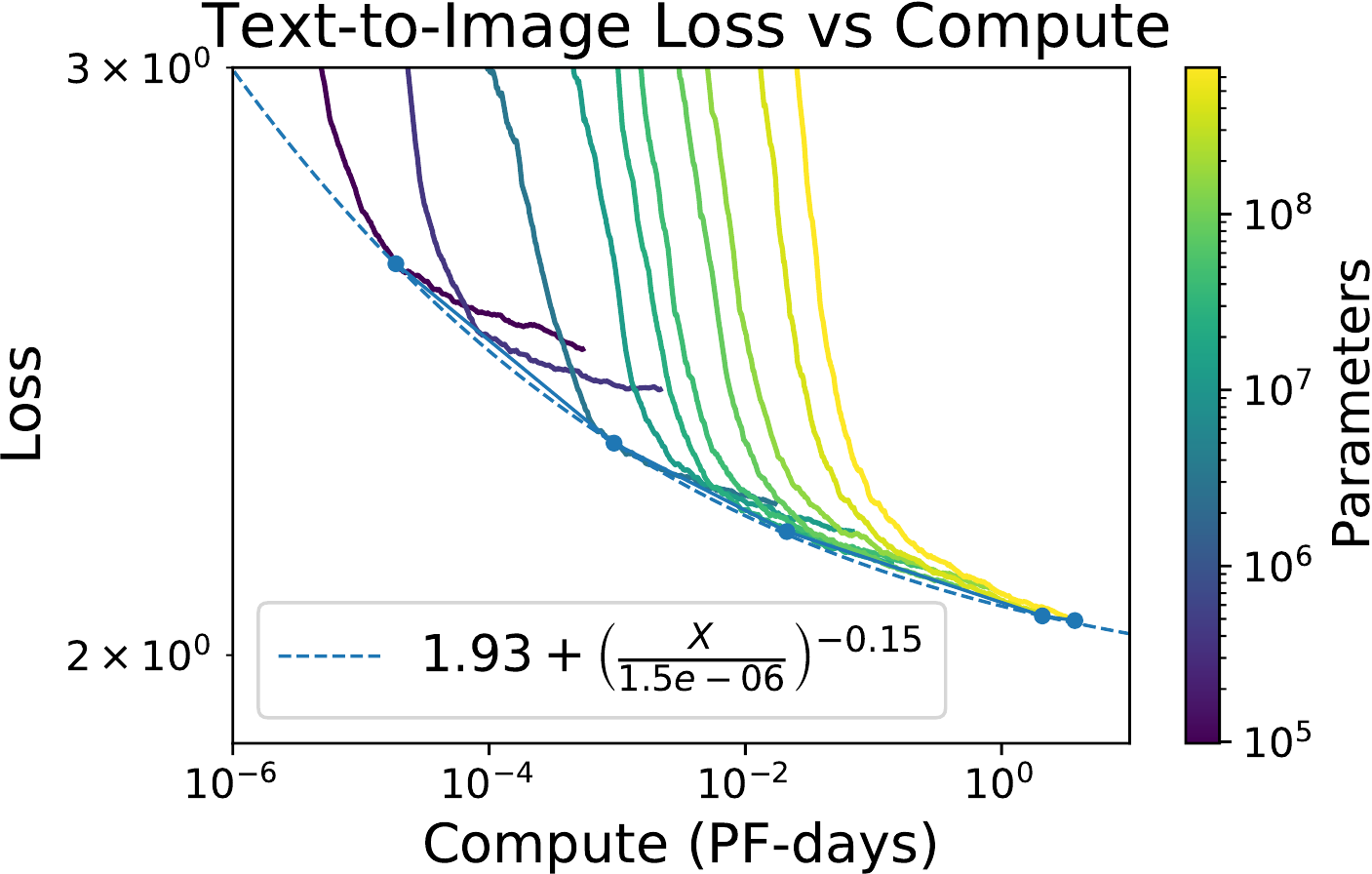}\hfill
\includegraphics[width=0.31\textwidth]{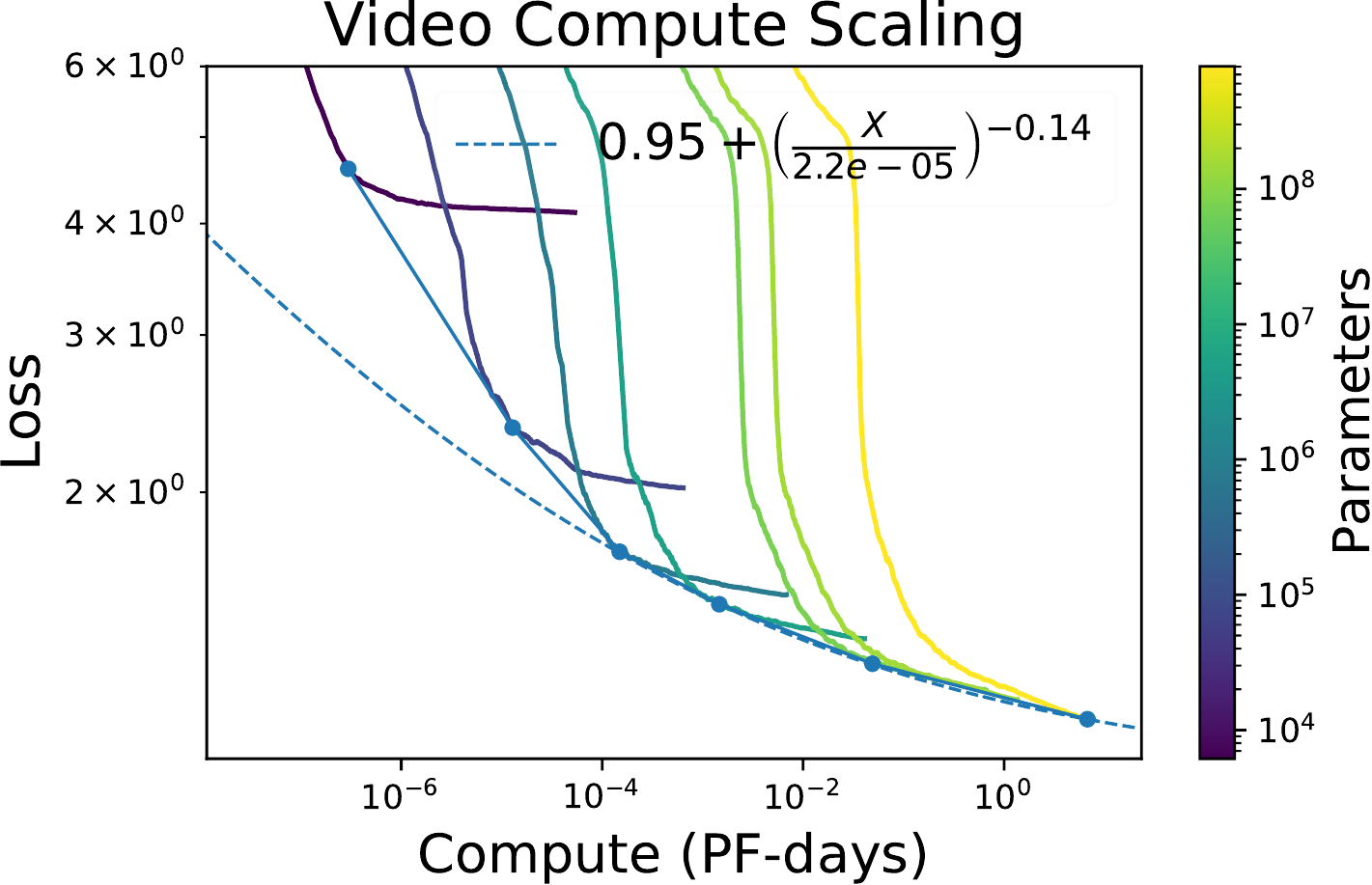}
\\
\vspace{1em}\includegraphics[width=0.31\textwidth]{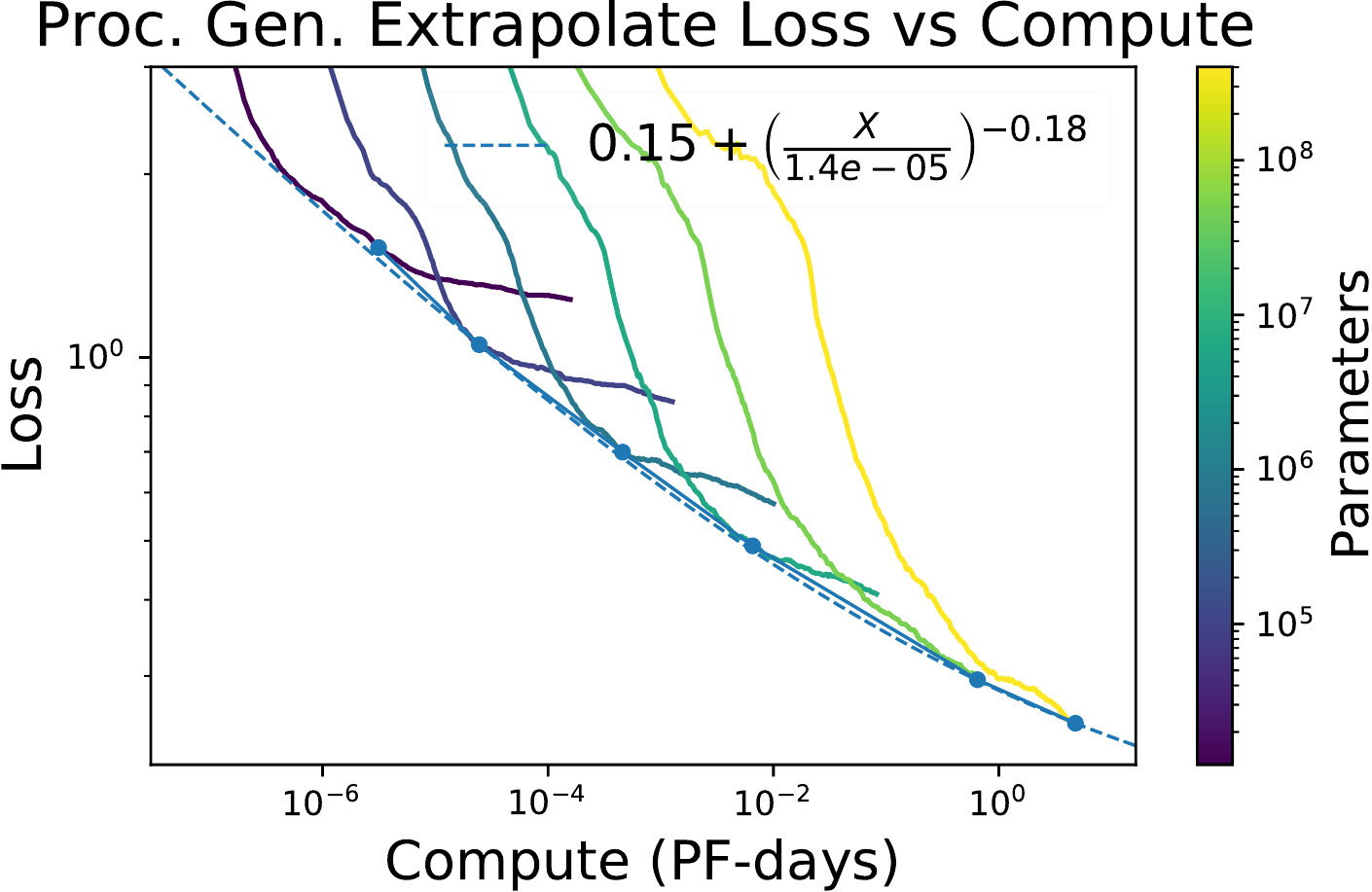}\hfill
\includegraphics[width=0.31\textwidth]{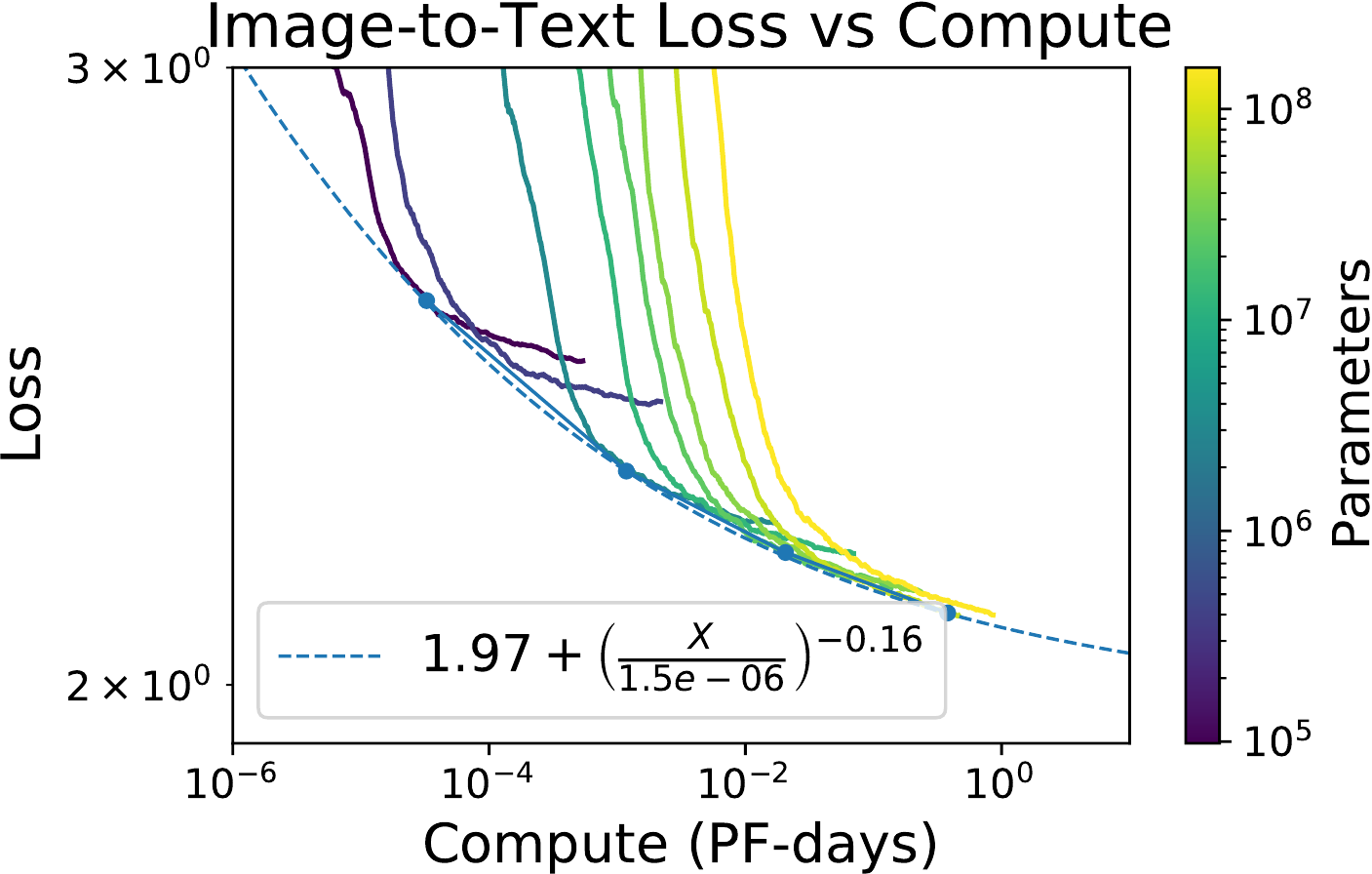}\hfill
\includegraphics[width=0.31\textwidth]{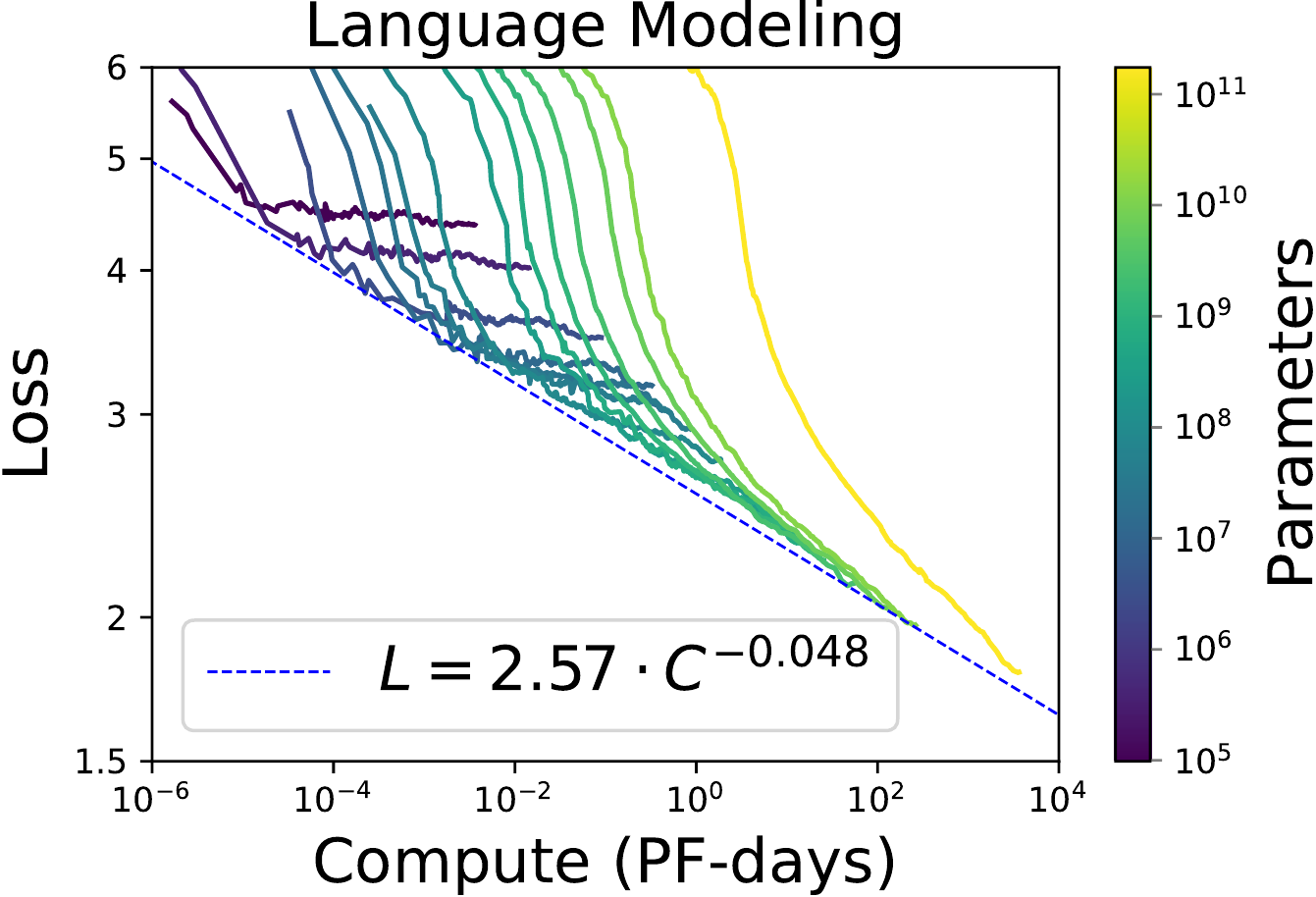}
\caption[Compute Scaling]{\textbf{Scaling laws with compute---} Scaling laws with compute (total estimated floating point operations) for various domains, along with power-law plus constant fits (dashed). This is identical to figure \ref{fig:ComputeScaling}, except that we do not subtract the fitted constant irreducible loss.  Note that very small models underperform compared to the trends when they model images or videos with very large contexts.  Note also that the largest language models \cite{brown2020language} were not trained to convergence. \label{fig:ComputeScalingFullLoss}}
\end{figure}

Instead of focusing on converged performance, one can study the loss $L$ achieved with a finite training compute budget $C$ when training with a large enough dataset to avoid overfitting.  We define $C$ theoretically rather than empirically, and approximate\footnote{The factor of 6 includes a factor of 2 for add-multiply and a 3 to include forward and backward passes.} it as $C \equiv 6 N E$ where $N$ is the non-embedding parameter count (model size) and $E = S B$ is the total number of tokens processed during training (with $S$ the number of parameter updates and $B$ the batch size in tokens).    The results for $L(C)$ from a variety of model sizes are depicted in figure \ref{fig:ComputeScalingFullLoss}, along with the pareto-frontier of optimal loss for a given compute budget, and a power-law plus constant fit forced to lie below this frontier. 

The compute  trends are most relevant for differentiating between the irreducible loss and reducible losses, since they avoid the issue of training to convergence, which makes the interpretation of $L(N)$ difficult.   We display the reducible loss trends for $L(C)$ in figure \ref{fig:ComputeScaling}, and emphasize that these appear to be pure power-laws, even when the reducible loss is much smaller than the irreducible loss.  

We can use the $L(C)$ trends to estimate the model size $N_{\rm opt}$ that optimizes the loss when training is constrained by a fixed compute\footnote{For a fixed amount of training compute, we can train smaller models at the expense of worse performance. Hence, when accounting for both inference and training compute, the optimal model size may be somewhat smaller than described here. See \cite{kaplan2020scaling} for a discussion this tradeoff.} budget $C$.  For this purpose we select points on the convex hull of the loss versus compute frontier; these can be seen as blue points in figure \ref{fig:ComputeScalingFullLoss}. The results for all domains together appear in figure \ref{fig:OptimalModelSizeAllDomains}, while each domain is shown separately with individual fits in figure \ref{fig:OptimalModelSizeScaling}.  In all cases we find that $N_{\rm opt}(C) \propto C^{\beta}$ can be fit  with a pure power-law, with all exponents fairly close to $\beta \sim 0.7$.  This suggests that one should spend most of a growing training compute budget by training much larger generative models.

When estimating $N_{\rm opt}(C)$, one might worry about errors due to a sub-optimal usage of data.  Specifically, if the batch size is too large early in training, then some compute may effectively be wasted.  This can be studied by identifying the critical batch size \cite{DBLP:journals/corr/abs-1712-06559, 1812.06162} above which there are diminishing returns to further data parallelism.  In prior work \cite{kaplan2020scaling} this was taken into account by measuring the critical batch size and using relations derived in \cite{1812.06162} to adjust compute estimates.   We have not made this adjustment here, as it would require a number of additional experiments in order to measureme the critical batch size in each domain.  For large model sizes and compute budgets these effects should be small, because most or all of training involves batches smaller than the critical batch size (which grows quickly during training \cite{1812.06162}), but this issue may be worth revisiting in the future.  

The total number of tokens processed during all of training is $E = \frac{C}{6N} \geq D$, where $D$ is the dataset size, with equality representing training for only a single epoch.  This means that $D \propto C^{1- \beta} \propto N^{\frac{1- \beta}{\beta}}$.  We clearly have $\beta > 0.6$ for all data modalities and by a comfortable margin, suggesting that dataset size should not grow faster than $D \propto N^{2/3}$ during compute-optimal training, with a more reasonable median estimate of $D \propto N^{0.4}$. This unambiguously sub-linear scaling across all data modalities runs somewhat counter to conventional wisdom.  As a word of caution, we have yet to train models in a regime where compute optimal training actually implies $D \ll N$ numerically.  We discuss this further in section \ref{sec:Paradox}.

\subsection{Loss versus Position in the Context Depends on the Structure of the Data}

\begin{figure}
\noindent \centering{} 
\includegraphics[height=0.22\textwidth]{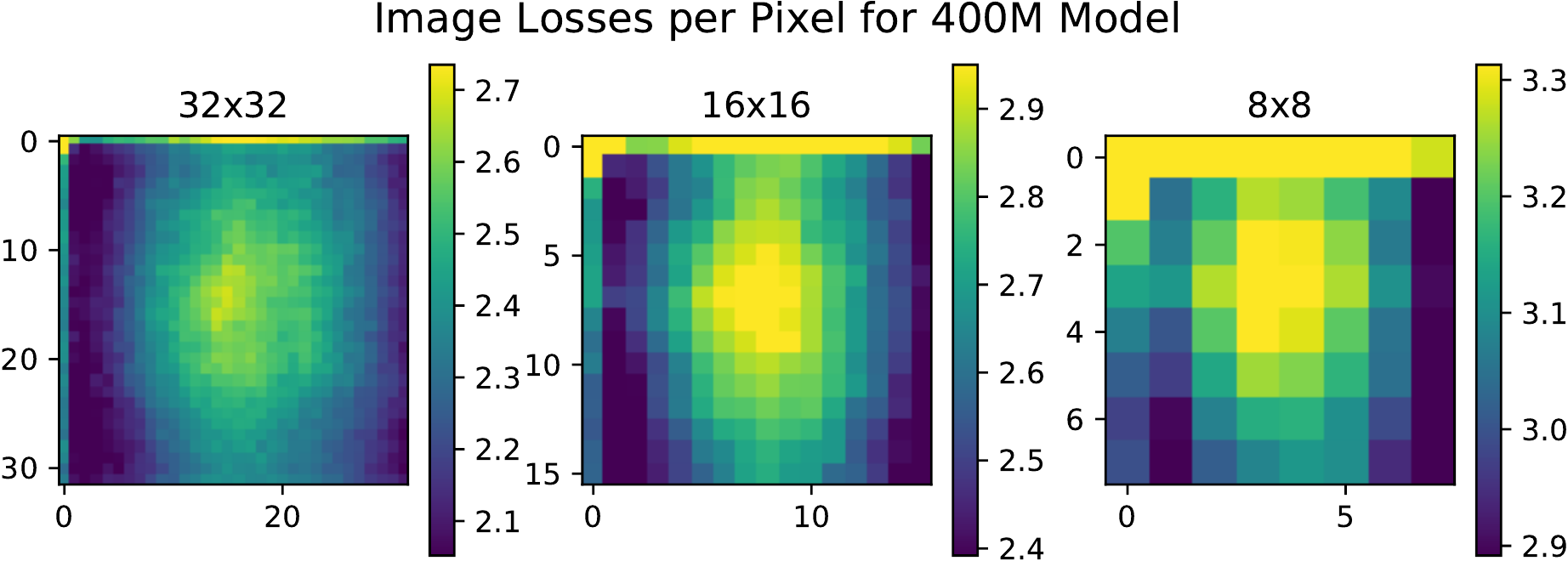}\hfill
\includegraphics[height=0.22\textwidth]{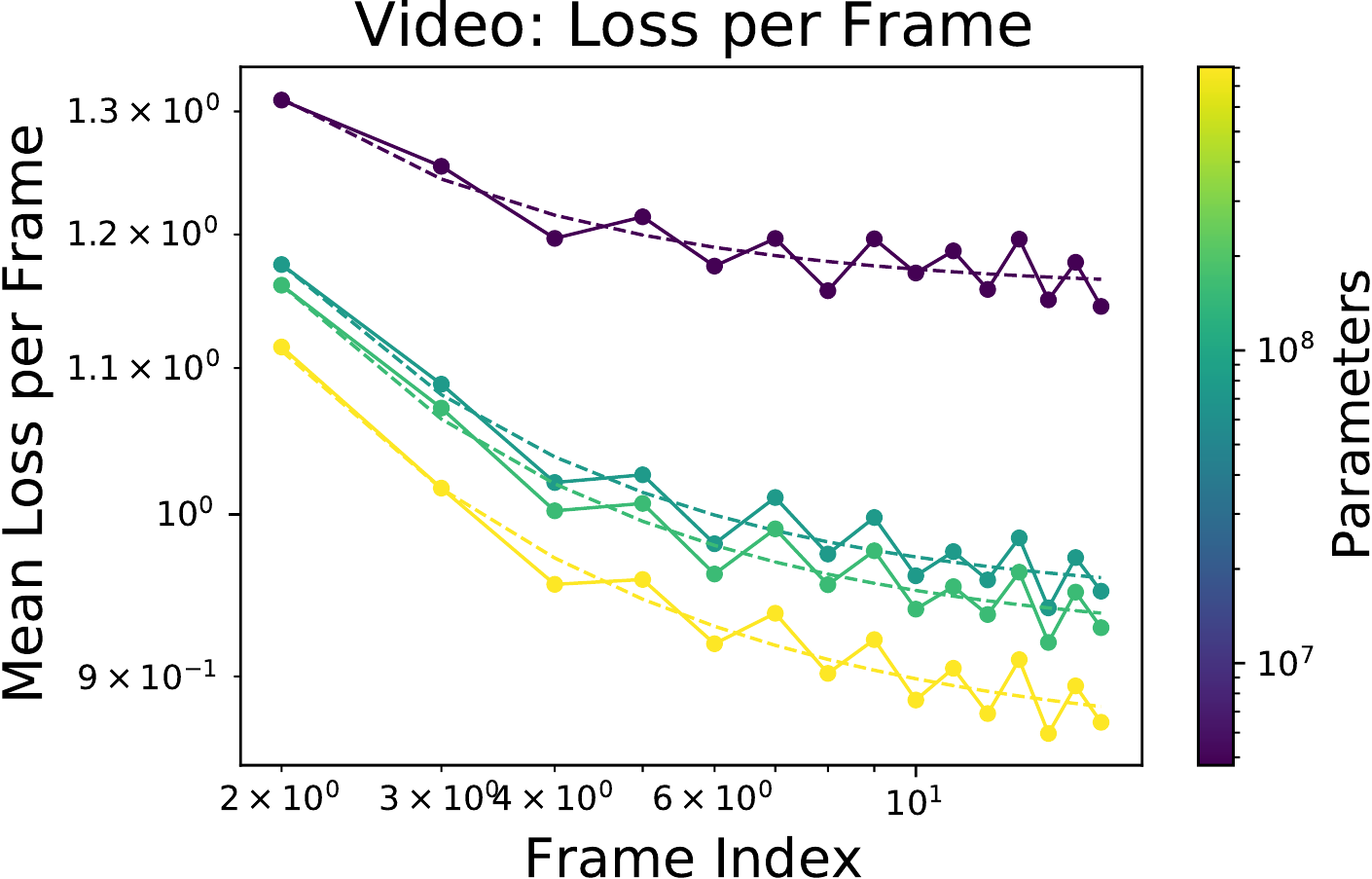}
\caption[Loss per Frame]{\textbf{Position-dependent loss for images and video---} We show trends for the loss as a function of position in the context for image and video models.  On the left we have the mean loss over the three colors for images of various resolutions.  The top-left pixel actually has significantly higher loss, off the color scale, which was set to make the pattern clear for the image as a whole.  On the right we see the mean loss per frame for video models, as a function of the frame index.  The oscillatory behavior per frame is due to the video encoding.  \label{fig:LossvsContext}}
\end{figure}

Some trends in the loss are highly dependent on the structure of the data.  A clear example of this is the loss as a function of the position in the context, ie the loss per token for language models, loss per frame for video models, or the loss per pixel in visual domains.  We provide two examples in figure \ref{fig:LossvsContext}. Note that for images the very first pixel typically has a large loss, outside the color range shown; we chose not to extend the color range as it would have obscured the patterns in the remainder of the image.

Language \cite{kaplan2020scaling} and videos (per frame) show a power-law plus constant trend as a function of context position, as their data is naturally sequential.  However, these trends do not apply at all to image modeling, where the loss is largest for the first pixels and near the center of the image.  Thus power-law correlations in the context depend in an essential way on the nature of the data, and are not universal.  In contrast, the form of the compute and model size scaling laws appears to be largely independent of the data distribution.

\section{Image and Video Modeling, the Reducible Loss, and Downstream Tasks}
\label{sec:ImageandVideo}

Image data can be presented at a wide variety of resolutions, or it may be compressed, for example with  VQ codes \cite{oord2018neural}.  These settings provide a way to modify the complexity of the data distribution, creating a useful arena for the study of neural scaling laws. Furthermore, we can finetune generative image models for classification to explore the quality of their learned features.  

We will use these tools to explore the nature of the reducible and irreducible loss.  In particular, at very low resolution (8x8) we can follow the power-law trend in the reducible loss all the way to a few nats/image, which can be achieved by models approaching a billion parameters.  This gives us some reason for optimism when extrapolating similar trends on larger images beyond the realm that we can currently explore.  It also strongly suggests that the power-law plus constant form of equation (\ref{eq:PowerLawPlusConstant}) will remain an excellent approximation.

Furthermore, we will show that improvement in fine-tuned classification performance continues smoothly even as the generative loss approaches the irreducible loss. This result strongly suggests that representation quality continues to improve smoothly even when the generative loss trend appears to taper off.

\subsection{Varying the Image Resolution and Encoding}

\begin{figure}
\noindent \centering{} 
\includegraphics[width=0.45\textwidth]{ImageLossvsModelSize}\hfill
\includegraphics[width=0.45\textwidth]{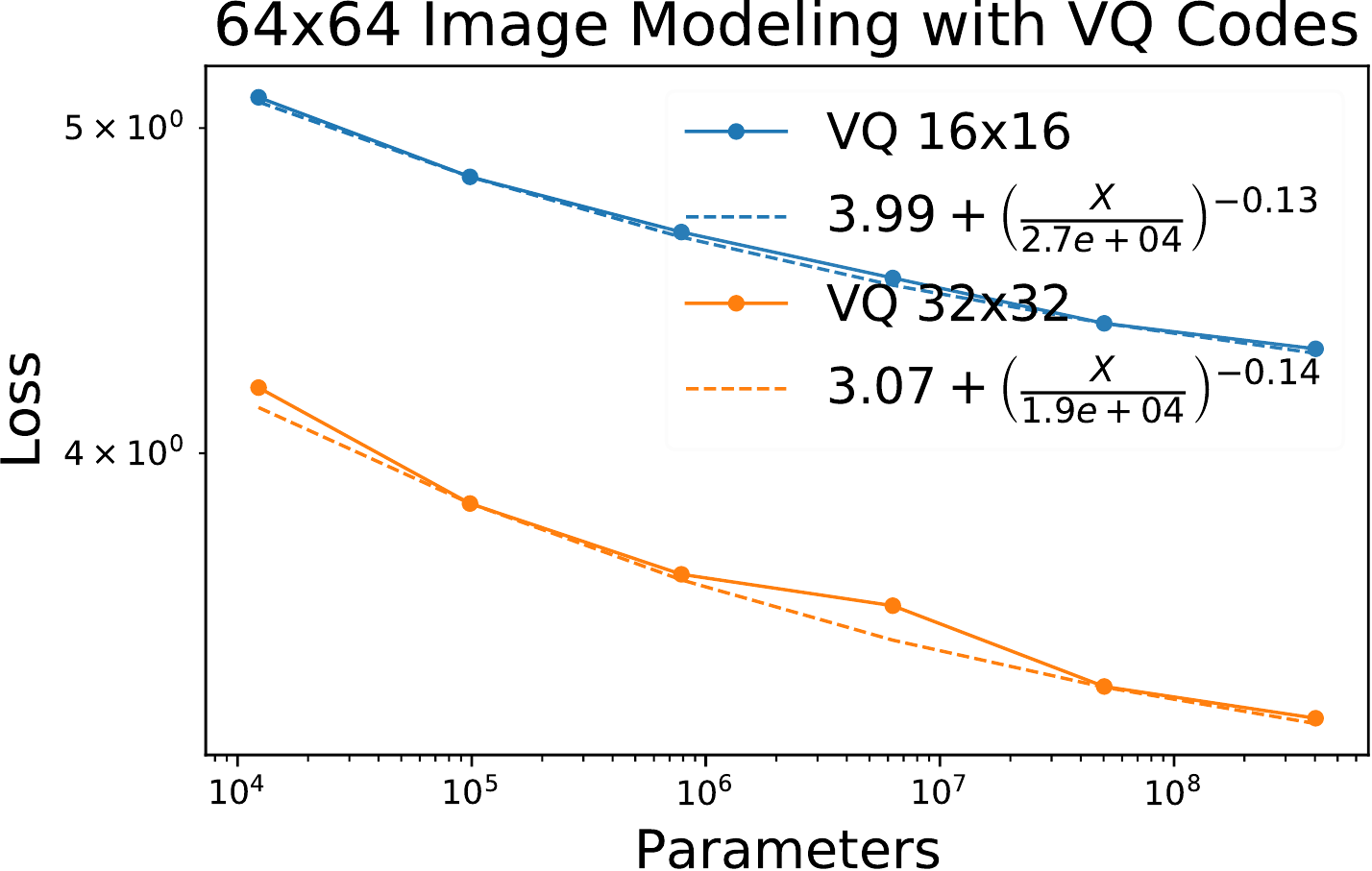}\\
\vspace{1em}
\includegraphics[width=0.45\textwidth]{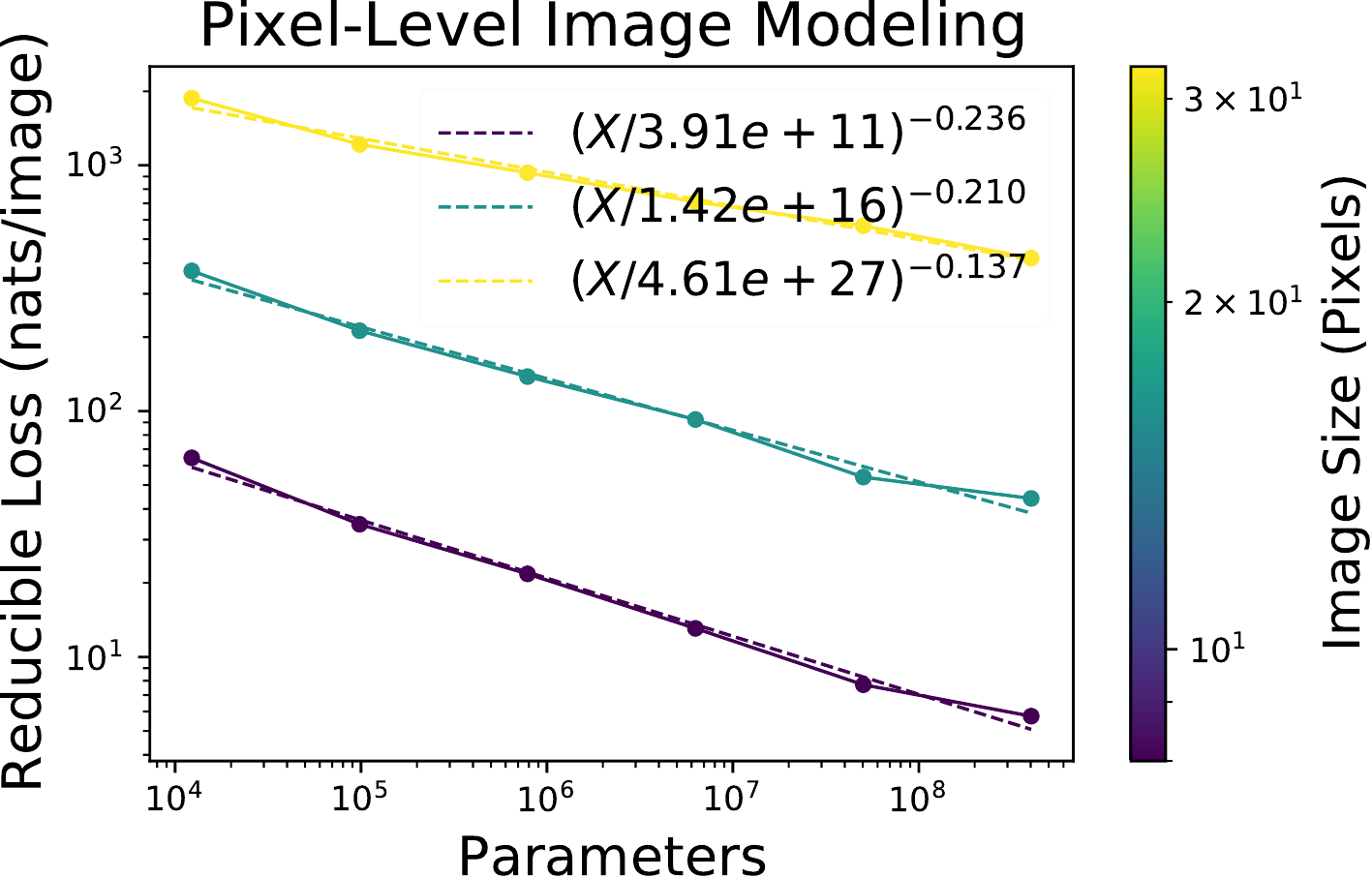}\hfill
\includegraphics[width=0.45\textwidth]{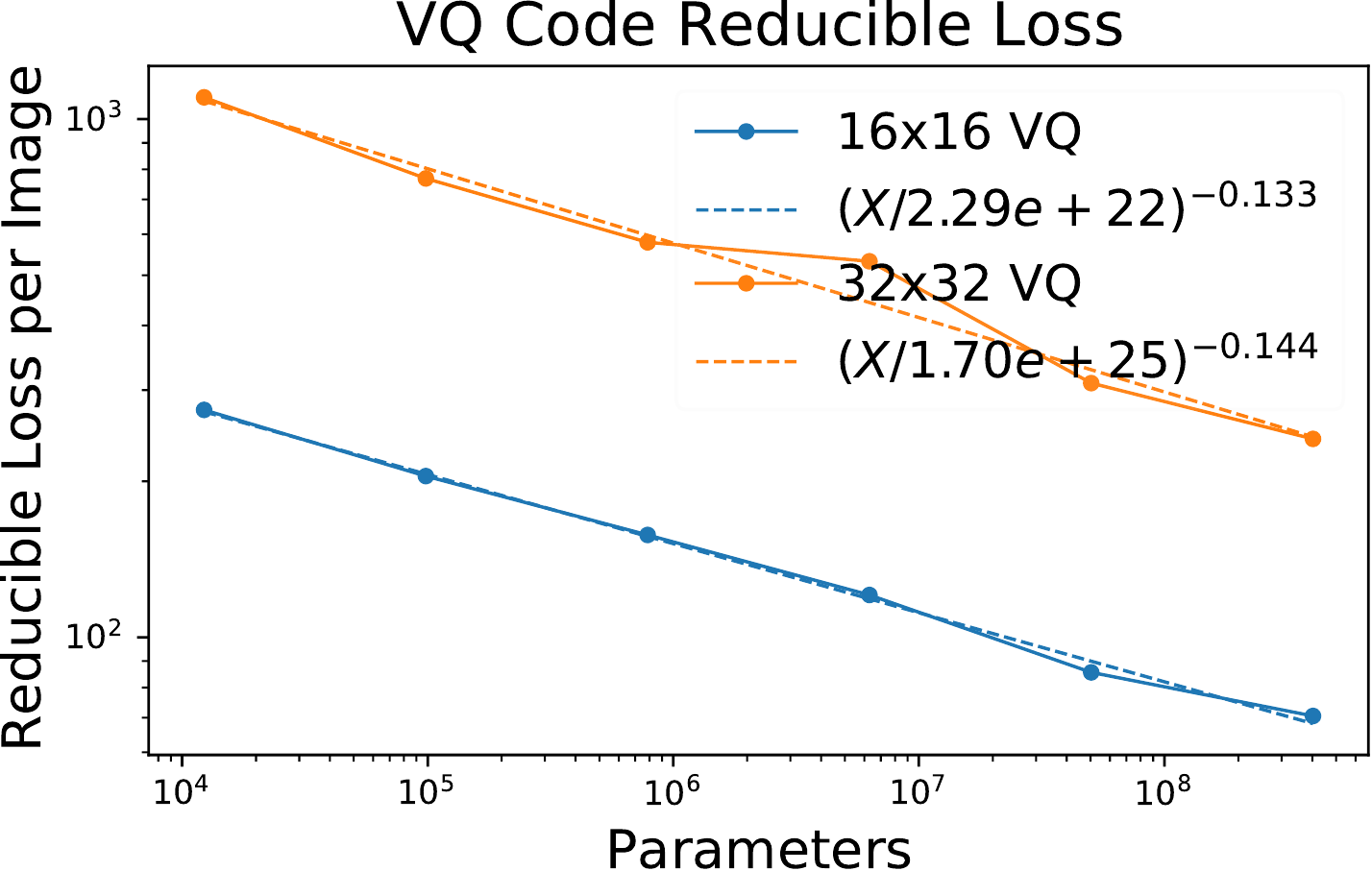}
\caption[Image Resolution Model Size Scaling]{\textbf{Comparison of image resolutions (model size scaling)---} {\bf Top:} We display scaling laws with model size for various image resolutions, and also with various VQ encodings, along with power-law plus constant fits (dashed) to equation (\ref{eq:PowerLawPlusConstant}).  The fits for pixel-level image modeling are shown in table \ref{tab:ImageResolutionTrends}.  Note that the tiniest (10k non-embedding parameter) pixel models underperform at higher resolutions; we suspect they have difficulty recognizing relative positions in larger images.  These deficiencies are even more clearly visible in the compute trends. \label{fig:ModelSizeScalingImageResolution} \label{fig:ReducibleLossImagesbyResolution}
{\bf Bottom:} We show the reducible losses, which estimate the KL divergence between the true probability distribution over images and the distribution predicted by our models.  We show the result as a function of model size and image resolution or encoding, along with pure power-law trends. 
}
\end{figure}

\begin{figure}
\noindent \centering{} 
\includegraphics[width=0.32\textwidth]{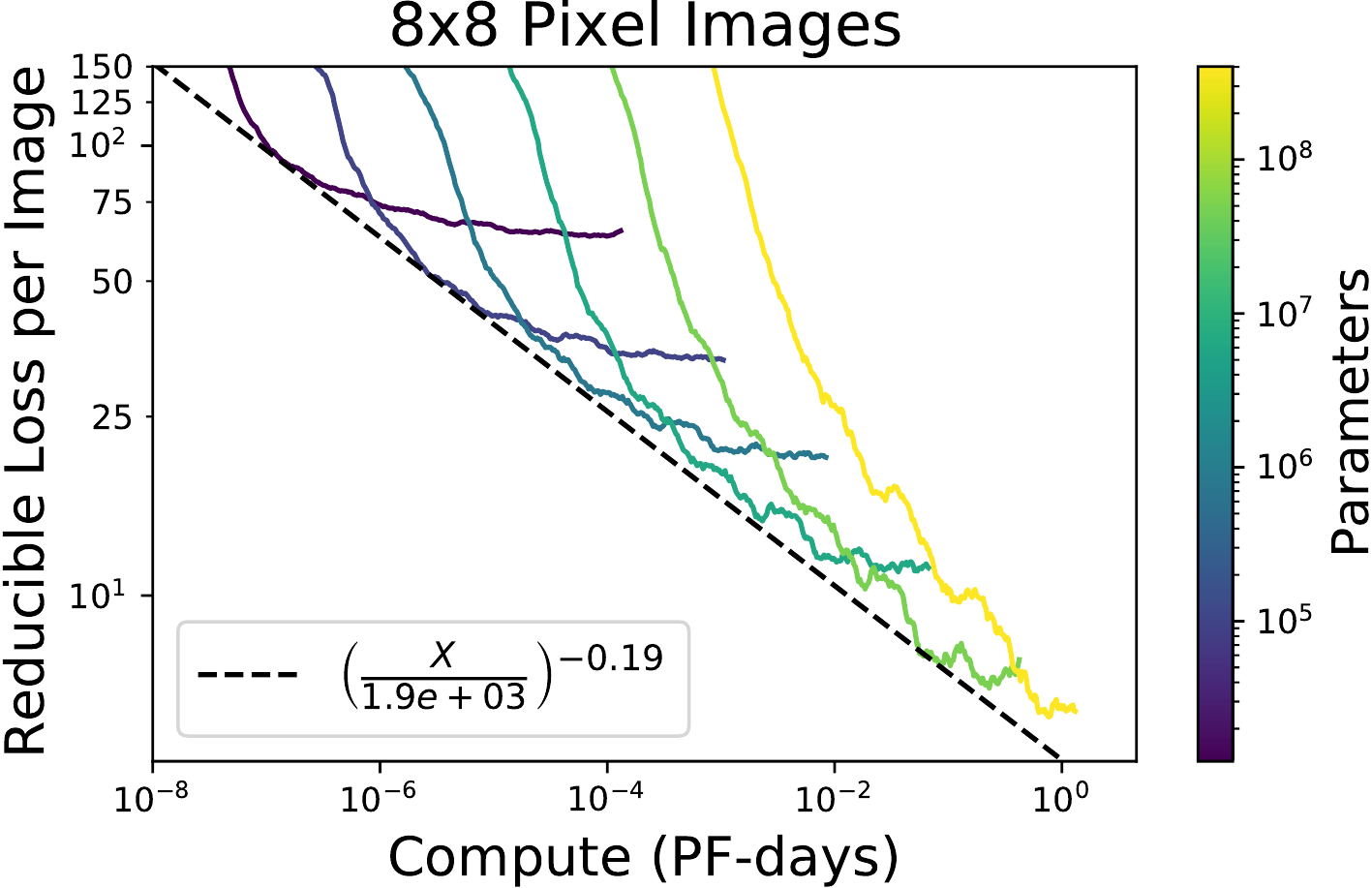}\hfill
\includegraphics[width=0.32\textwidth]{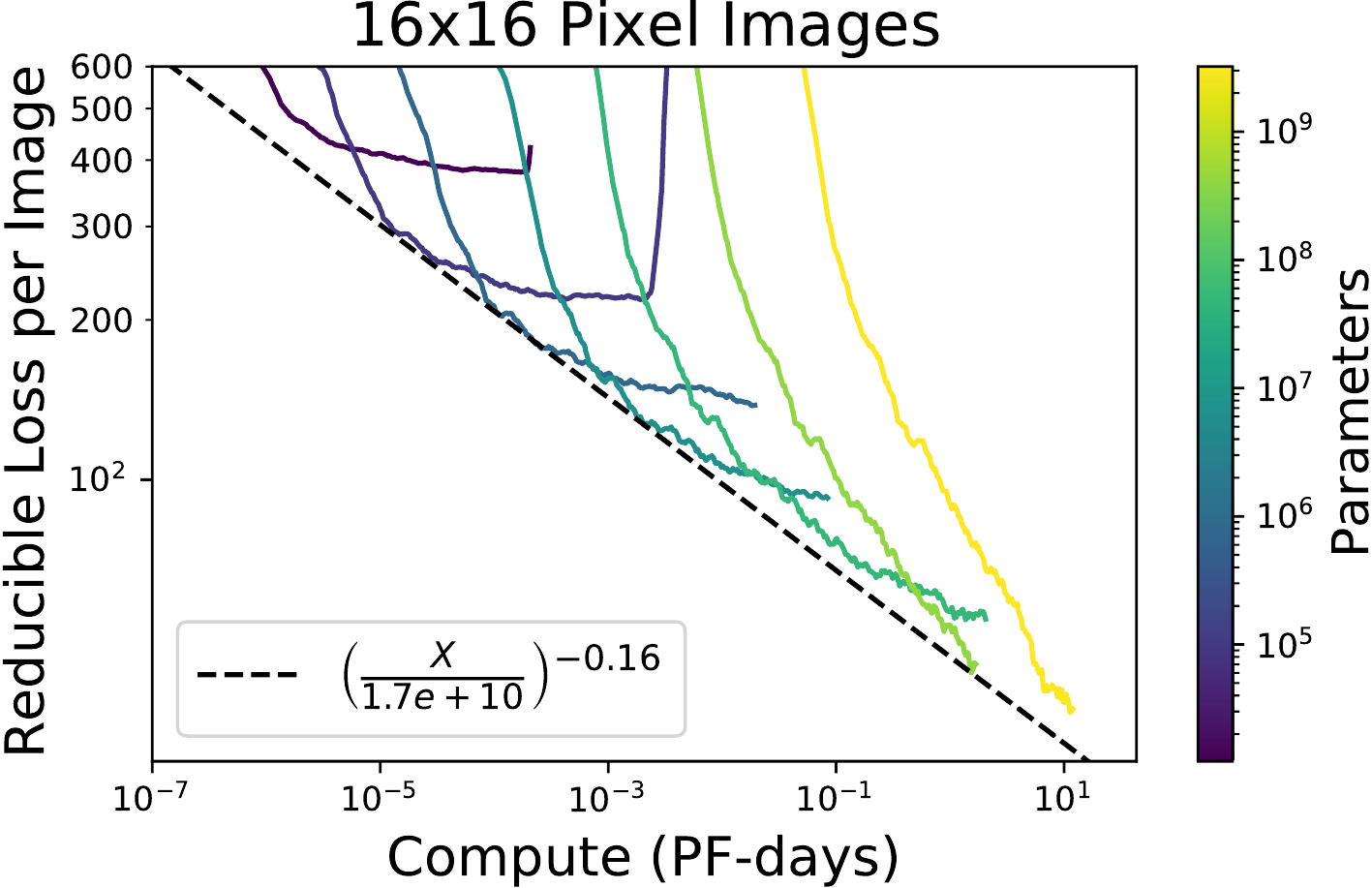}\hfill
\includegraphics[width=0.32\textwidth]{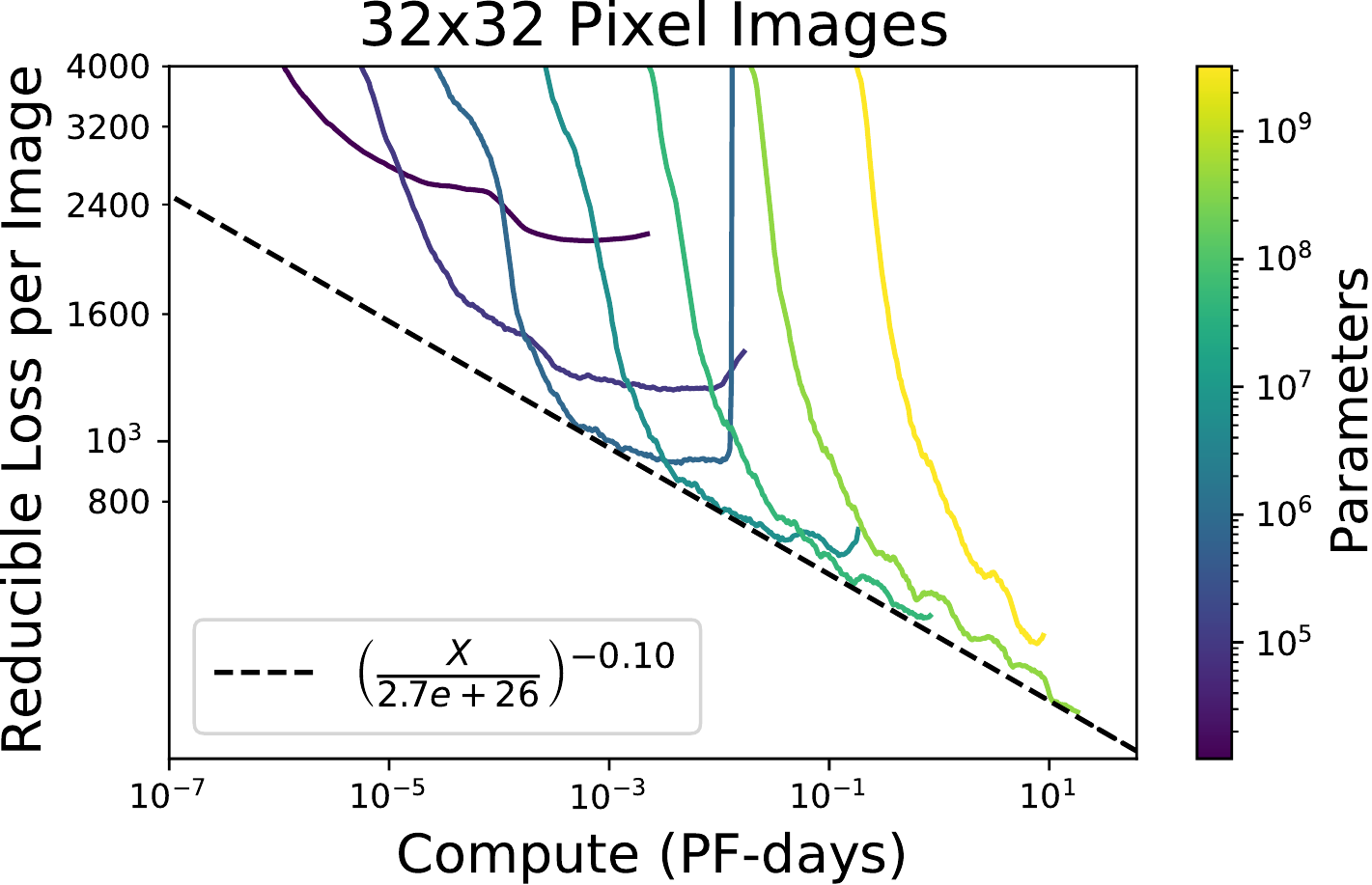} \\\vspace{1em}
\includegraphics[width=0.32\textwidth]{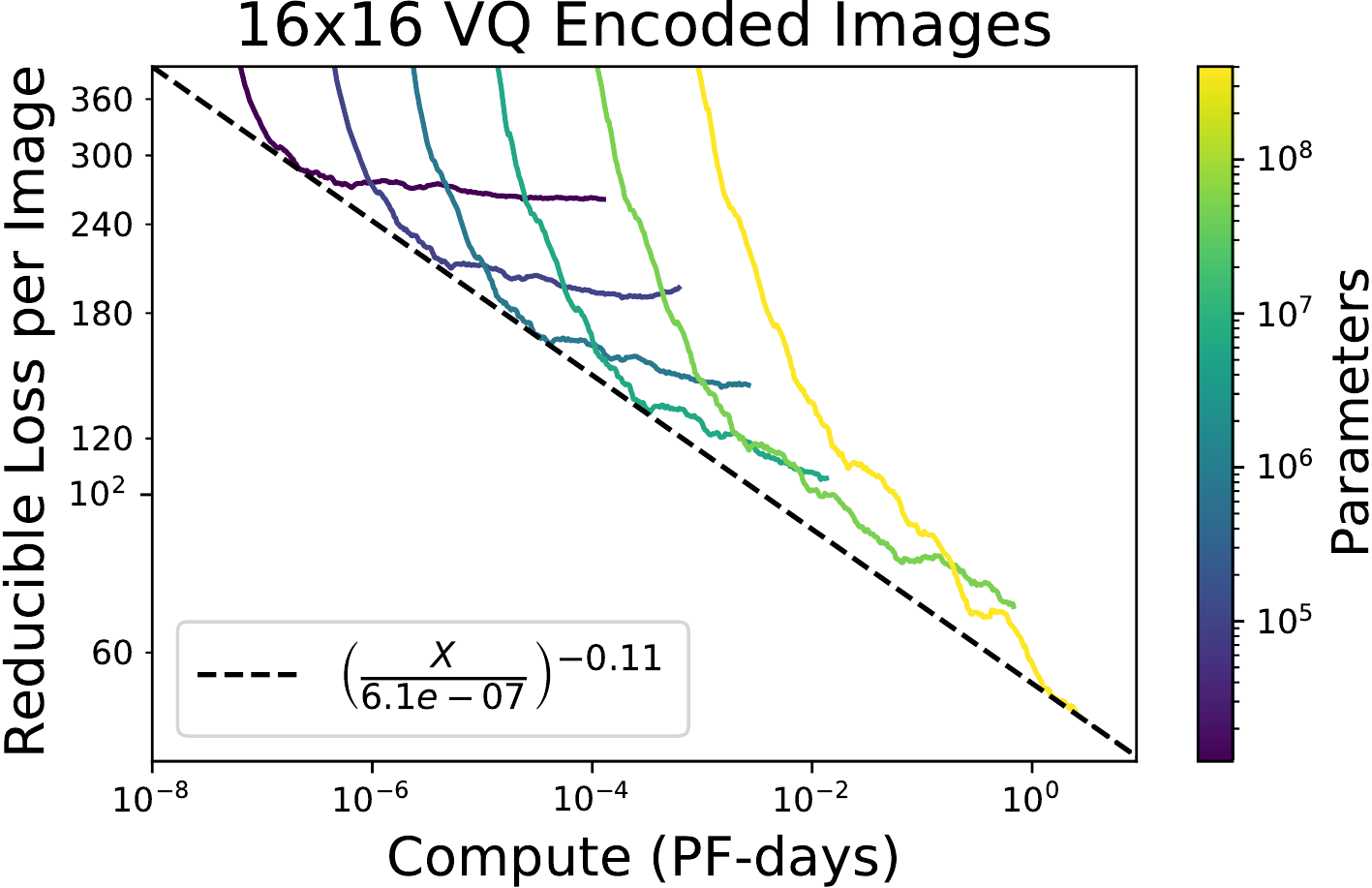}\hspace{2em}
\includegraphics[width=0.32\textwidth]{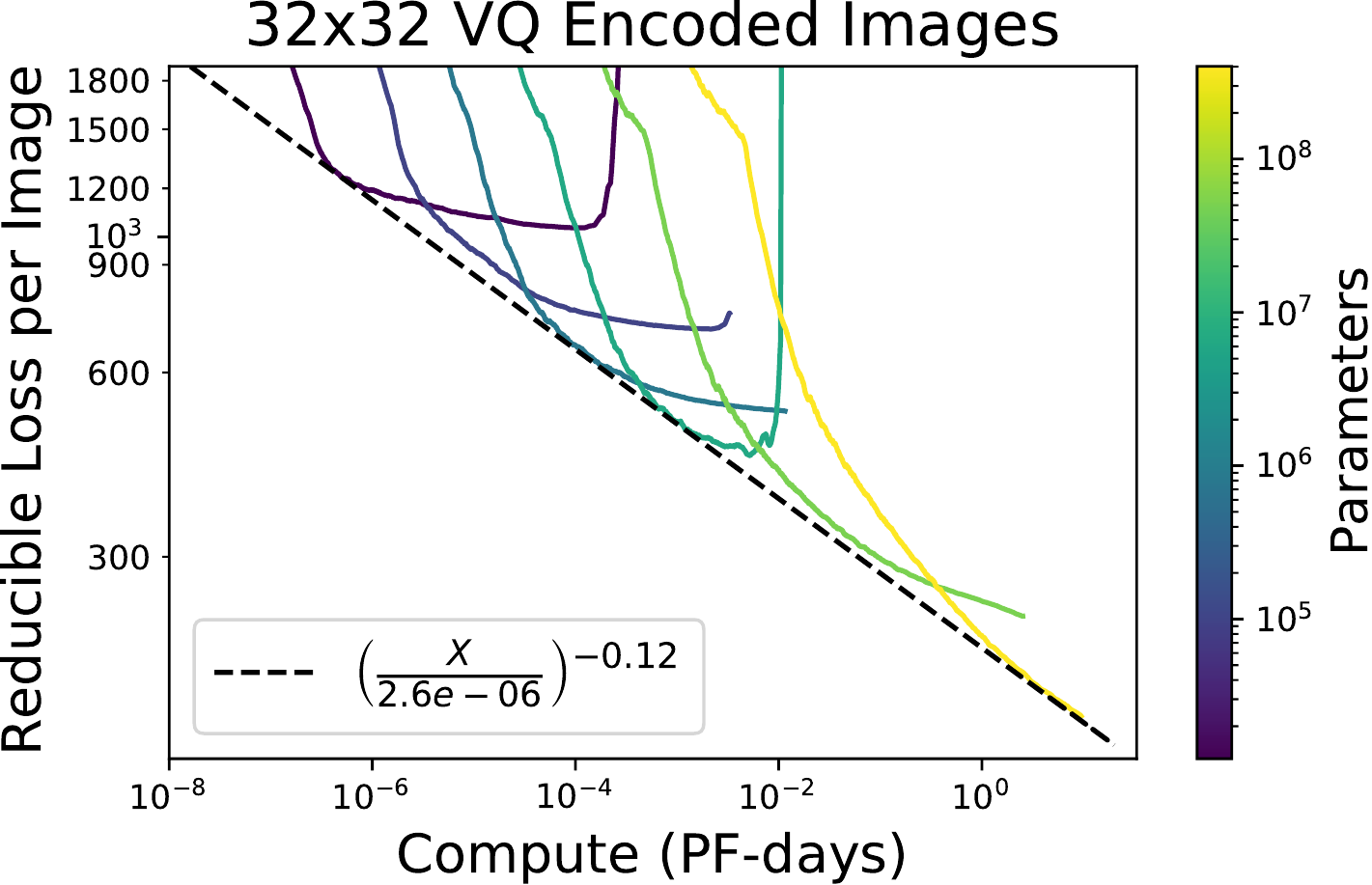}
\caption[Image Resolution Compute Scaling]{\textbf{Comparison of image resolutions (compute scaling)---} We display scaling of the reducible loss with compute for pixel-level image modeling at various resolutions (first line), and for various VQ encodings of 64x64 images (second line).  We show the test loss, but we did not observe any  train/test gap for these models.  A few models diverged late in training.  \label{fig:ComputeScalingReducibelLossImageResolution}}
\end{figure}

We trained Transformers on the YFCC100m  dataset after scaling images down to 8x8, 16x16, and 32x32 pixel resolutions, along with 64x64 images encoded with VQ codes \cite{oord2018neural} with 16x16 and 32x32 VQ code patterns.  We display the trends for the reducible loss per image as a function of the compute budget in figure \ref{fig:ComputeScalingReducibelLossImageResolution} (see figure \ref{fig:ComputeScalingImageResolution} in the appendix for trends for the full loss).  We include these figures to emphasize that the reducible loss for an optimally-allocated compute budget follows a power-law trend, even when the reducible loss becomes very small.  

Note that the smallest models underperform as compared to the trends at resolutions greater than 8x8.  We see this both for the compute trends in figure \ref{fig:ComputeScalingReducibelLossImageResolution} as well as in model-size trends in figure \ref{fig:ModelSizeScalingImageResolution}.  We speculate that this is due to difficulty utilizing the positional encodings.  For example, our smallest models have only 10k non-embedding parameters, while 32x32 images include $3072$ tokens in their context, each with a distinct positional embedding.

\begin{table}
  \begin{center}
    \begin{tabular}{c|c|c} 
      {Resolution} & {Reducible Loss per Image} (nats) & Irreducible Loss per Image (nats) \\
      \hline
      8x8  &  $ \left(\frac{C}{ 1.9 \times 10^3 }\right)^{-0.19}$  &   $602$  \\
      16x16&   $ \left(\frac{C}{ 1.7 \times 10^{10}}\right)^{-0.16}$ & $2026$ \\
      32x32 & $\left(\frac{C}{ 2.7 \times 10^{26} }\right)^{-0.1}$  & $6806 $ \\
      64x64 (16x16 VQ) & $\left(\frac{C}{ 4.7 \times 10^{15} }\right)^{-0.11}$  & $1047 $ \\
      64x64 (32x32 VQ) & $\left(\frac{C}{ 3.1 \times 10^{19} }\right)^{-0.12}$  &  $3246$ \\
    \end{tabular}
        \caption{\textbf{Per-image loss trends---} Fits for the reducible and irreducible loss as a function of compute for various image resolutions, shown \emph{per-image} rather than per-token as in table \ref{table:AllScalingExponents}. Here compute $C$ is measured in PF-days, so the denominators estimate the amount of compute needed to achieve a reducible loss of 1 nat/image.  The irreducible losses estimate the entropy of the YFCC100M data distribution \cite{DBLP:journals/corr/ThomeeSFENPBL15}. \label{tab:ImageResolutionTrends}}
  \end{center}
\end{table}

To understand the significance of the reducible loss trends in table \ref{tab:ImageResolutionTrends}, recall that the  cross entropy loss between the true distribution $P$ and the model distribution $Q$ is 
\be
 {\mathbb E}_{x \sim P} \left[ \log \frac{1}{Q(x)} \right] =  D_{\mathrm{KL}}(P || Q)  + S(P)
\ee
The KL divergence vanishes when $P = Q$, and is otherwise strictly non-negative.  Thus we can identify the irreducible loss with $S(P)$, the constant entropy of the true distribution.  Then the reducible loss estimates the KL divergence between the true distribution and the distribution predicted by the model.  This interpretation can only make sense if in the limit of infinite data and compute, we expect the transformer to perfectly model the data distribution.  We have focused on $L(C)$ trends because the asymptotic limits of the model size trend $L(N)$ could be misleading if the models have not all been trained fully to convergence.

The power-law trends in $D_{\mathrm{KL}}$ can be extrapolated down to the level of just a few nats per image.  Models powerful enough to reach this level of performance model the distribution of images with near-perfect fidelity.  In fact we see that models with $\sim 1$B parameters nearly achieve this feat for 8x8 `images'. However, we see that for larger images we would need  enormous quantities of compute to perfectly model the true image distribution.  

The consistency of the trends among distinct image resolutions in figure \ref{fig:ReducibleLossImagesbyResolution} and the strikingly small reducible loss for the 8x8 case suggests that if we could run much larger models, we would continue to see smooth improvements at higher resolution.  It seems that compute requirements for a near-perfect model of the data distribution grow as a steep power-law or even an exponential in the image resolution.  Of course we do not expect to need a perfect model of the probability distribution of real-world images for  practical tasks.  

\subsection{Video Modeling and Individual Frames}

\begin{figure}
\noindent \centering{} 
\includegraphics[width=0.32\textwidth]{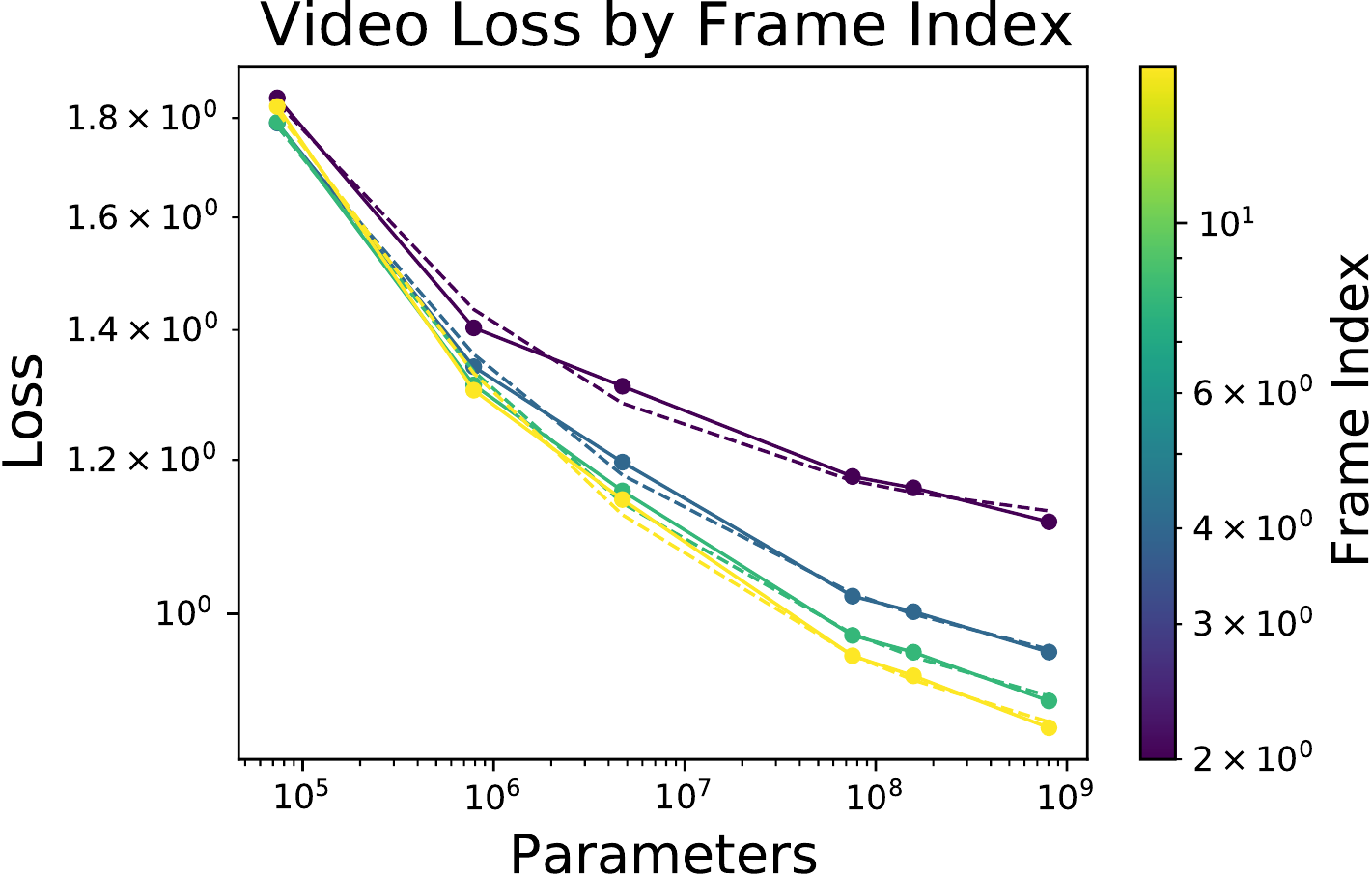}\hfill
\includegraphics[width=0.32\textwidth]{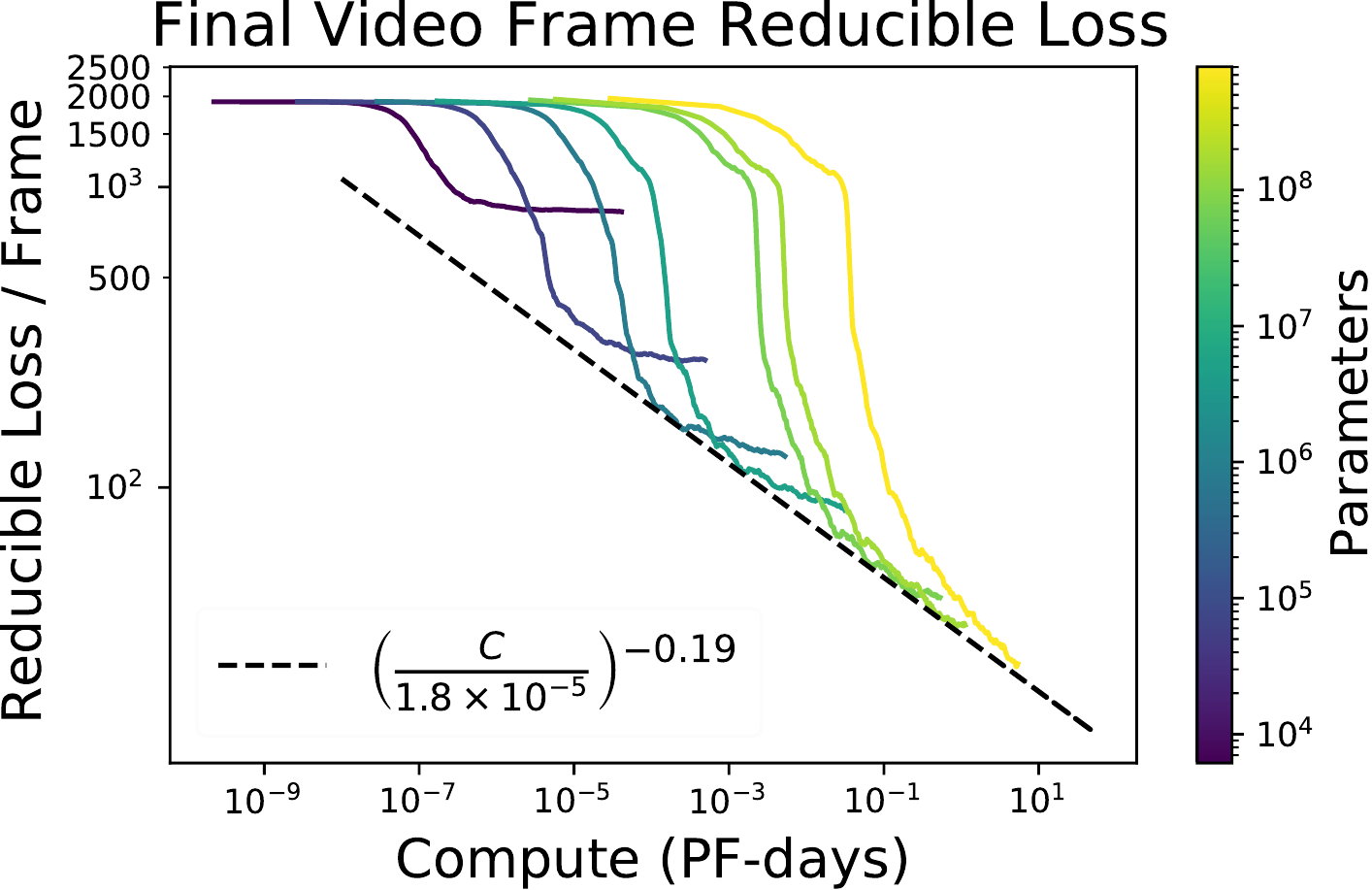}\hfill
\includegraphics[width=0.32\textwidth]{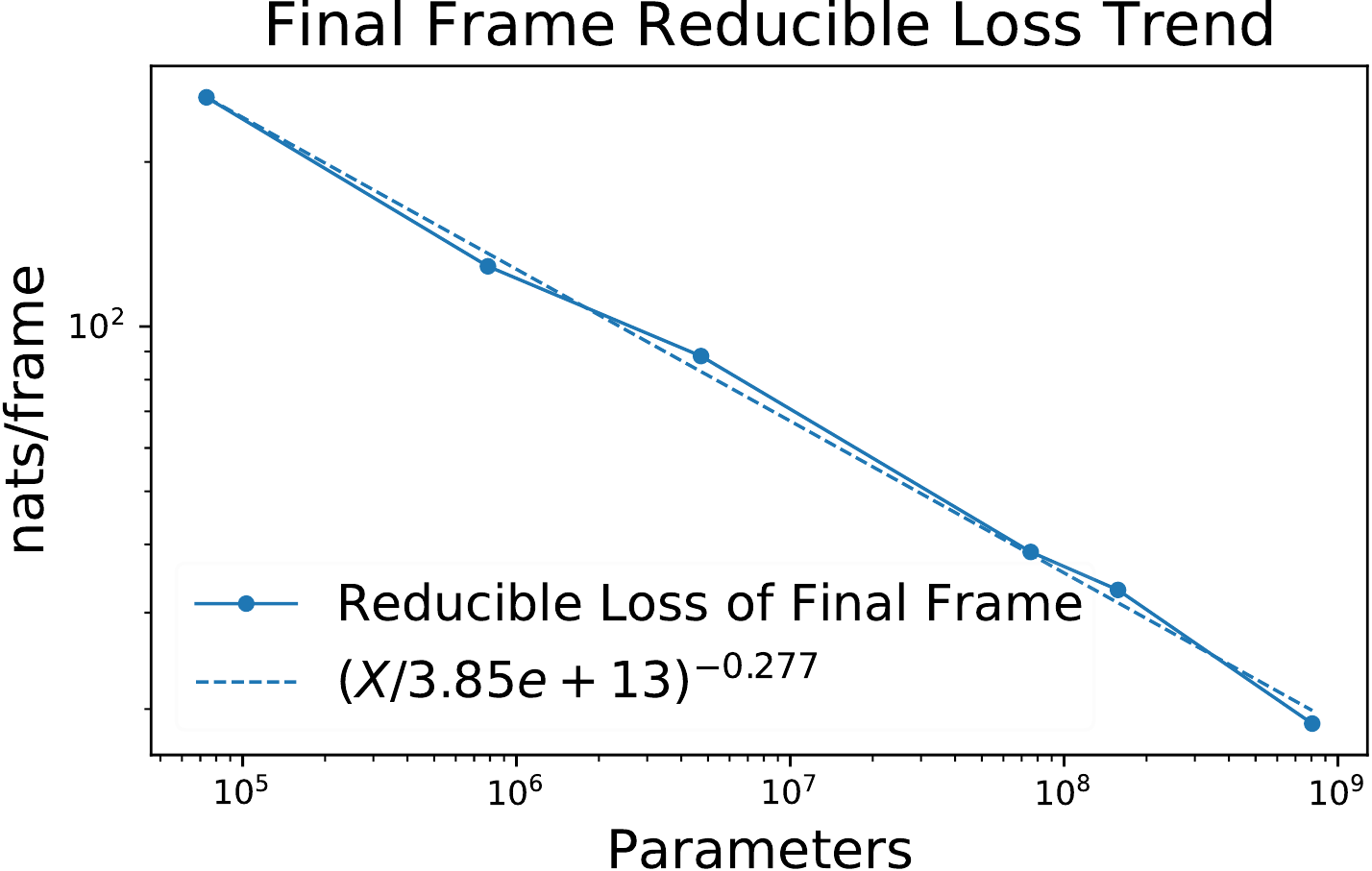}
\caption[Per Frame Video]{\textbf{Per-frame video performance trends ---} On the left we show scaling trends for specific frames in 16-frame videos.  In the center we show the reducible loss as a function of compute for the final frame of the video.  On the right we show the reducible loss and its pure power-law trend with model size for the final frame in a video. \label{fig:PerFrameVideoScaling}}
\end{figure}

For the case of video modeling, it is natural to extend the overall trends to the study of specific frames.  We display several frame-dependent results in figure \ref{fig:PerFrameVideoScaling}.  On the left we show loss as a function of model size, omitting the first frame, which has a much larger loss and should be considered an image modeling problem.  In the center we show compute scaling of the reducible loss on the final frame.  On the right in the same figure we  show the reducible loss for the final (16th) frame, which is of particular interest when generating a continuation of an existing video.  Much like the trends for image modeling, we see that the reducible loss is very well approximated by a power-law, making it possible to forecast that we would need a model size around $\sim 10^{13}$ parameters and compute of around $10^4$ PF-days to achieve a loss of just a few nats/frame on the final frame of this type of video. 

\subsection{Scaling Trends for Individual Images}

We have observed very consistent scaling trends on a variety of data modalities.  This raises a question -- does the loss achieved by different sized models on specific, individual data examples scale in the same way?  Or are the distribution-level trends an aggregate of many different trends on individual examples?

To answer these questions, we evaluated the loss of all the pixel-level 32x32 image models on a thousand randomly chosen images from the test set.  When plotting the loss as a function of model size for individual, randomly chosen examples, in essentially all cases we observe a smooth, power-law plus constant trend.  

\begin{figure}
\noindent \centering{} 
\includegraphics[width=0.45\textwidth]{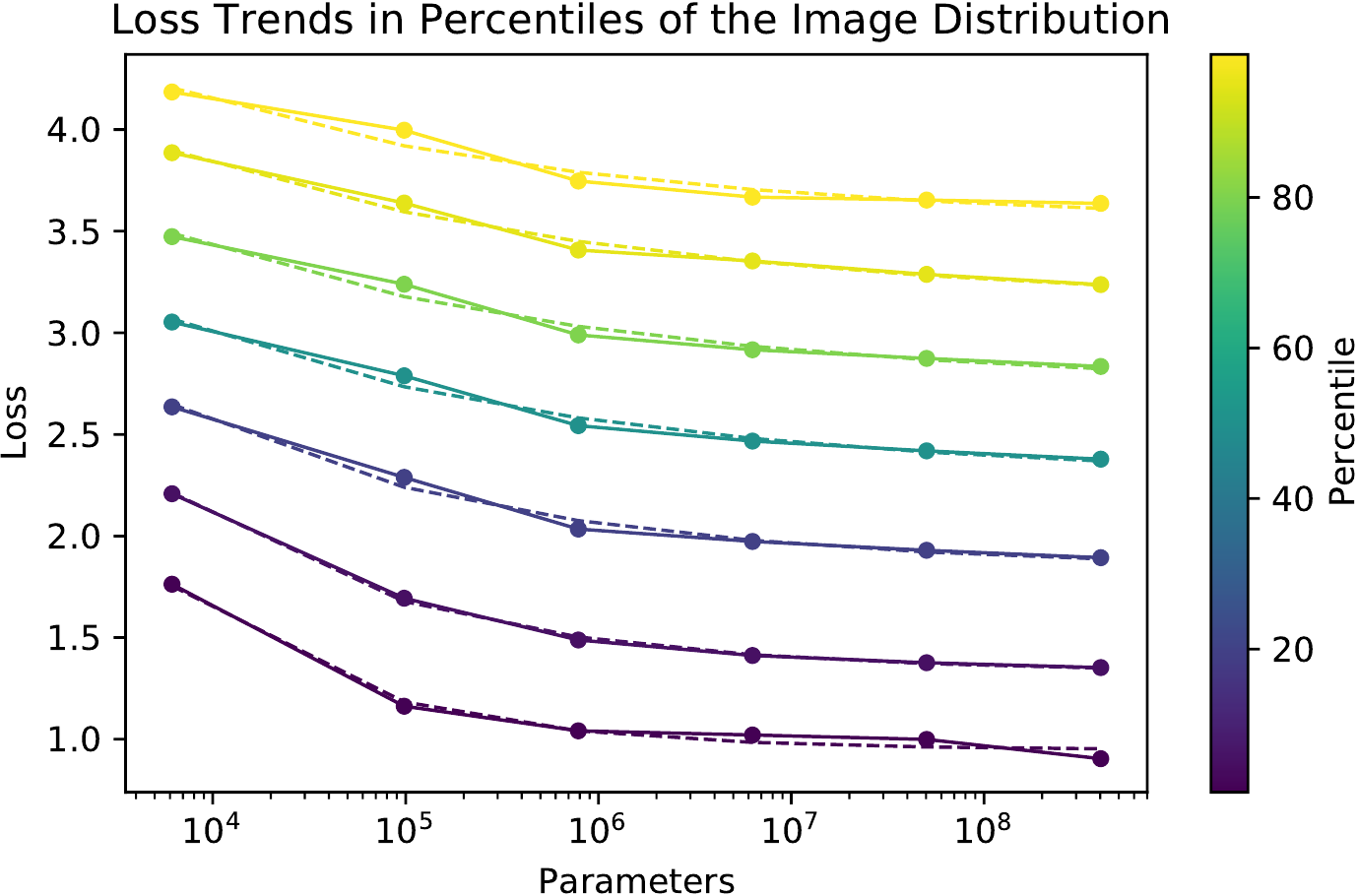}
\caption[Image Percentile Trends]{\textbf{Performance trends for image dataset percentiles---} We selected one thousand  images from the  32x32 image test set, and evaluated the loss of all models on each image.  In this figure we plot the trends in the 1, 5, 20, 50, 80, 95, 99  percentiles of the loss distribution over these images, along with power-law plus constant fits (dashed). We also observe similar trends for randomly chosen individual images (figure \ref{fig:TrendsIndividualImages}).  \label{fig:ImagePercentileTrends}}
\end{figure}

To convey this information, for each model size we evaluate the 1,5,20,50,80,95, and 99 percentile of the loss among a thousand images in the distribution, for each model size.  We then plot the trends in these percentile losses in figure \ref{fig:ImagePercentileTrends}.  We see very similar trends among all percentiles of the loss distribution, and all are well-described by equation (\ref{eq:PowerLawPlusConstant}). We  show model size trends for eight randomly chosen individual test images in figure \ref{fig:TrendsIndividualImages}. We also display the most and least improved 10 images from a sample of one thousand test images in figure \ref{fig:MostLeastImprovedImages}.  Finally, we visualize the trends in a different way, by generating conditional samples at each model size, in figure \ref{fig:CompletionsBySize}.

We would expect that these findings also apply to other data modalities.  On a quick inspection, we found the same patterns for randomly chosen text sequences and language models of different sizes.

\subsection{Finetuning on ImageNet at  32x32 Resolution}

By finetuning generative models for image classification we  gain another handle on the scaling of performance  with  model size.  We use the scaled-down 32x32 resolution ImageNet \cite{DBLP:journals/corr/ChrabaszczLH17} and finetune the 32x32 resolution pixel-level generative image models.  

To turn these models into classifiers, we remove their final embedding matrix and use the mean-pooled (over all pixels) activations of the transformer's final layer as the input to a new single-layer classifier.  During finetuning we backpropagate through the full transformer, and we do not freeze any of its weights.  As a comparison, we also train equivalent randomly initialized transformer models `from scratch' on only the classification task.

\begin{figure}
\noindent \centering{} 
\includegraphics[width=0.45\textwidth]{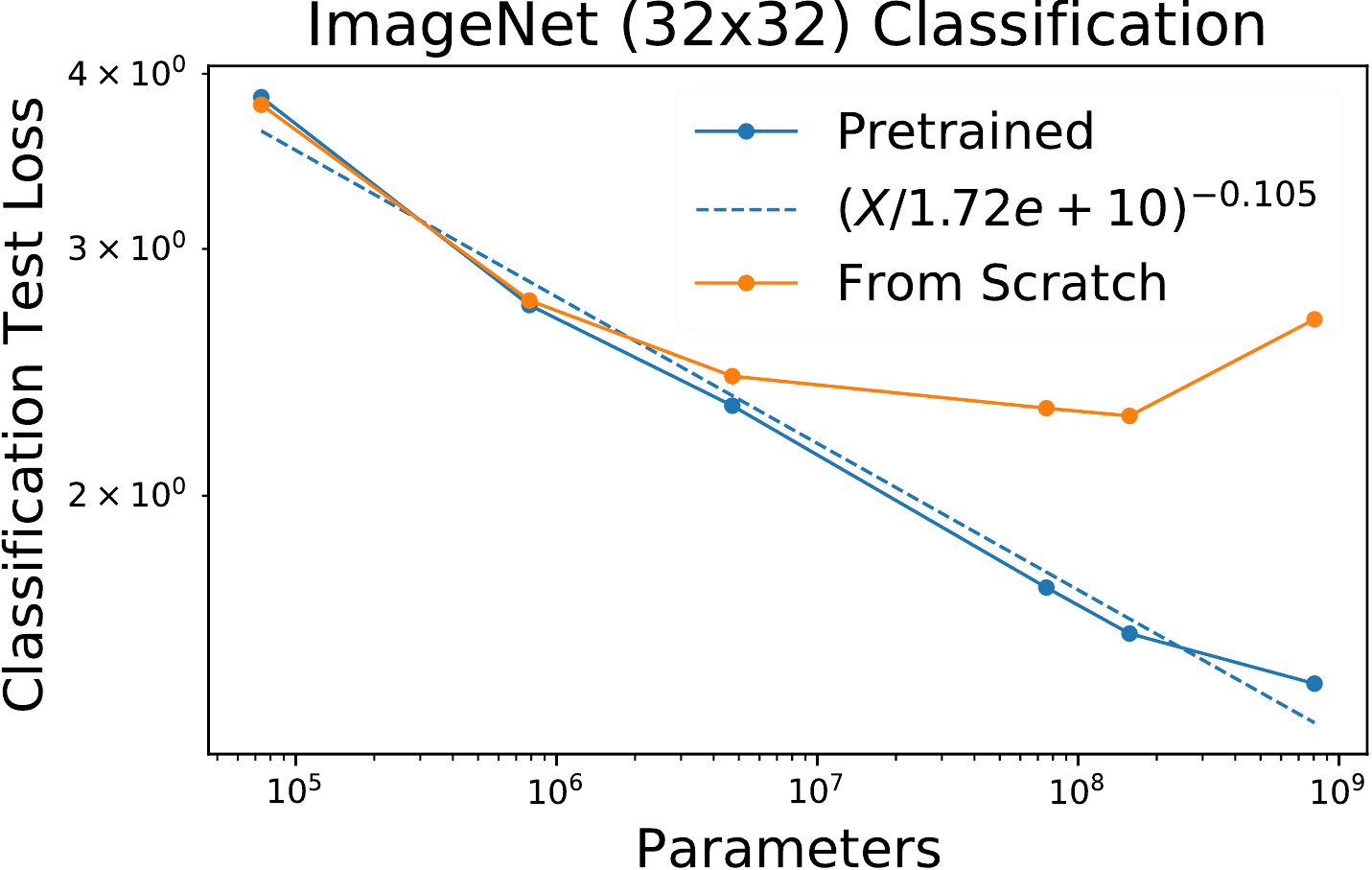}\hfill
\includegraphics[width=0.45\textwidth]{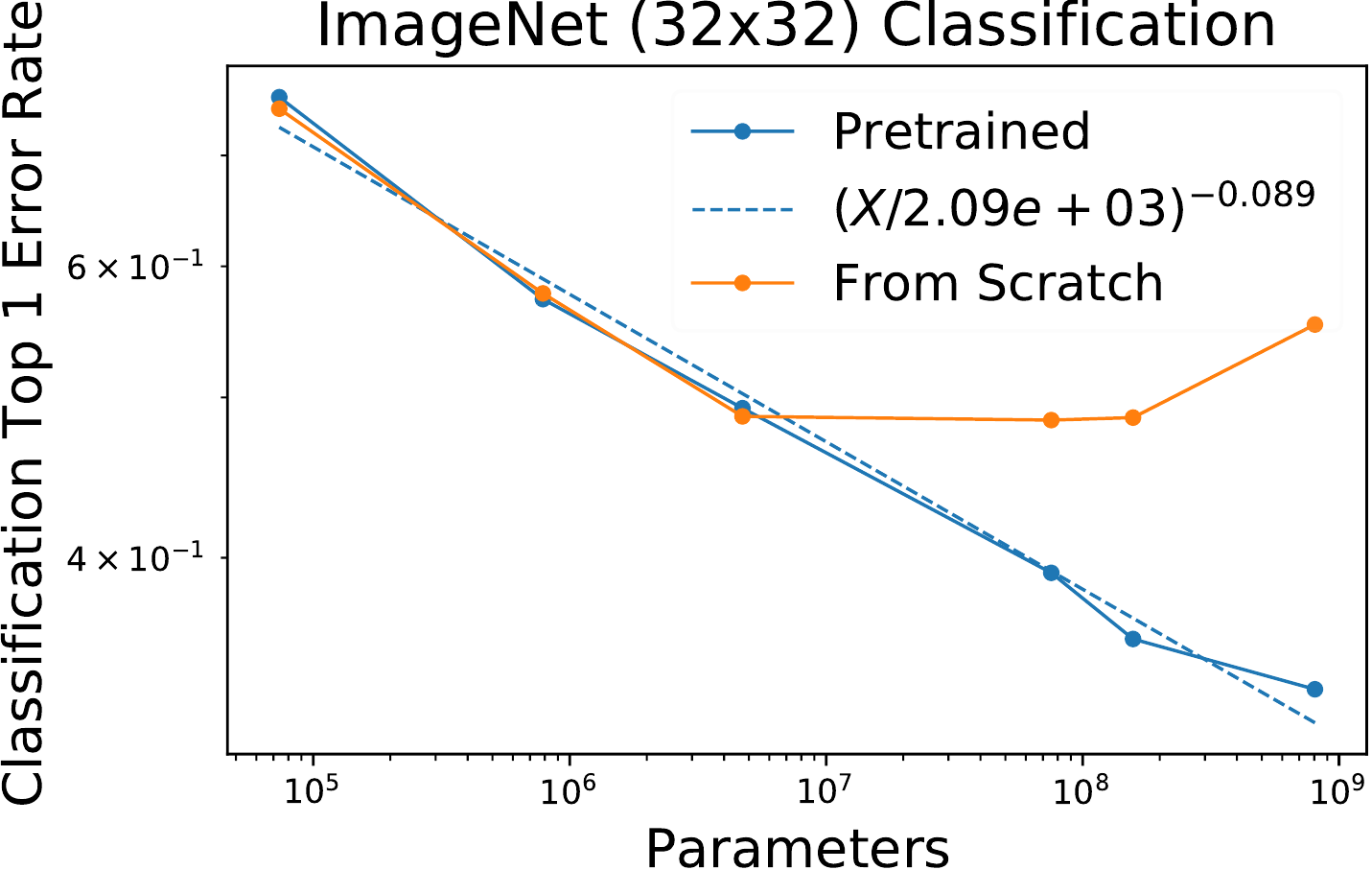}\\\vspace{1em}
\includegraphics[width=0.45\textwidth]{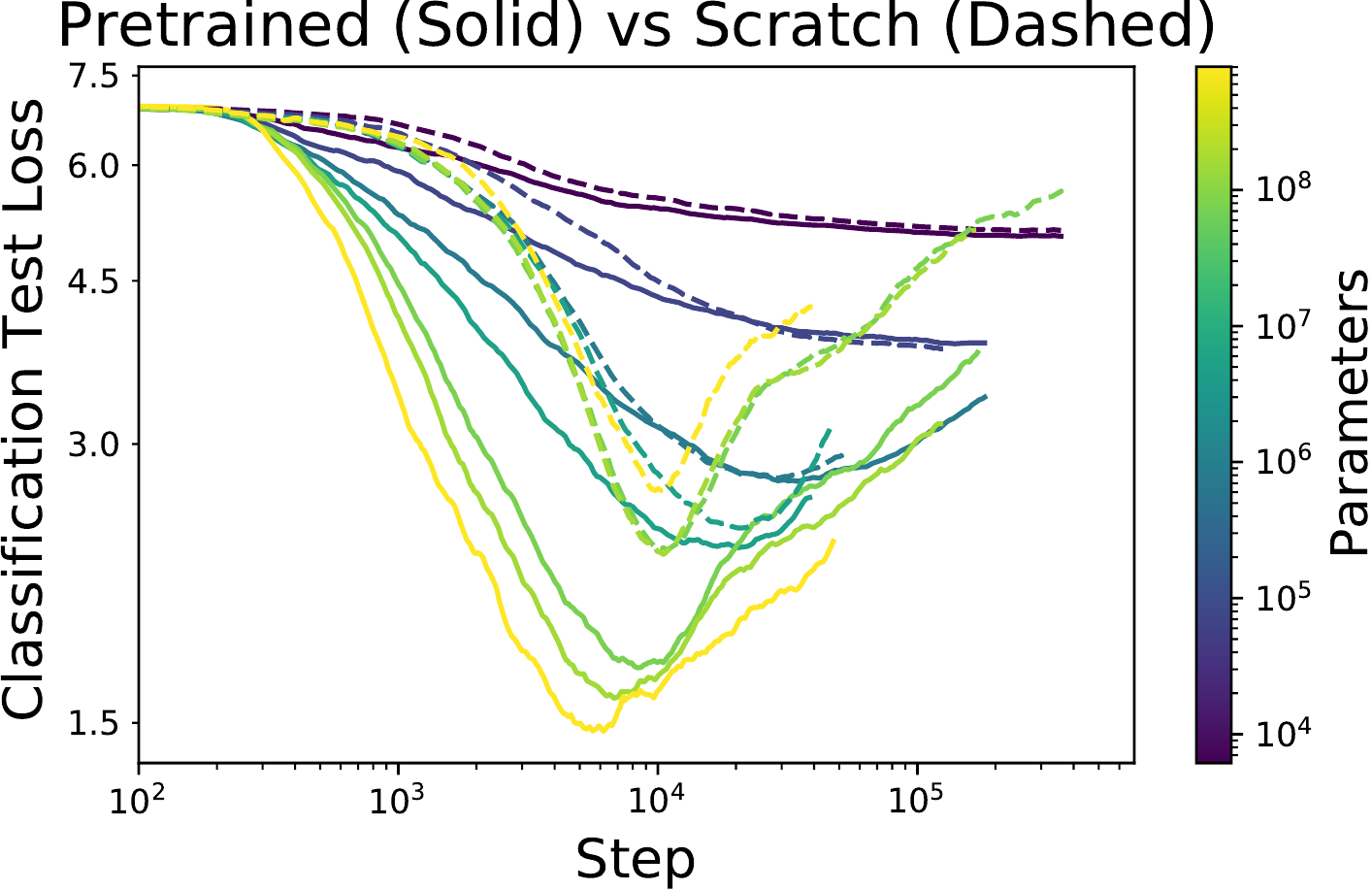}\hfill
\includegraphics[width=0.45\textwidth]{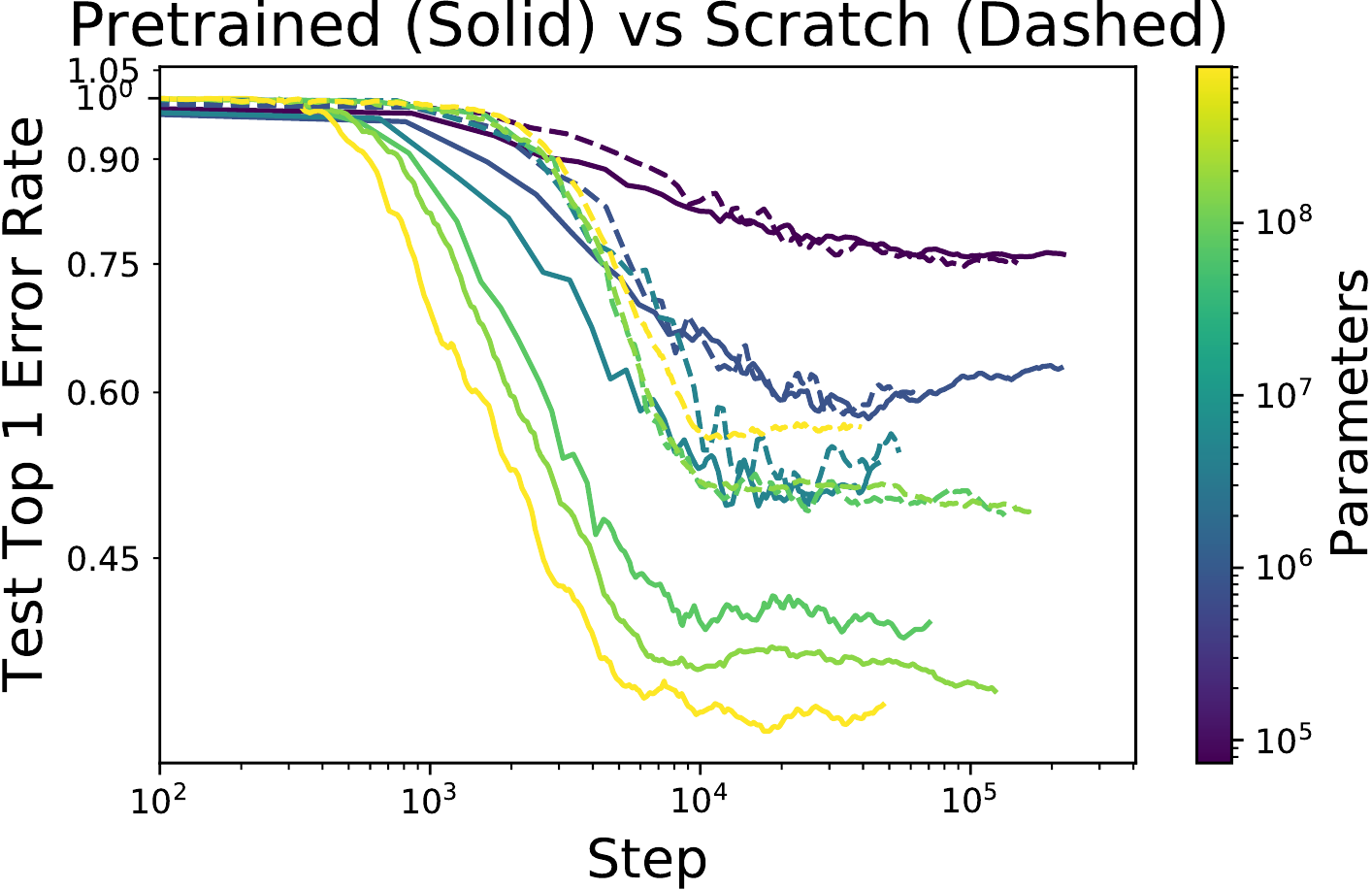}
\caption[ImageNet Trends]{\textbf{Trends in image classification performance---}
\textbf{Top:} We show model size scaling results for 32x32 pixel ImageNet \cite{DBLP:journals/corr/ChrabaszczLH17} classification.  We compare models trained from scratch on ImageNet classification (ie with no pre-training) to finetuned generative models.  Though the generative loss trend bends as it approaches the irreducible loss (figure \ref{fig:ModelSizeScalingImageResolution}), the pretrained models exhibit a straight power-law trend in classification performance vs model size, which also continues far beyond the point where the models that were trained from scratch exhibit overfitting.  
\textbf{Bottom:} Larger pre-trained models fine-tune significantly faster, and to significantly better performance, despite the approach to the irreducible generative loss.  The same does not hold when training from scratch.
\label{fig:ImageNetClassificationTrends}}
\end{figure}

Finetuning learning  curves for both pretrained and randomly initialized models are available in figure \ref{fig:ImageNetClassificationTrends}.  In all cases we use a batch size of $1024$ images, and we use the same learning rate schedule for finetuning as was used for pretraining.   We see that for small models, pretraining affords almost no benefit compared to training from scratch, but it greatly enhances the performance of larger models.

More importantly, in figure \ref{fig:ImageNetClassificationTrends} we show the model-size trends of ImageNet classification performance for pretrained and randomly initialized models.  We see that the pre-trained models follow a smooth, pure power-law\footnote{We have not encountered a clear irreducible loss in the range of model sizes that we have explored.} trend in both loss as well as error rate ($1-$ accuracy).  The very existence of these trends on a downstream finetuning task provides a striking confirmation of the importance of neural scaling laws for AI capabilities.  In the case of language, GPT-3 \cite{brown2020language} provides many more examples.   

We also emphasize that the proximity to the irreducible loss does not necessarily indicate diminishing returns with regards to model performance.  The trends in figure \ref{fig:ImageNetClassificationTrends} continue smoothly, even though  the green curve corresponding to 32x32 resolution in figure \ref{fig:ModelSizeScalingImageResolution} suggests a close approach to the irreducible loss for models with $> 10^7$ parameters.  Apparently, a great deal of important semantic information lies in the `last few bits' near the irreducible loss.
We may also interpret this as the pre-training process  providing  a highly effective regularizer for downstream tasks.

\section{Multimodal Models and Information Gain}
\label{sec:Multimodal}

Is a picture worth a thousand words?  With multimodal models we can study the amount of information that one domain provides about another.  
For this purpose we study the empirical mutual information between images and text and the infogain defined in equation (\ref{eq:InfogainDefinition}). The latter has the interesting property that it must lie in the interval $[0,1]$, with larger values suggestive of better performing multimodal models.  

To estimate the empirical mutual information between the image and text for text-to-image models, we subtract the captioned-image loss  from the  image loss in the presence of a blank caption.  Similarly, we subtract text losses with and without corresponding images for image-to-text models. 

However, these measurements have a potentially serious flaw -- if the models have only been trained on multimodal data, then blank captions and blank images may be out of distribution.  We minimize this issue by measuring the mutual information only after finetuning our models for $10^4$ steps on an even mixture of data with and without captions (for text-to-image) or with and without images (for image-to-text).  Empirically we find that without this finetuning, the mutual information is measured to be about twice as large.  In the case of text-to-image models, we also tried training from scratch on a 95/5 mixture of mulitmodal and blank caption data, and found very similar results.  The learning curves for the mutual information and some other comparisons can be found in appendix \ref{app:MoreMultimodal}.

\begin{figure}
\noindent \centering{} 
\includegraphics[width=0.4\textwidth]{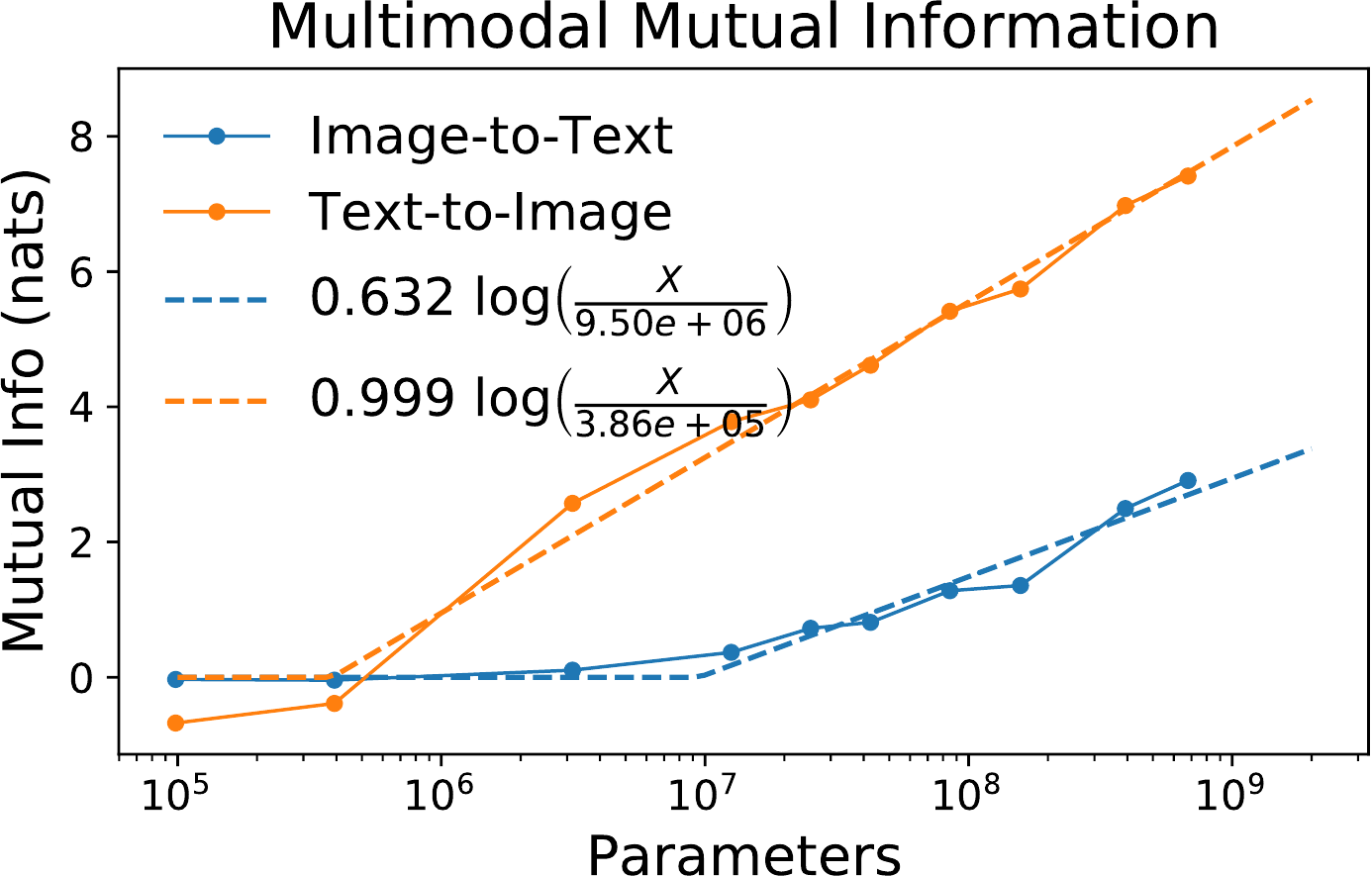}\hspace{2em}
\includegraphics[width=0.4\textwidth]{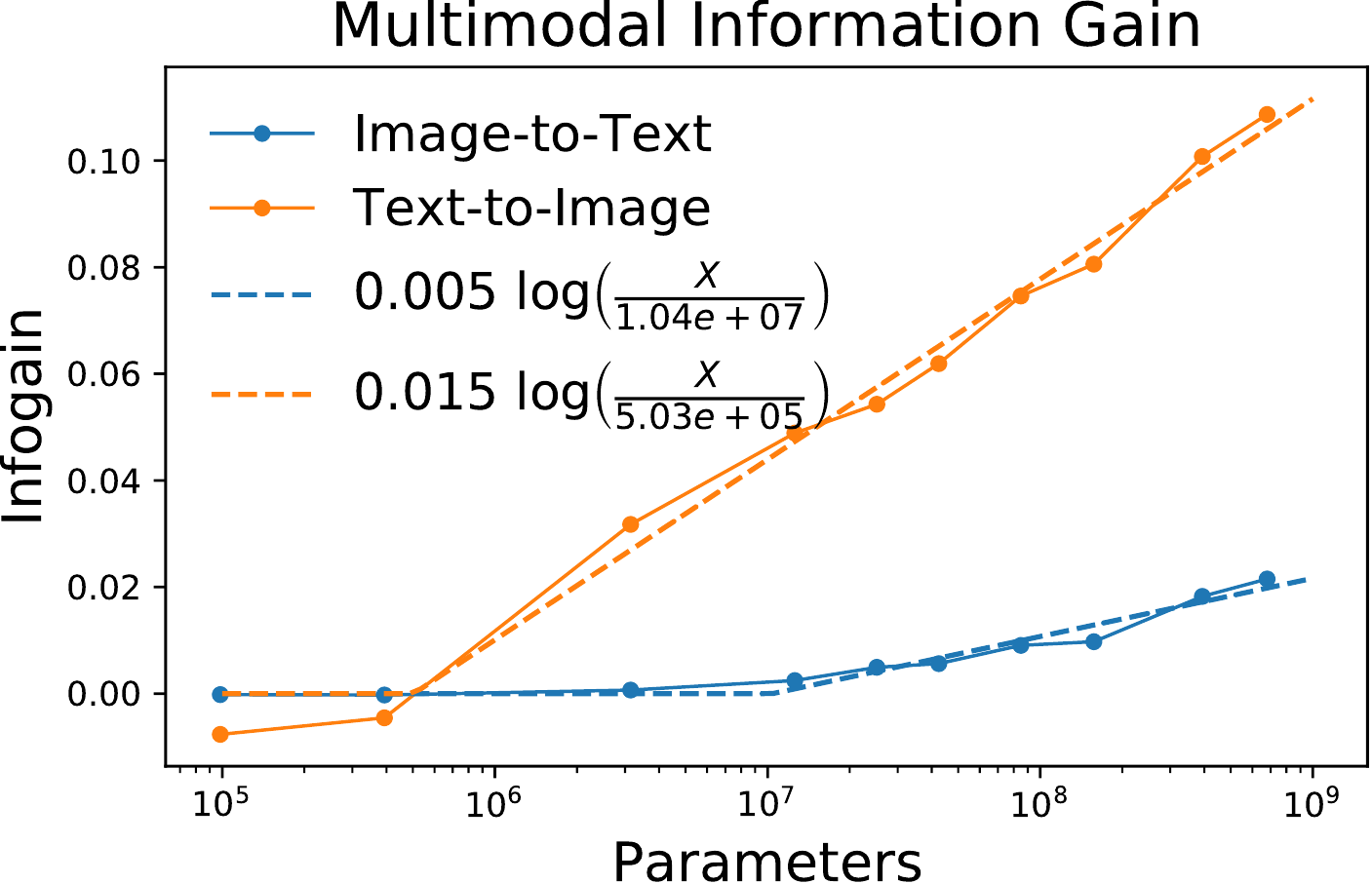}
\caption[InfoGain Trends]{\textbf{Mutual information trends for multimodal models---}
We show the empirical mutual information between image and text in multimodal models (left) and the Infogain (right), which is  the ratio of the empirical mutual information to the empirical entropy of the text.   The results in these plots were compiled after finetuning multimodal models for 10k steps on half multimodal, half blanked caption/image data, to ensure that blank captions/images were not out of distribution.  The largest text-to-image models use about 10\% of the information in the text when constructing images.  \label{fig:InfoGain}}
\end{figure}

We plot the mutual information and the infogain ratio in figure \ref{fig:InfoGain}. We see that billion-parameter, decoder-only transformer  models extract about 8 nats of information concerning the image from an average text caption in the test set.
In the case of both Image-to-Text and Text-to-Image multimodal models, we observe empirically that mutual information and infogain varies with model size as
\be
\label{eq:ScalingInfo}
I({\rm text}, {\rm image}), \  {\rm Infogain} \approx \lambda \log \left( \frac{N}{N_c} \right)
\ee
with different $\lambda$ and $N_c$ for the two cases.
We can derive this approximate formula from  plausible assumptions, as discussed in appendix \ref{app:MutualInfoInfoGainScaling}.
If this trend holds over a large range of $N$, it might be used in combination with the upper bound infogain $< 1$ to roughly estimate the maximal productive model size.  

However, the trends identified in figure \ref{fig:InfoGain} suggest a very slow growth of infogain with $N$ for these models, so it seems unrealistic to extrapolate all the way to an infogain $=1$.  Furthermore, in the  data distribution  the text and images are not always closely correlated, as in many examples much of the text has little to do with the accompanying image.  So instead we might ask when $20\%$ of the information in the text will be used to define the image, doubling the infogain of a 1B parameter model.  For text-to-image models, this threshold will be met with models of size $N \approx 3$ trillion  parameters, though for image-to-text models this remains far out of reach.  Other architectures may improve on these results, but we conjecture that they will display similar trends with model size.

Text-to-image models have much larger mutual information and infogain, as compared to image-to-text models.  We speculate that this is due to the fact that much more processing is required to extract semantic information from images than from text.

We can now revisit the question of how many words a picture is worth.  Figure \ref{fig:ModelSizeScaling} shows the loss per text token, including padding tokens; if we exclude padding tokens, the largest image-to-text models achieve a loss of 2.6 nats per text token, or about 3.4 nats per word. Comparing the image-to-text mutual information of 8 nats, we find that a 32x32 image is worth only about 2-3 words to our best models.

\section{Mathematical Problem Solving and Extrapolation}
\label{sec:Math}

In the context of machine learning, generalization most often refers to the gap between test and training performance.   But on a conceptual level, generalization can also refer to the more ambitious possibility of extrapolation from the training distribution to a larger or more diverse distribution.  Mathematical problem solving lends itself very naturally to the study of extrapolation, because we can extend the range of numbers or operations used to create math problems, or the recursive/compositional depth \cite{hupkes2019compositionality} required for a solution.  

We studied this phenomenon in the fundamental figure \ref{fig:ModelSizeScaling}, where we evaluate problem solving performance using a variety of test sets indexed by a numerical level, which corresponds to an `entropy' used for generation  \cite{DBLP:journals/corr/abs-1904-01557}.  We observe fairly smooth power-law plus constant trends for the loss on all of these test sets, but with different exponents and offsets depending on the difficulty level.  So extrapolation performance improves with model size.

\begin{figure}
\noindent \centering{} 
\includegraphics[width=0.45\textwidth]{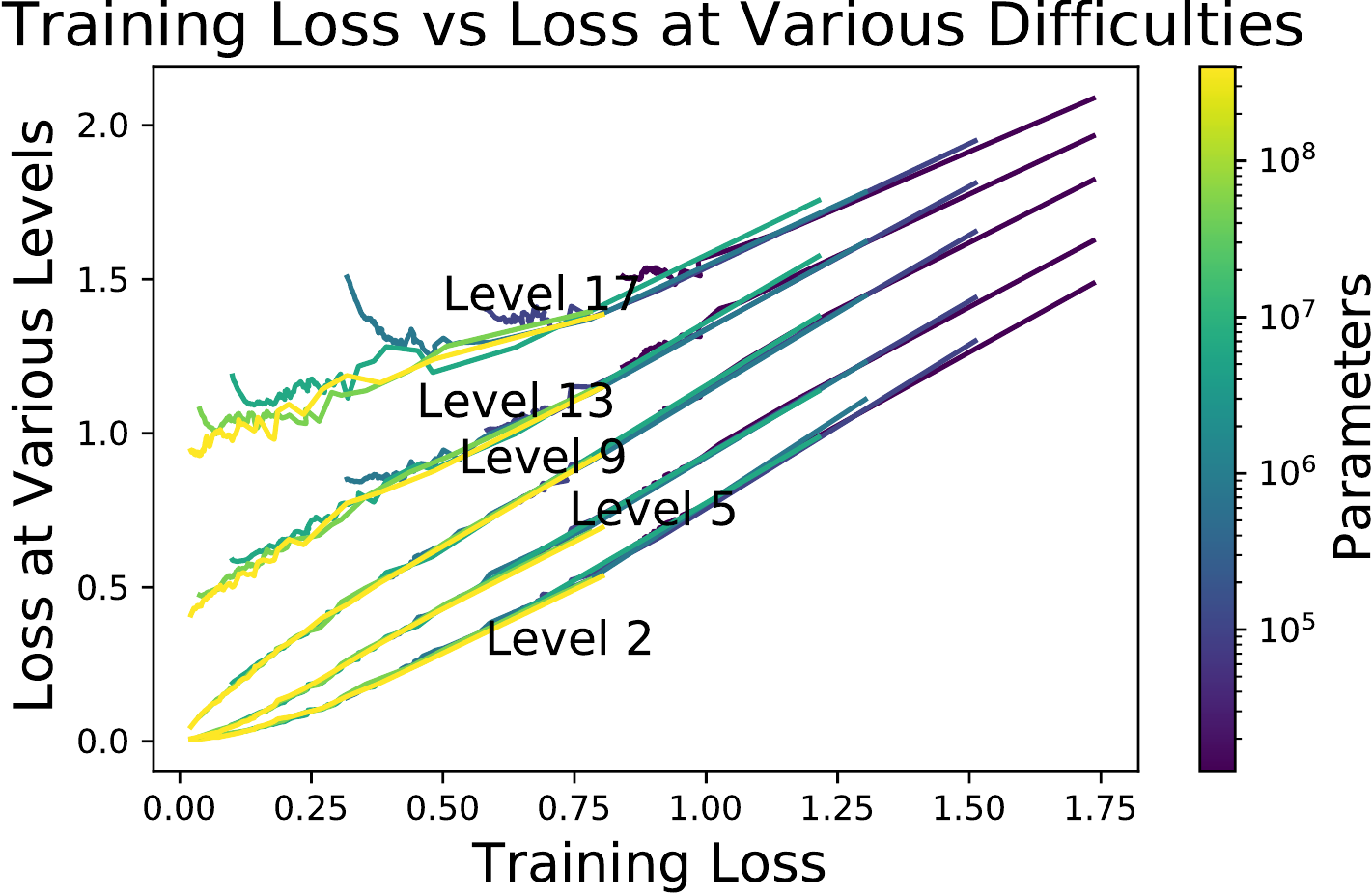}\hfill
\includegraphics[width=0.45\textwidth]{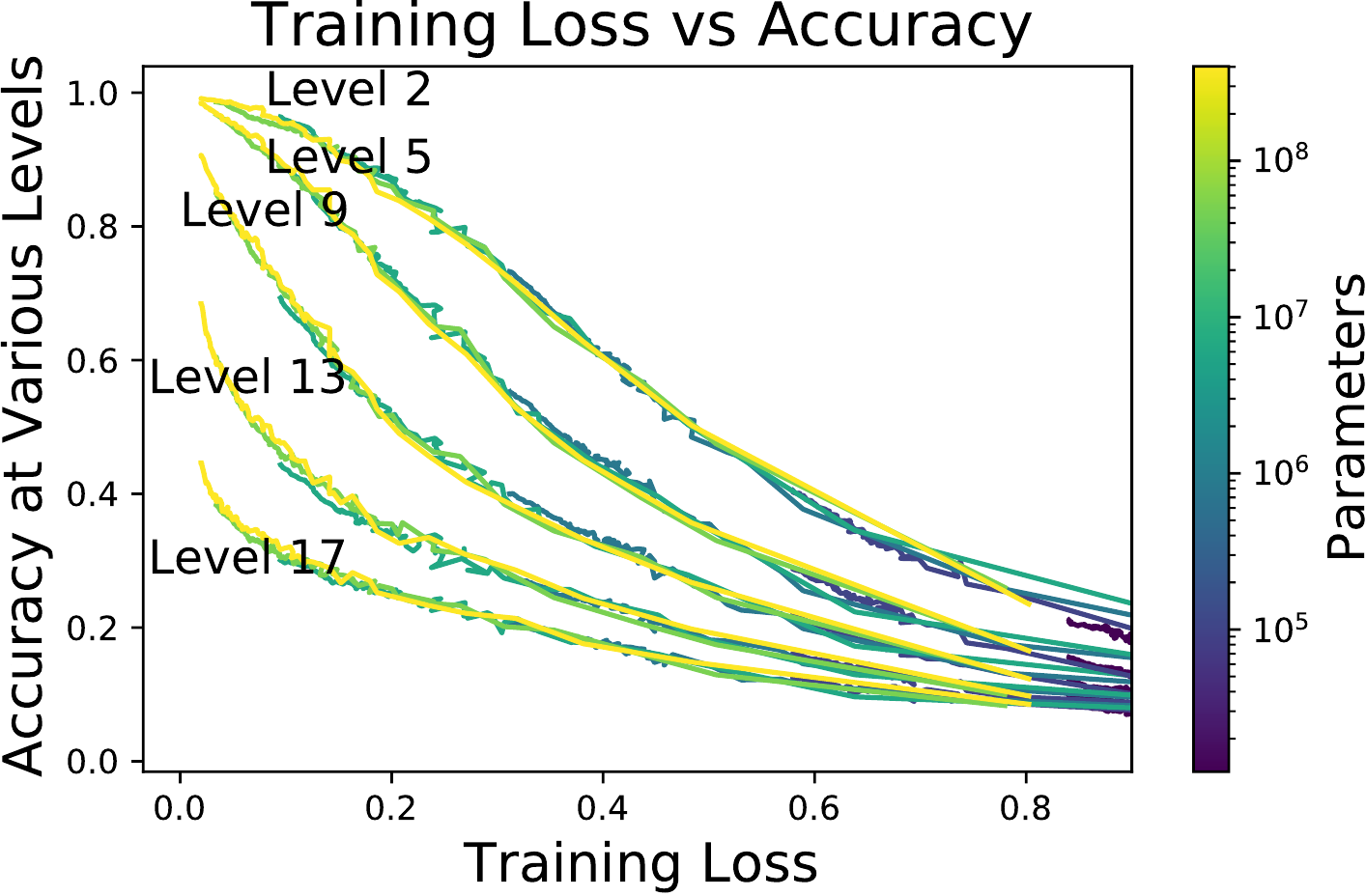}
\caption[Extrapolation]{\textbf{Mathematics difficulty levels---} We show the  loss (left) and accuracy (right) during training, as a function of the training loss, for math problems at various difficulty levels.  We emphasize that models of different size perform nearly identically when we hold the training loss fixed.  Thus in the case of math problem solving, both interpolation and extrapolation performance depends on model size primarily through the training loss.  Note the difficulties $\leq 10$ are within the training distribution; for levels  $> 10$ we expect non-zero test loss even as the training loss tends to zero.   \label{fig:Extrapolation}}
\end{figure}

However, as we show in figure \ref{fig:Extrapolation}, the extrapolative capabilities of these models predominantly  depends on the models' performance on the training distribution.  That is, models of different sizes that achieve the same loss on the training distribution perform about equally on the various test distributions.  In this sense, increasing the model size does not automatically improve extrapolation, except insofar as it improves performance on the training distribution.  Similar results were found  in \cite{kaplan2020scaling} when extrapolating from one text distribution to another.

Finally, for completeness we note that the information theoretic interpretation of the loss has a somewhat different meaning in the context of math problem solving, where the answers are deterministically related to the questions, so that the entropy should truly vanish.  For much more detailed results on math performance and a great many more trends see appendix \ref{app:MathDetails}.

\section{An Inconsistency in Compute and Datasize Scaling Laws}
\label{sec:Paradox}

\begin{figure}
\noindent \centering{} 
\includegraphics[width=0.48\textwidth]{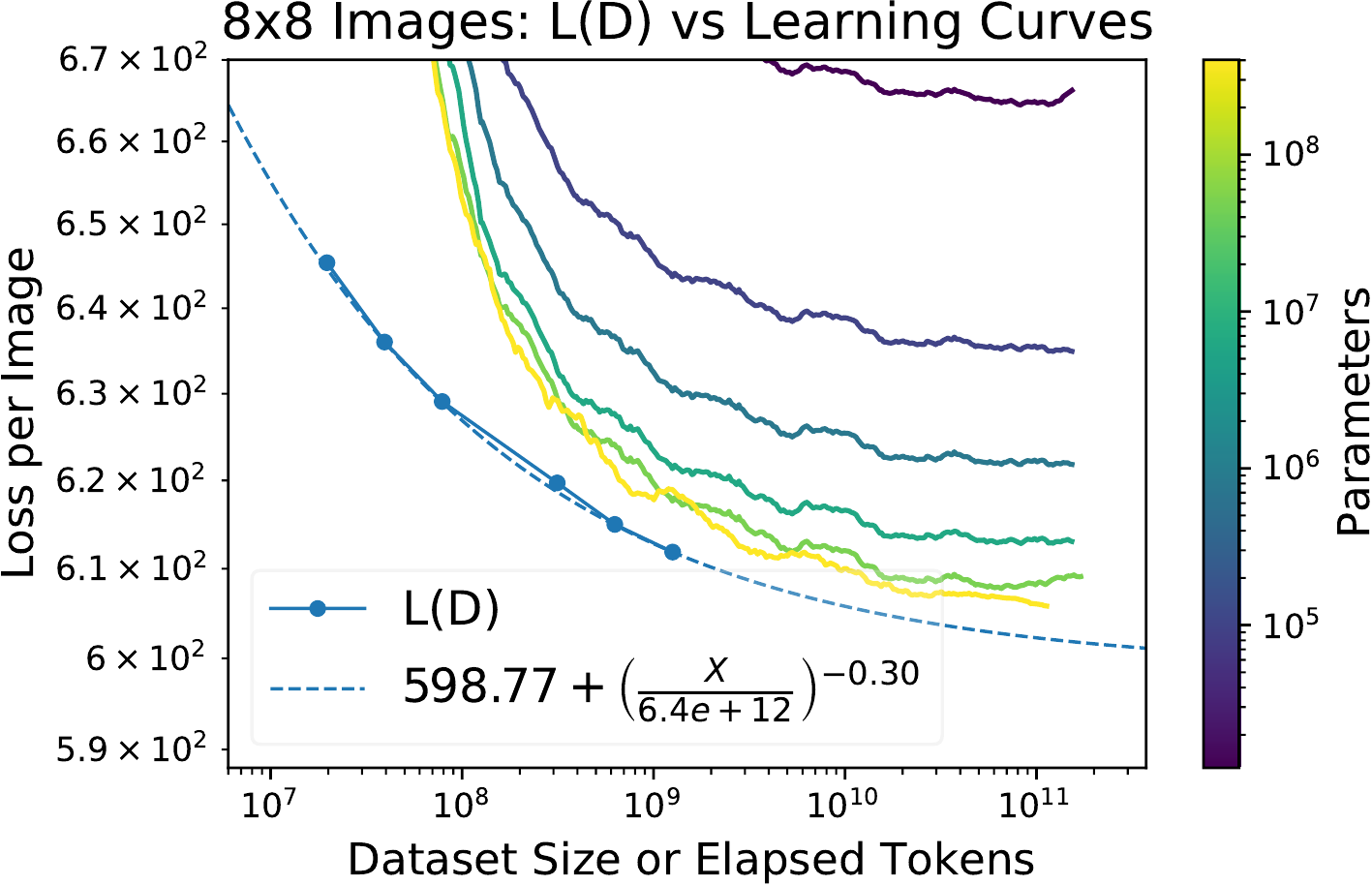}\hfill
\includegraphics[width=0.48\textwidth]{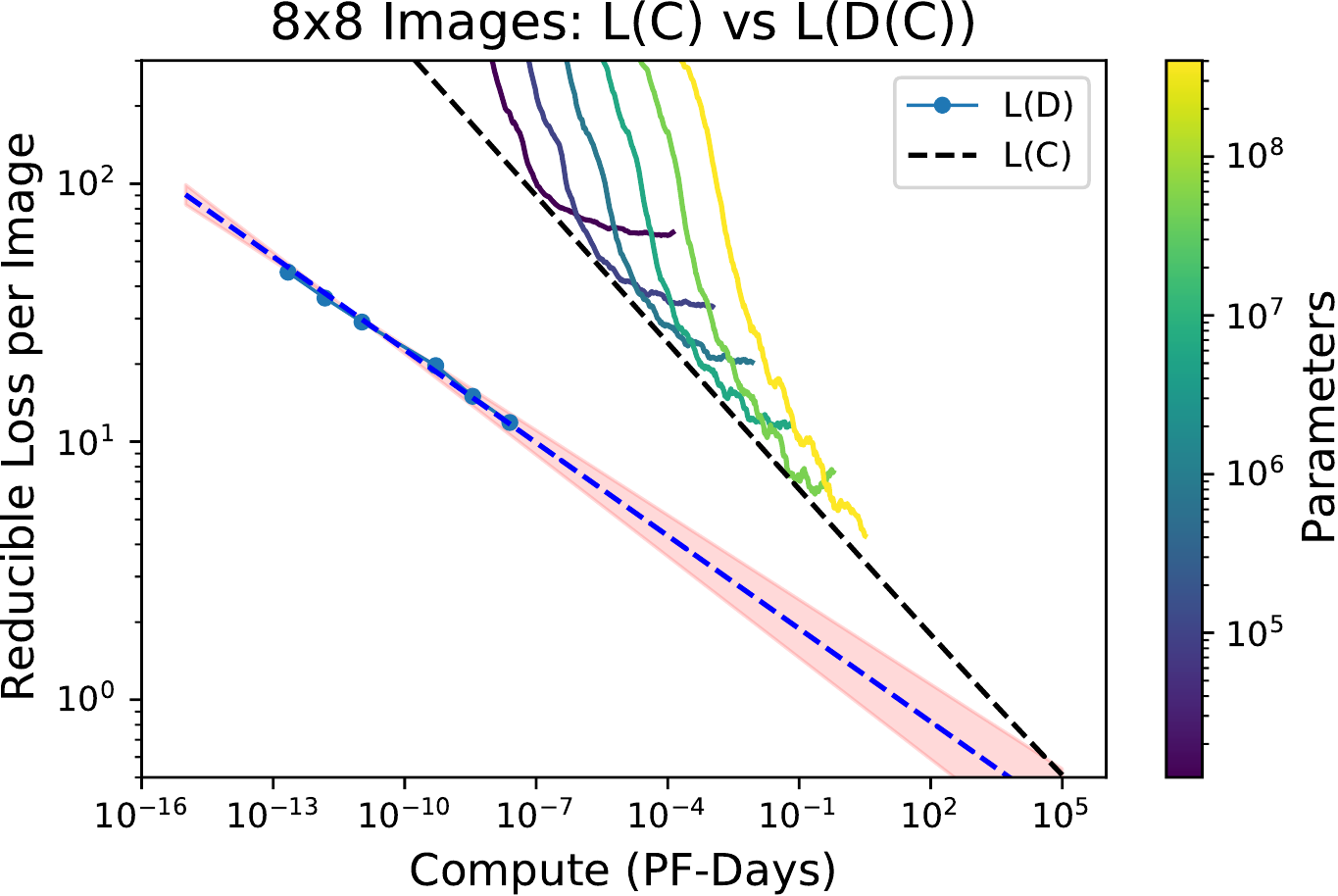}
\includegraphics[width=0.48\textwidth]{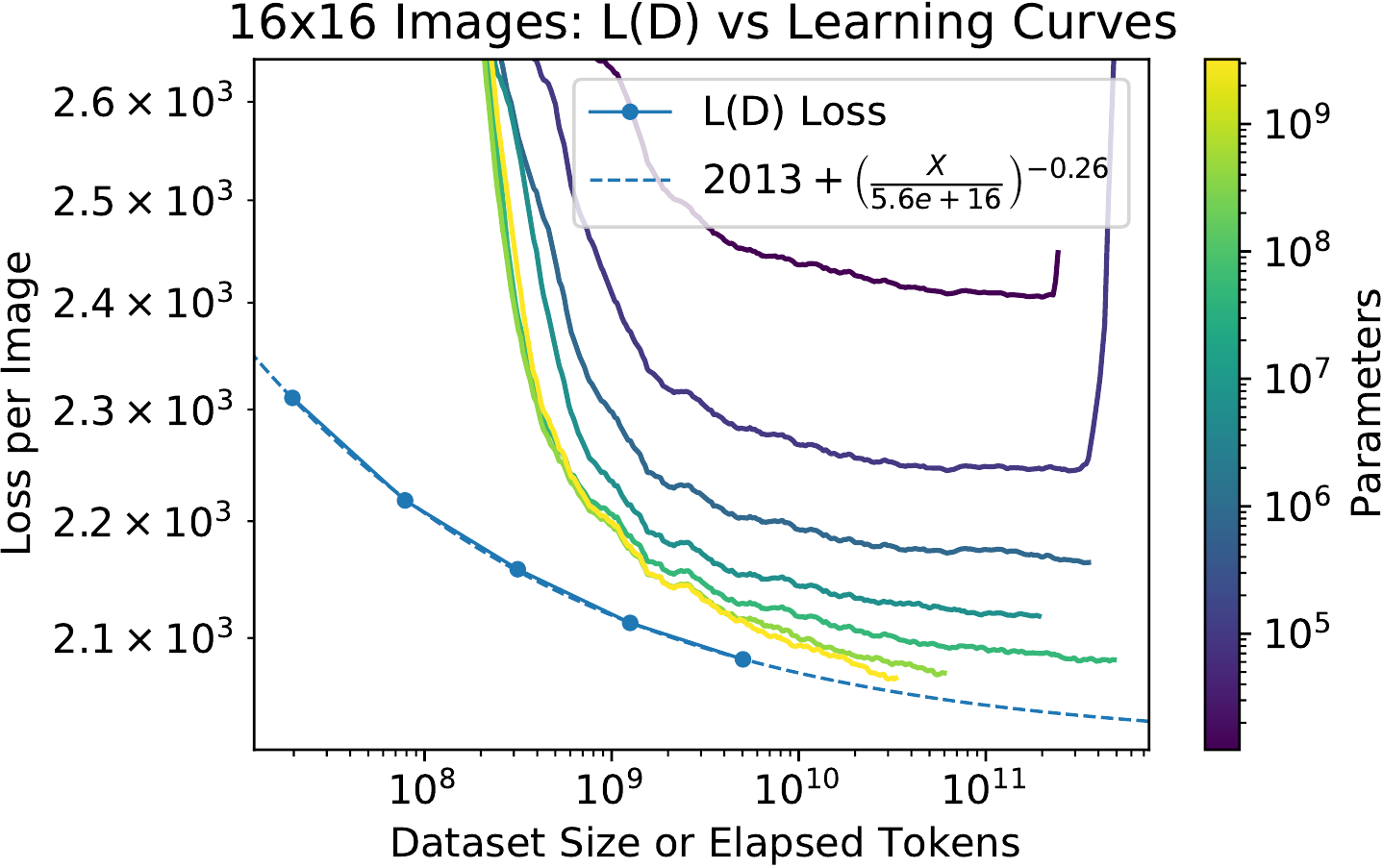}\hfill
\includegraphics[width=0.48\textwidth]{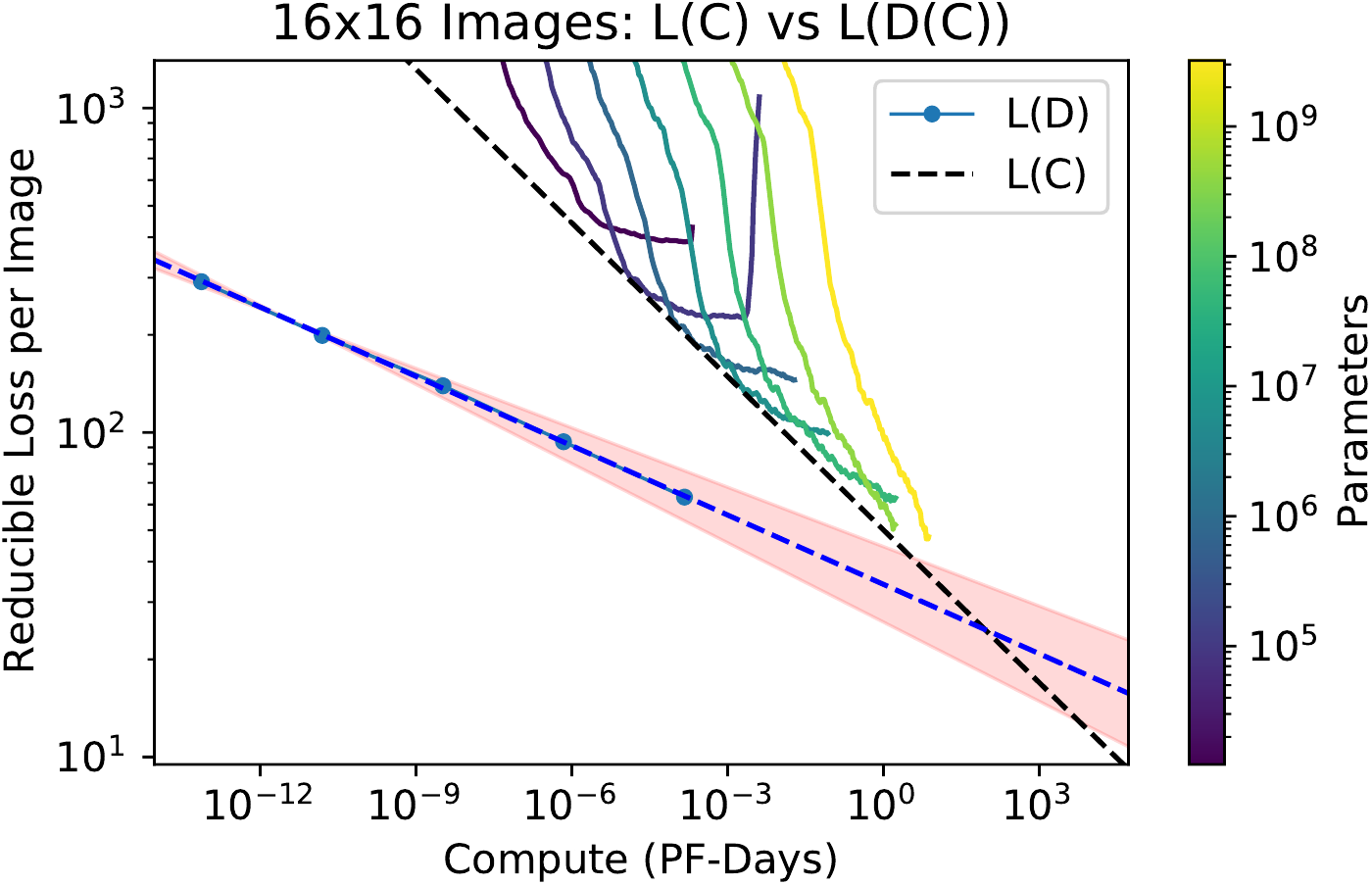}
\caption[Paradox]{\textbf{Training speed approaches a limit---} {\bf Left}: These figures show learning curves for various model sizes, along with the trend for fully trained, early-stopped $L(D)$, identifying the dataset size in tokens with the number of elapsed tokens during training.  We observe that the learning curves are approaching $L(D)$ as model size increases. {\bf Right}:  We show learning curves along with the $L(C)$ trend in black.  On the same plot we show $L(D)$ vs $C(D)$ in blue, where the latter is determined by identifying the optimal proportion of compute to allocate to tokens, and then assuming this corresponds to one epoch of training.  By construction all learning curves must lie above and to the right of the blue dashed line, so the intersection of the black and blue lines suggests a breakdown of some trend. The red shaded region corresponds to altering the optimal model size exponent by $\pm 5$\%, illustrating that projections are extremely sensitive to these trends. \label{fig:Paradox}  }  
\end{figure}

\begin{figure}
\noindent \centering{} 
\includegraphics[width=0.48\textwidth]{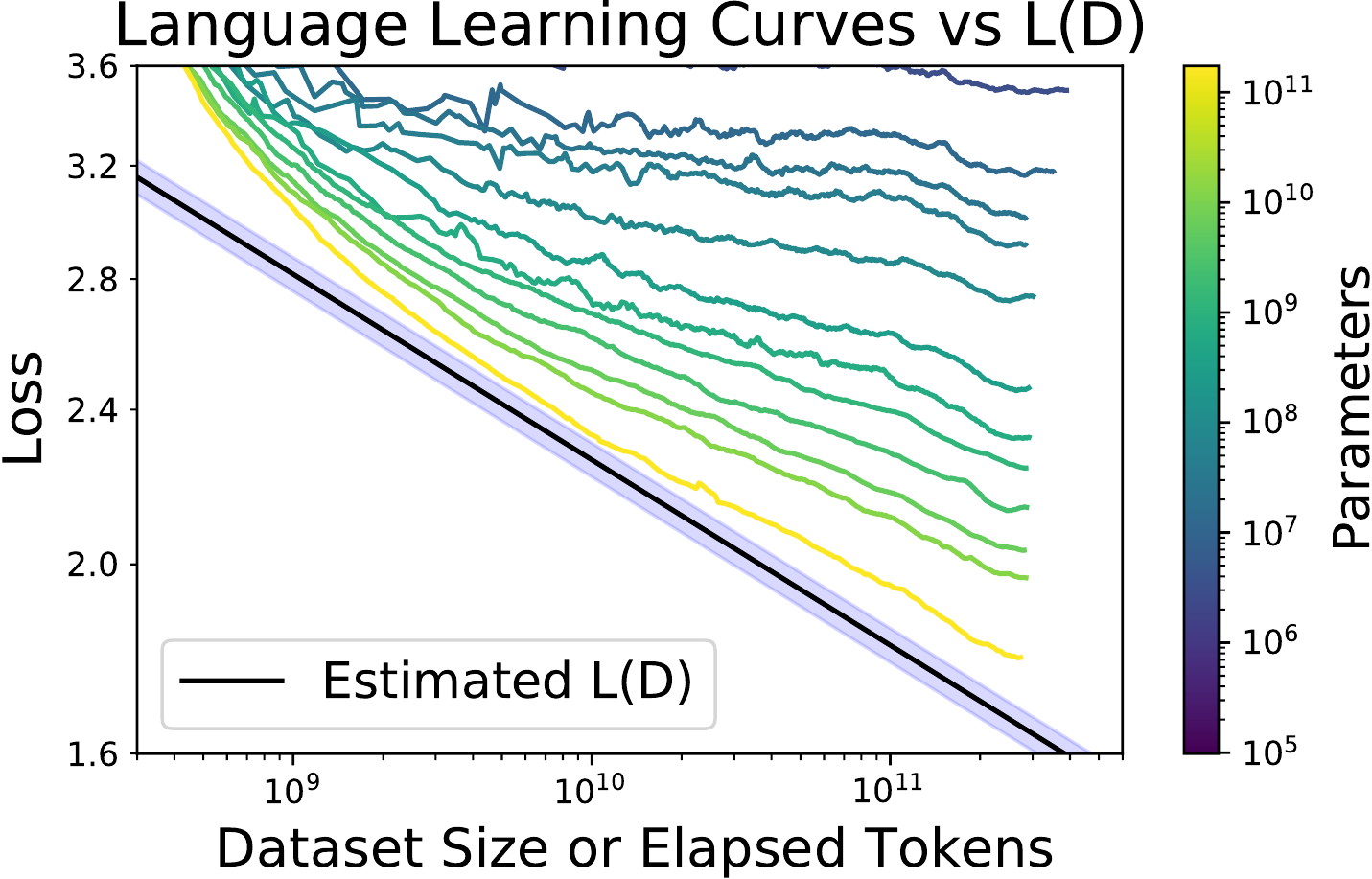}
\caption[Paradox]{\textbf{Training speed approaches a limit (language)---} Here we show an approximation of $L(D)$ with 2\% estimated errors, and the language modeling learning curves from \cite{brown2020language}.  The $L(D)$ trend comes from \cite{kaplan2020scaling}, but the models in that work were trained on a slightly different data distribution and with half the context length of \cite{brown2020language}.  \label{fig:LanguageLofDvsLearning}}
\end{figure}

An inconsistency among the datasize and compute scaling laws was observed in \cite{kaplan2020scaling}.  In this section we will study the same phenomenon using image models on low resolution images, though we expect the  results will be qualitatively the same on any of the datasets we have covered.

Before discussing the inconsistency, consider the plots on the left of figure \ref{fig:Paradox}.  We show both learning curves and the trend $L(D)$ for trained models, identifying the dataset size with the number of tokens seen by various models during training.  The learning curves lie above the $L(D)$ trend because the optimization process fails to achieve the minimum loss in a single epoch.  If the optimizer were perfect (in a sense), then $L(D)$ would coincide with the learning curve, assuming performance is not limited by model size.  Note that as model size increases, the learning curves appear to approach ever closer to the $L(D)$ trend.  This means that larger models learn faster, and it also implies that optimization becomes increasingly effective as model size increases.  But learning curves  will always be bounded by $L(D)$, which sets the  sample efficiency.  We show the same phenomena for language in figure \ref{fig:LanguageLofDvsLearning}, though we can only estimate\footnote{We need to account for slightly different data distributions, and context lengths differing by a factor of 2.  We estimate that these produce errors less than about 2\% of the loss, which we show as a shaded region on the plot.} $L(D)$ for these models.

To see an apparent inconsistency, we must compare the projections from two different trends.  For the $L(C)$ compute trend we can just reproduce results from figure \ref{fig:ReducibleLossImagesbyResolution}.  To plot $L(D)$ with compute on the x-axis, we will use the power-law trend $N_{\rm opt}(C) \approx (2.8 \times 10^8)  C^{0.74}$ for 16x16 images (see figure \ref{fig:OptimalModelSizeScaling}), where $C$ is measured in petaflop-days.  From this we can solve for the optimal number of tokens processed during training using $C = 6 D N$, which leads to $C(D) \approx (5 \times 10^{-42}) D^{3.9}$ where $D$ is measured in tokens. A similar analysis applies to 8x8 images. Using these results we can plot $L(D)$ vs $C(D)$ parametrically, as shown on the right of figure \ref{fig:Paradox} for the reducible\footnote{For these figures we subtract the irreducible loss measured from each of $L(D)$ and $L(C)$, respectively, since numerically the irreducible losses from these two measurements are not exactly equal.} loss (chosen for clarity on the log plot).  We have also included a shaded region showing the effect of changing the empirically extracted $N_{\rm opt}(C)$ trend exponent by $\pm 5$\%.  

The inconsistency arises because all learning curves must lie above the $L(D)$ trend on the right of figure \ref{fig:Paradox}, but the extrapolation of $L(C)$ eventually intersects and passes below $L(D)$.  Either $L(D)$, $L(C)$, or the $N_{\rm opt}(C)$ trend must break down at or before this intersection point.  Note that the existence of this intersection is an inevitable consequence of the power-law form of the trends, since these lead to straight lines on a log-plot, and two straight lines must intersect.

We do not know for certain how this inconsistency or its equivalent for language \cite{kaplan2020scaling} are resolved.  However, the observation of the left of figure \ref{fig:Paradox} and our earlier discussion suggests a  plausible hypothesis.  As we increase model and dataset sizes, optimization becomes increasingly efficient, until eventually learning curves begin to merge with the $L(D)$ trend, so that there are no benefits to be gained from training for more than a single epoch \cite{1906.06669}.  Near the intersection point, the compute frontier would bend and become coincident with $L(D)$.  From this point of view, the fact that $L(C)$ appears steeper than $L(D(C))$ is due to a deficiency with optimization, which requires more than one epoch to reach a local minimum of the test loss.  It would be  interesting to investigate this hypothesis in the future.  If it is true, it suggests that the relative scaling of optimal model and dataset sizes may eventually change, and perhaps will ultimately be set by trends for overfitting  such as those found in \cite{rosenfeld2019constructive, kaplan2020scaling}.

Finally, we note that the irreducible loss from dataset size trend is measured at $L(D =\infty) \approx 2013$ nats/image (16x16), and 599 nats/image (8x8), while that extracted from compute trends is $L(C=\infty) \approx 2023$ nats/image (16x16), and 602 nats/image (8x8).  These estimates for the entropy of low-resolution YFCC100M images are quite similar, and provide a consistency check on our results.

\section{Related Work}

Predictable scaling trends for modern neural networks have been studied by a variety of groups, beginning with \cite{1712.00409}.  
More recently \cite{rosenfeld2019constructive, li2020train, roller2020recipes, 1906.06669, rosenfeld2020predictability} studied scaling relations using many model architectures and datasets, with the work on language modeling in \cite{kaplan2020scaling} closest to our approach here.  Work on the 175B parameter GPT-3 model \cite{brown2020language} was partially motivated by neural scaling laws.  

There has not been a great deal of  work on  theoretical explanations for the very precise scaling relations we and others have identified. A simple theory connecting scaling exponents to the inverse of the dimension of the data manifold was proposed in \cite{sharma2020neural}.  Expansions in the model size, particularly at large width  \cite{1902.06720, jacot2018neural} may provide another useful framework for thinking about some of our scaling relations, if they are in fact applicable \cite{lewkowycz2020large} to optimally tuned hyperparameter settings. 

The models and data modalities we used have been widely studied in the past. Autoregressive image models have been trained starting with PixelRNN \cite{DBLP:journals/corr/OordKK16}, with the recent work \cite{icml2020_6022} nearly identical to our models and training procedure.  Transformer-based video models were trained in \cite{weissenborn2019scaling} and multimodal models in \cite{tsai2019multimodal}.  The original authors trained various models, including transformers, on the math problem dataset \cite{DBLP:journals/corr/abs-1904-01557}, and it has also been studied with more specialized architectures \cite{schlag2019enhancing}.  Our models are typically simpler than many of those that have been previously discussed, as we exclusively use decoder-only \cite{liu2018generating} transformers with dense or sparse \cite{DBLP:journals/corr/abs-1904-10509} attention.

\section{Discussion}

We have argued that a single neural architecture, the Transformer, can be applied to the generative modeling of images, videos, multimodal data, and math, along with language \cite{kaplan2020scaling, brown2020language}.  We identified common scaling laws for the loss achieved on all data modalities as a function of both  model size and compute budget.  As in the case of language, these results imply that larger models become more sample efficient. Furthermore, we found that in some important cases, finetuned performance on downstream tasks also follows similar scaling laws. This  suggests that trends in the generative modeling loss translate into advantages in practical capabilities.

A greater surprise was the approximately universal trend (figure \ref{fig:OptimalModelSizeAllDomains}) for optimal model size as a function of the training compute budget -- we did not anticipate that the exponent $N_{\rm opt} \propto C^{0.7}$ would be largely independent of the data distribution.  This trend implies a dual trend for the number of tokens elapsed during optimized training, as a function of $C$ or $N$, and leads to the conclusion that larger compute budgets should be `spent' mostly on larger models, rather than much longer training runs.  So this lesson from language modeling \cite{kaplan2020scaling} generalizes.
These empirical regularities beg for theoretical explanation -- why do these  scaling relations hold?  

The scaling laws also suggest a shift in perspective away from the particularities of neural architectures, loss functions, and training algorithms and towards the broader commonalities that appear when machine learning is studied across a large hierarchy of model, data, and compute scales.  Work in ML often involves identifying  specific deficiencies in current capabilities and remedying them through the alteration of models and algorithms.  Perhaps many capabilities simply lie on a spectrum that can be continuously unlocked with increasing scale, as might be suggested by the metalearning capabilities of the GPT-3 model \cite{brown2020language}. 

We also discussed some information theoretic implications of the scaling laws.  Perhaps the most important point was that the two terms in equation (\ref{eq:PowerLawPlusConstant}) can be interpreted as the  entropy of the true data distribution, and the KL divergence between that distribution and a given generative model.  The identification of the entropy was made possible through the extrapolation of a precise trend, and would not be predictable using  the results from a single model.   We also observed intriguing scaling laws for the empirical mutual information between images and captions in multimodal models.  This is particularly interesting because the mutual information must be bounded by the entropy of the caption.

\section*{Acknowledgments}

We thank Yasaman Bahri, Miles Brundage, Yura Burda, Paul Christiano, Ajeya Cotra, Psyho Debiak, Ethan Dyer, Harri Edwards, Danny Hernandez, Jacob Hilton, Jaehoon Lee, Brice Menard, Chris Olah, Utkarsh Sharma, and Ilya Sutskever for discussions and feedback on this work.

Thanks as well to Chris Berner, Ben Chess, Eric Sigler, and Clemens Winter for managing and scaling the supercomputing clusters and research platform that allowed us to run these experiments.
\vfill
\section*{Contributions}
\label{sec:contributions}
\textbf{Tom Henighan} performed and analyzed the image and video modeling experiments, and maintained the codebases for experimentation and data analysis that enabled our results.

\textbf{Jared Kaplan} performed and analyzed the math experiments, led the overall data analysis, and wrote the paper.

\textbf{Mor Katz} performed the multimodal experiments and data analysis.

\textbf{Jacob Jackson}, \textbf{Chris Hesse}, \textbf{Heewoo Jun}, and \textbf{John Schulman} collaborated on video modeling experiments.

\textbf{Jacob Jackson}, \textbf{Heewoo Jun}, \textbf{Prafulla Dhariwal}, and \textbf{Alec Radford}, developed the VQ-VAE training strategies and codebase.

\textbf{Sam McCandlish} analyzed the progression of question-answering capabilities in language models.

\textbf{Aditya Ramesh} and \textbf{Alec Radford} provided guidance on multimodal modeling and optimization.

\textbf{Chris Hallacy} and \textbf{Alec Radford} curated the multimodal datasets.

\textbf{Heewoo Jun} and \textbf{Aditya Ramesh} curated the image datasets.

\textbf{Chris Hesse}, \textbf{Heewoo Jun}, and \textbf{Alec Radford} curated the video datasets.

\textbf{Mark Chen} provided guidance on image modeling and finetuning.

\textbf{Tom Brown}, \textbf{Scott Gray}, \textbf{Benjamin Mann}, \textbf{Nick Ryder}, \textbf{Prafulla Dhariwal}, and \textbf{Daniel Ziegler} built, optimized, and maintained our codebase for training large transformer models.

\textbf{Dario Amodei} advocated for a broad study of scaling laws for generative modeling.

\textbf{Sam McCandlish} and \textbf{Jared Kaplan} led the research.

\newpage
\appendix

\begin{figure}
\noindent \centering{} 
\includegraphics[width=0.32\textwidth]{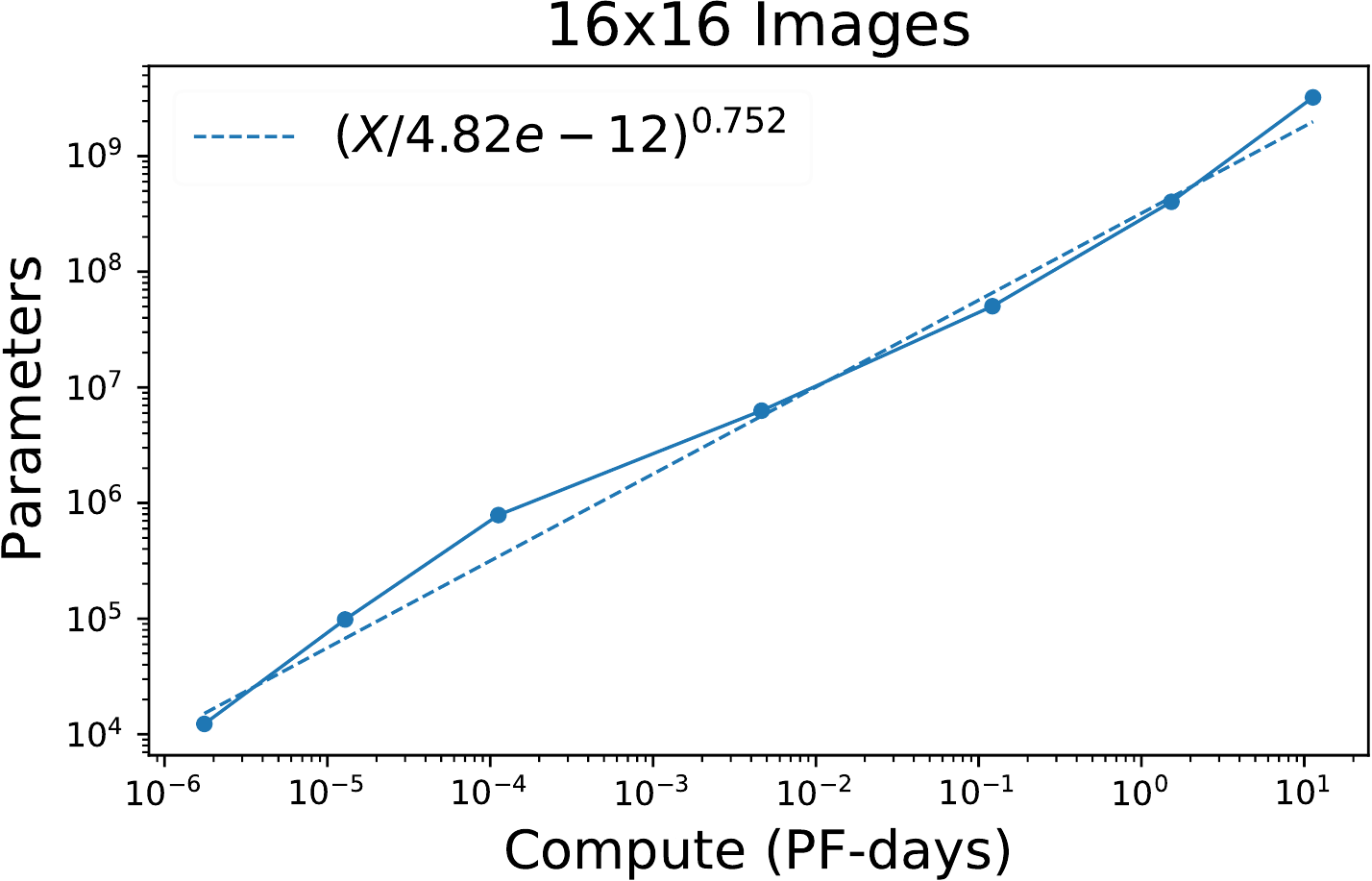}\hspace{2em}
\includegraphics[width=0.32\textwidth]{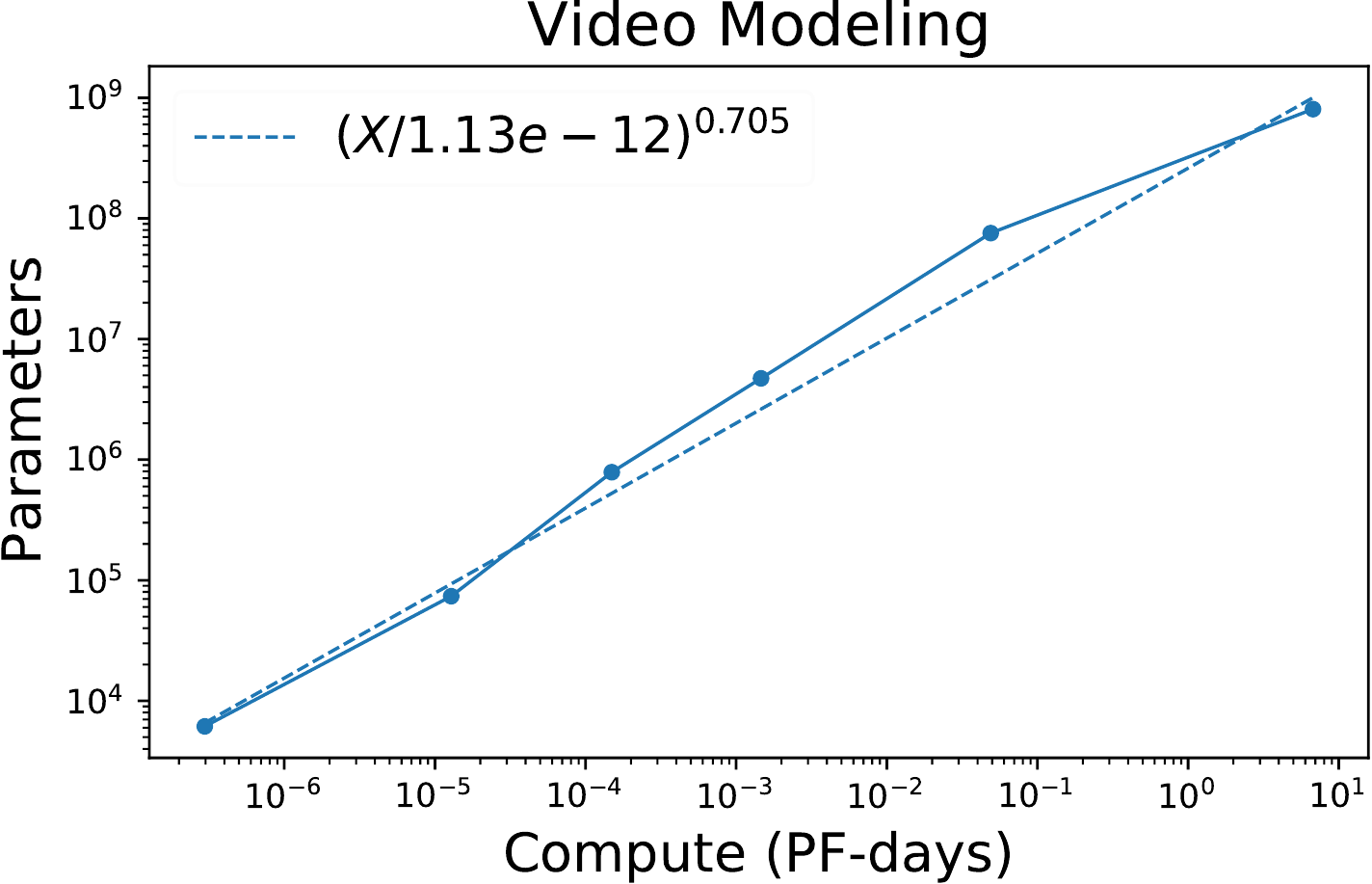}
\\\vspace{1em}
\includegraphics[width=0.32\textwidth]{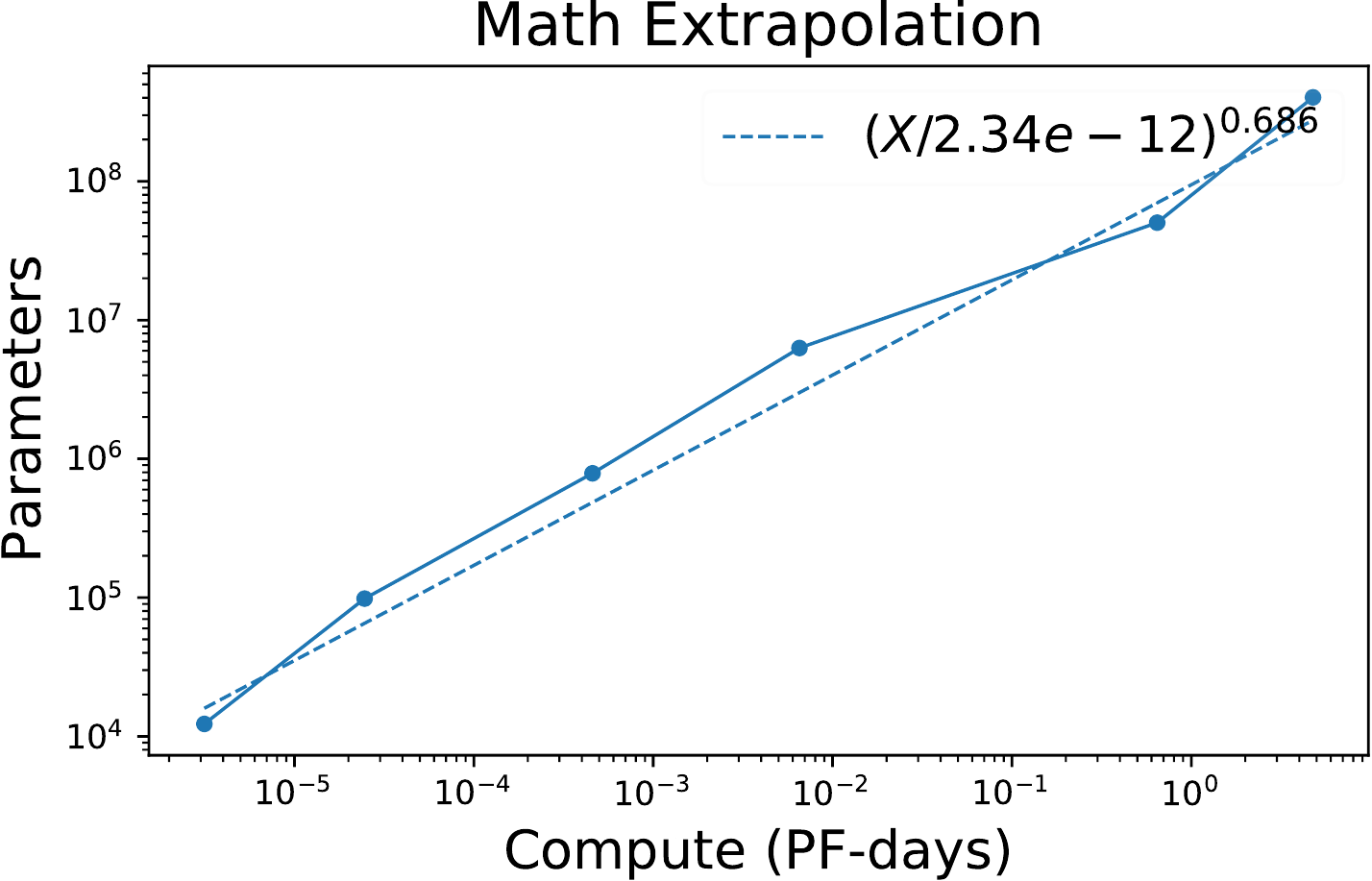}\hfill
\includegraphics[width=0.32\textwidth]{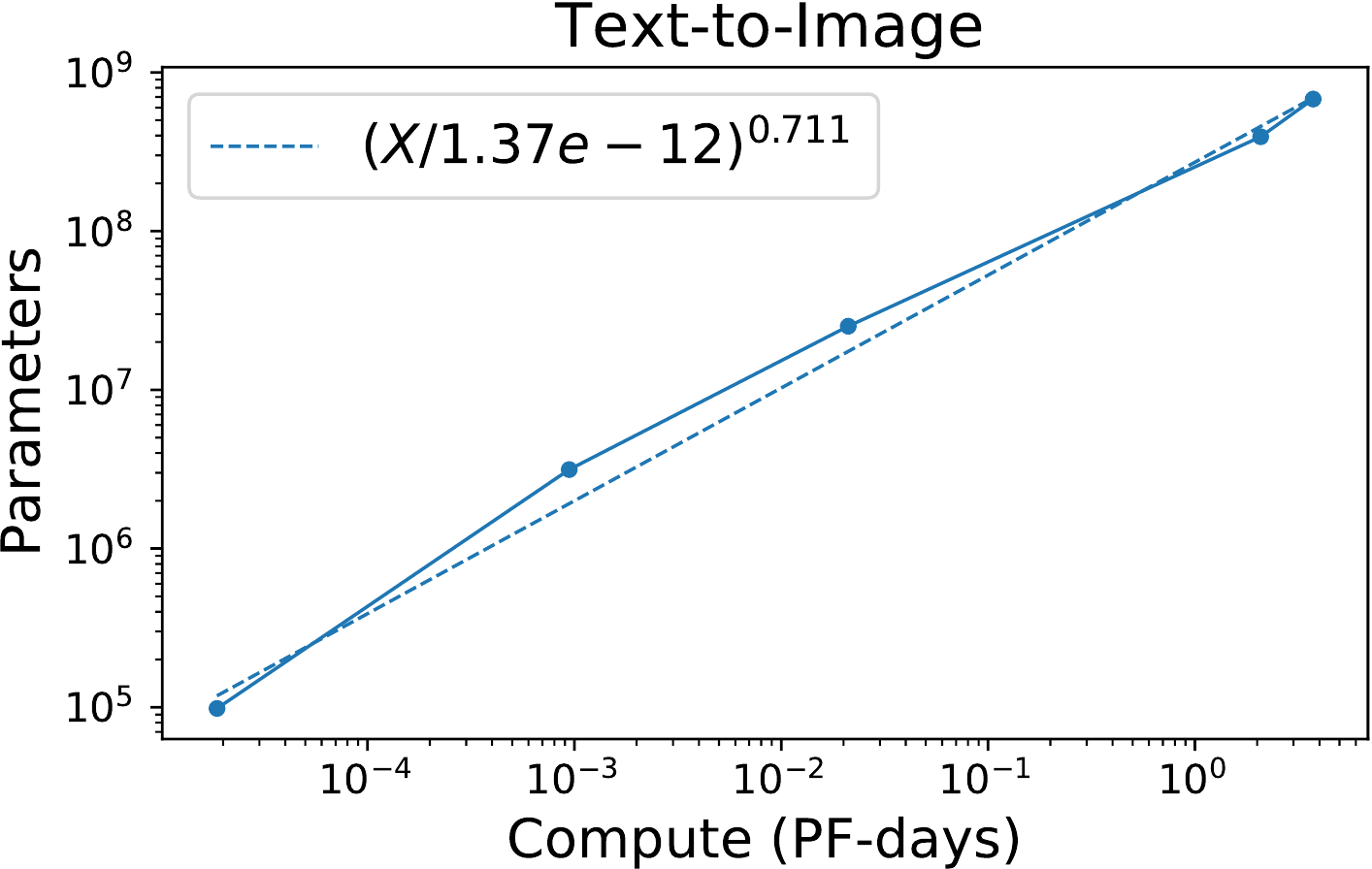}\hfill
\includegraphics[width=0.32\textwidth]{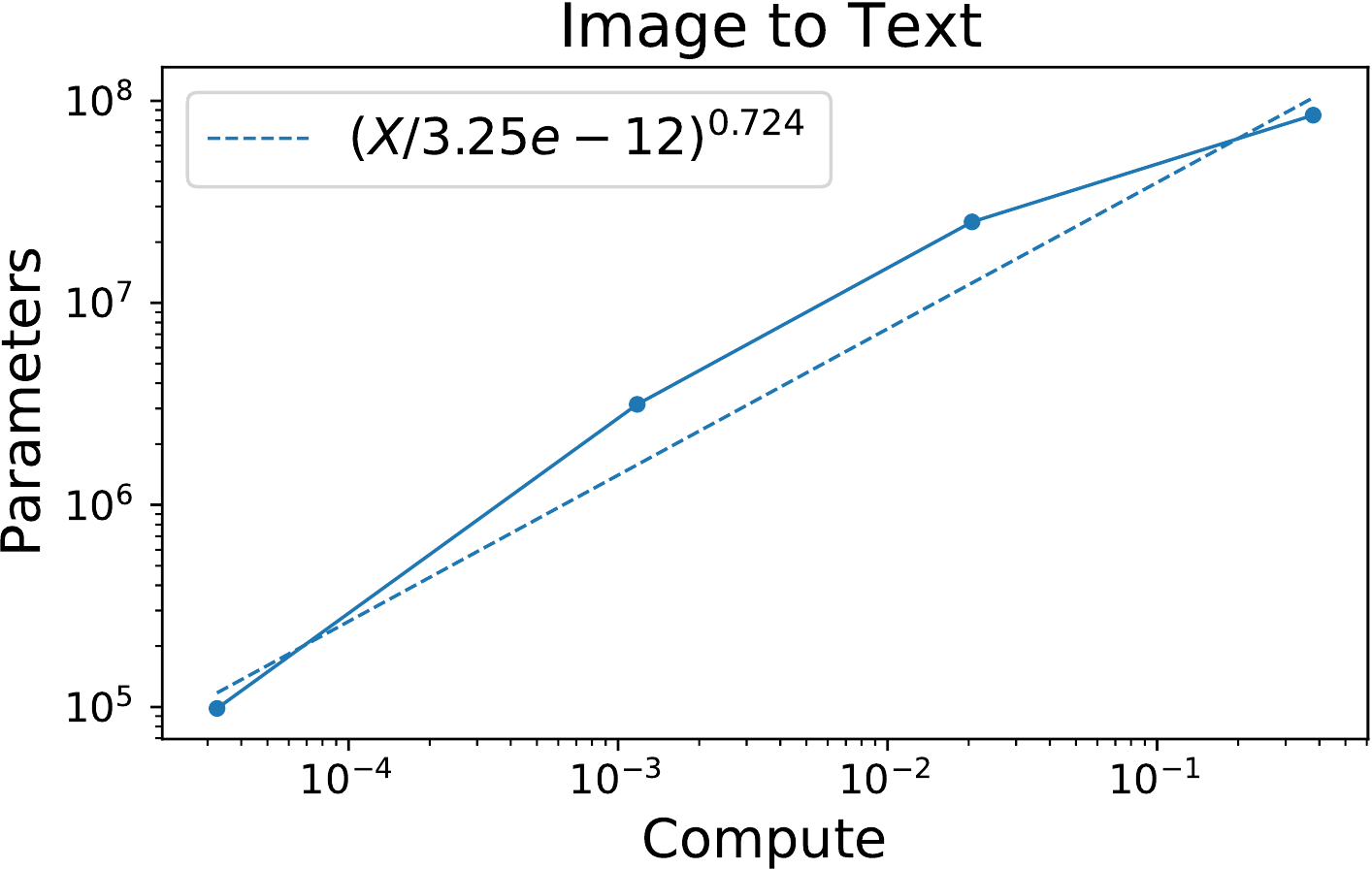}
\caption[Optimal Model Size]{\textbf{Optimal model size (individual trends)---} We show the optimal model size for a given compute budget, along with power-law fits, based on the points at the compute-efficient frontier of figure \ref{fig:ComputeScalingFullLoss}. These trends are combined in figure \ref{fig:OptimalModelSizeAllDomains}. \label{fig:OptimalModelSizeScaling}}
\end{figure}

\section{More Details on Image Modeling}
\label{app:MoreOnImageModels}

\begin{figure}
\noindent \centering{} 
\includegraphics[width=0.9\textwidth]{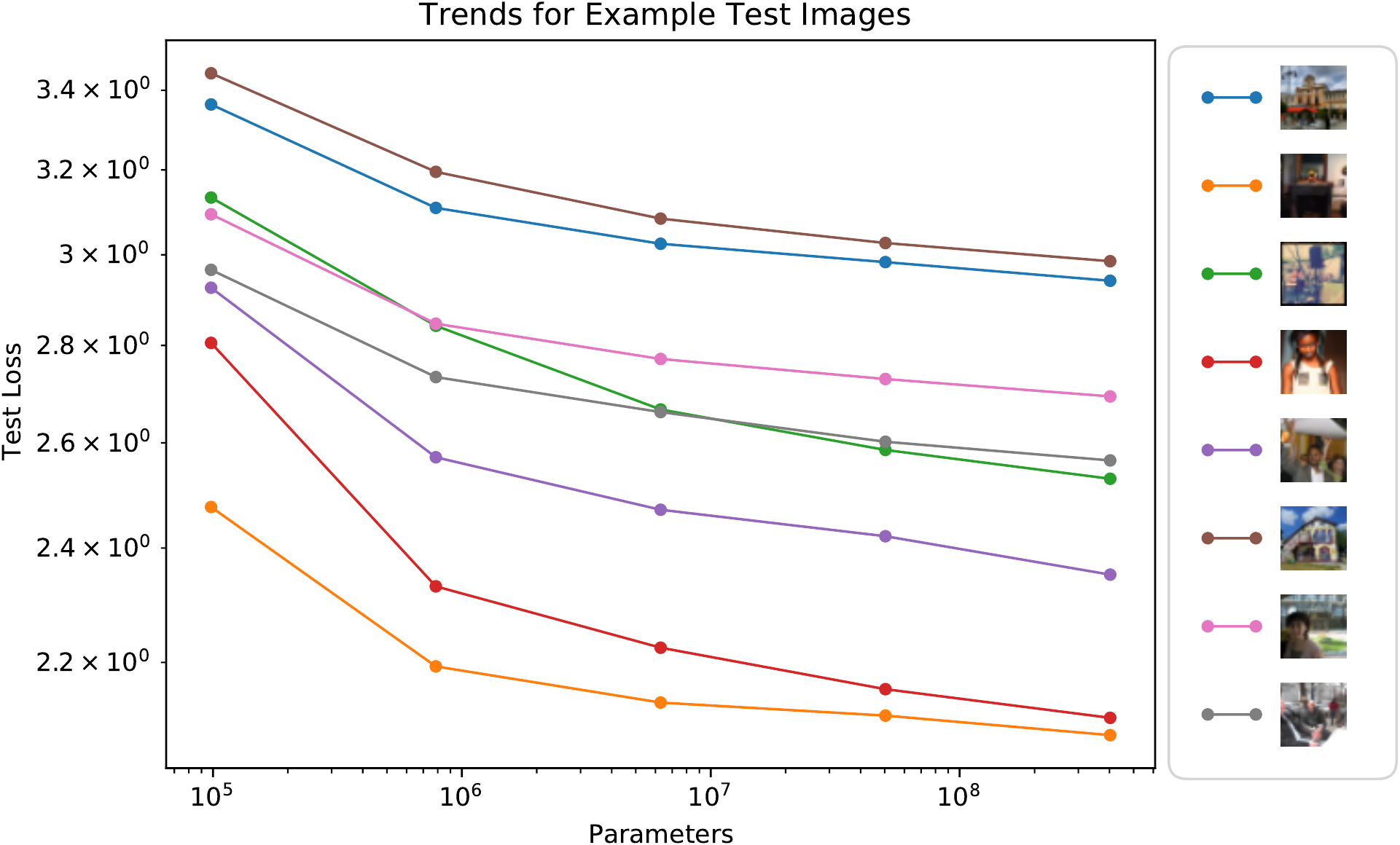}
\caption[Image Improved]{\textbf{Loss trend for individual images---} We show the loss trend for eight randomly chosen images from the test set.  These results are fairly typical. \label{fig:TrendsIndividualImages}}
\end{figure}

\begin{figure}
\noindent \centering{} 
\includegraphics[width=0.3\textwidth]{Image8x8ComputeFullLoss}\hfill
\includegraphics[width=0.3\textwidth]{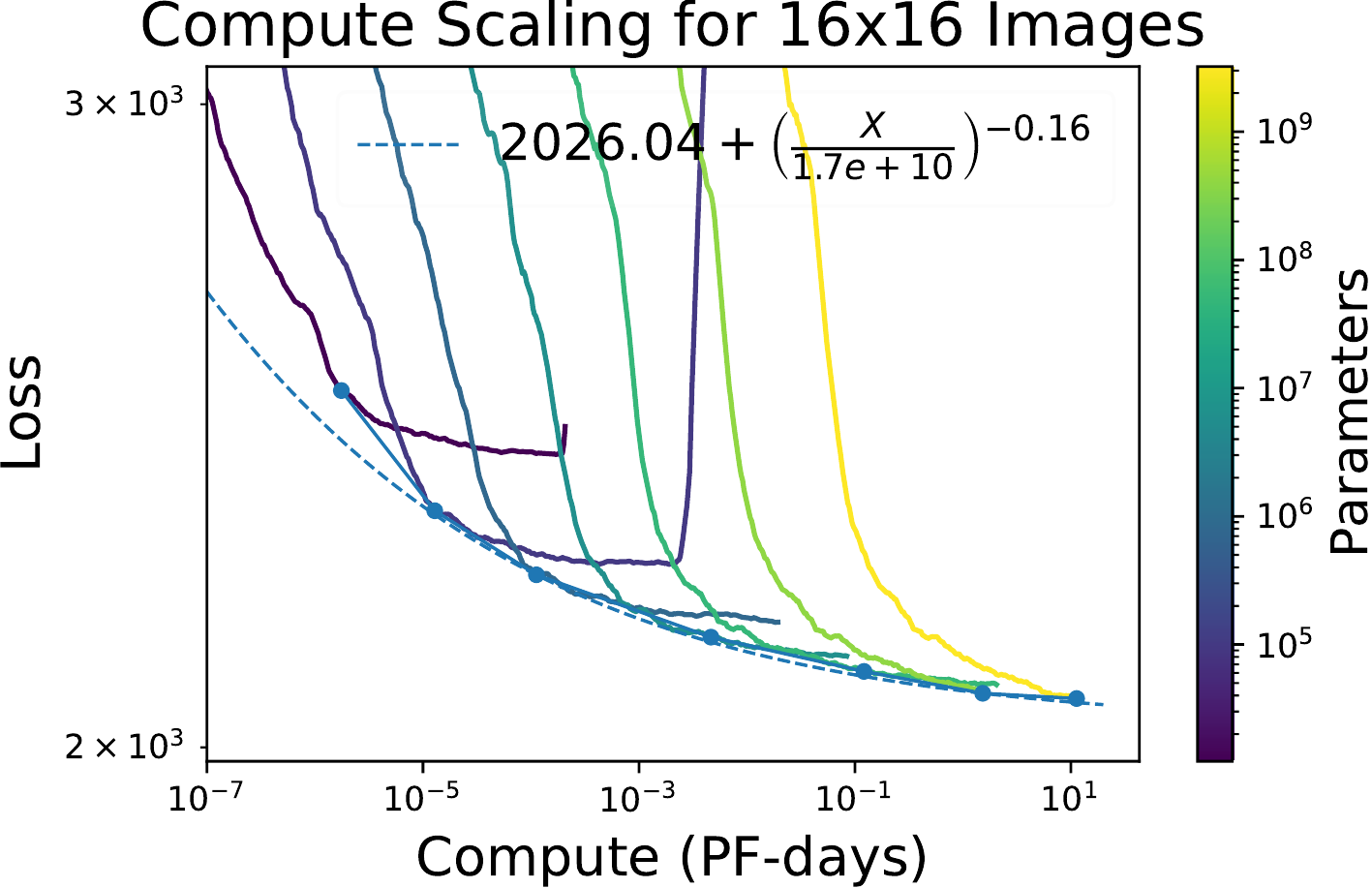}\hfill
\includegraphics[width=0.3\textwidth]{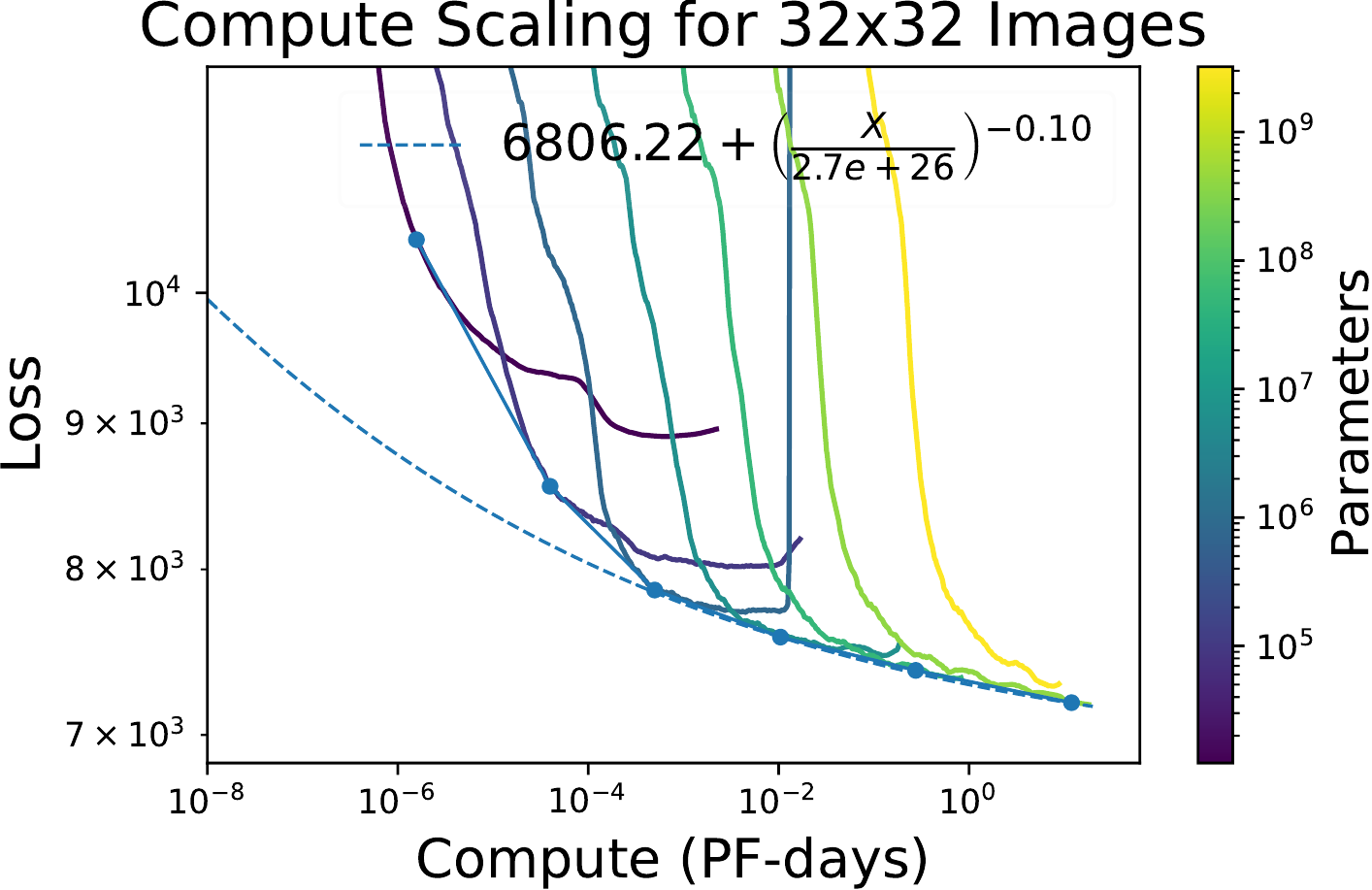} \\\vspace{1em}
\includegraphics[width=0.3\textwidth]{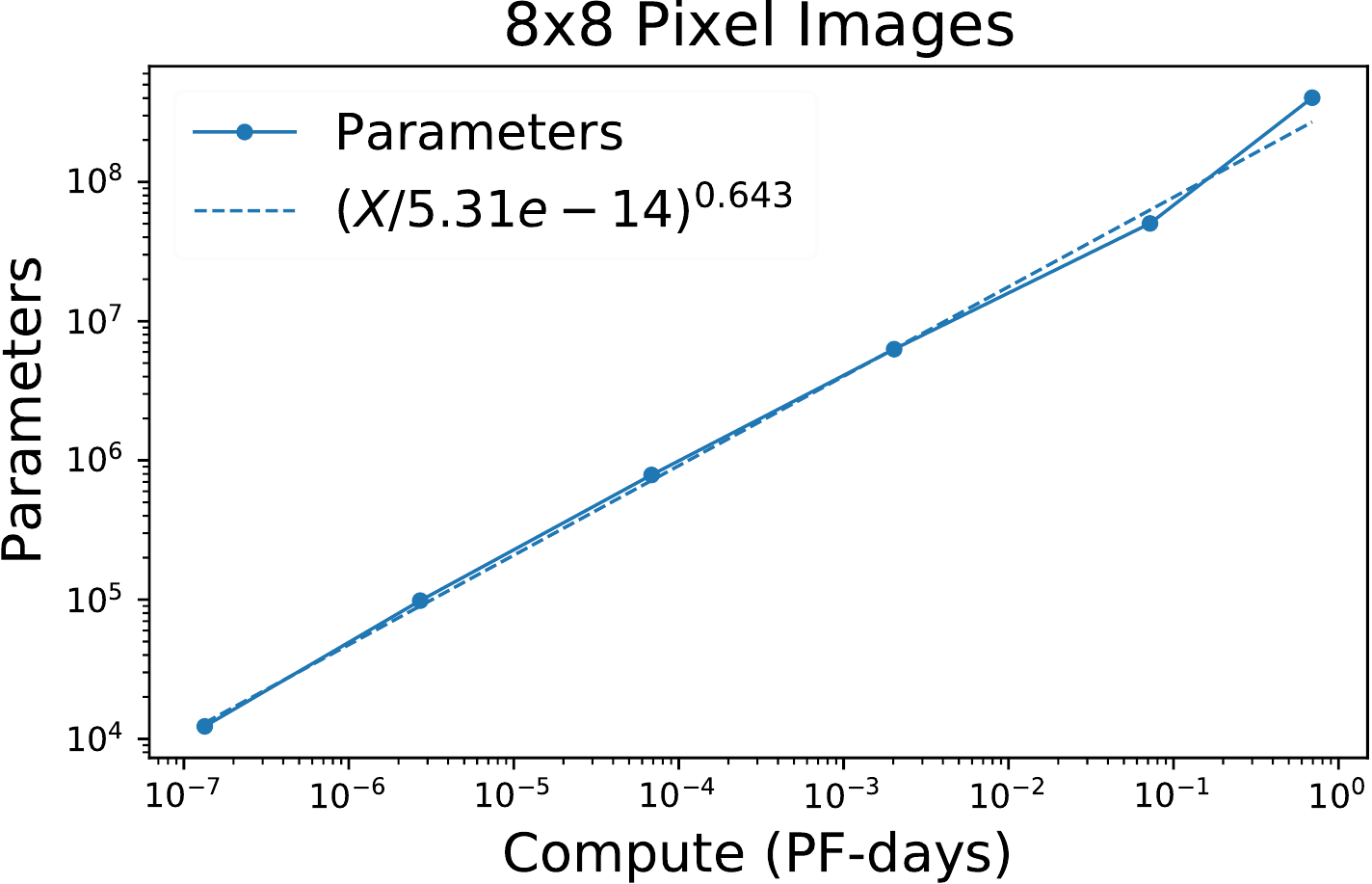}\hfill
\includegraphics[width=0.3\textwidth]{Image16OptimalParametersvsCompute}\hfill
\includegraphics[width=0.3\textwidth]{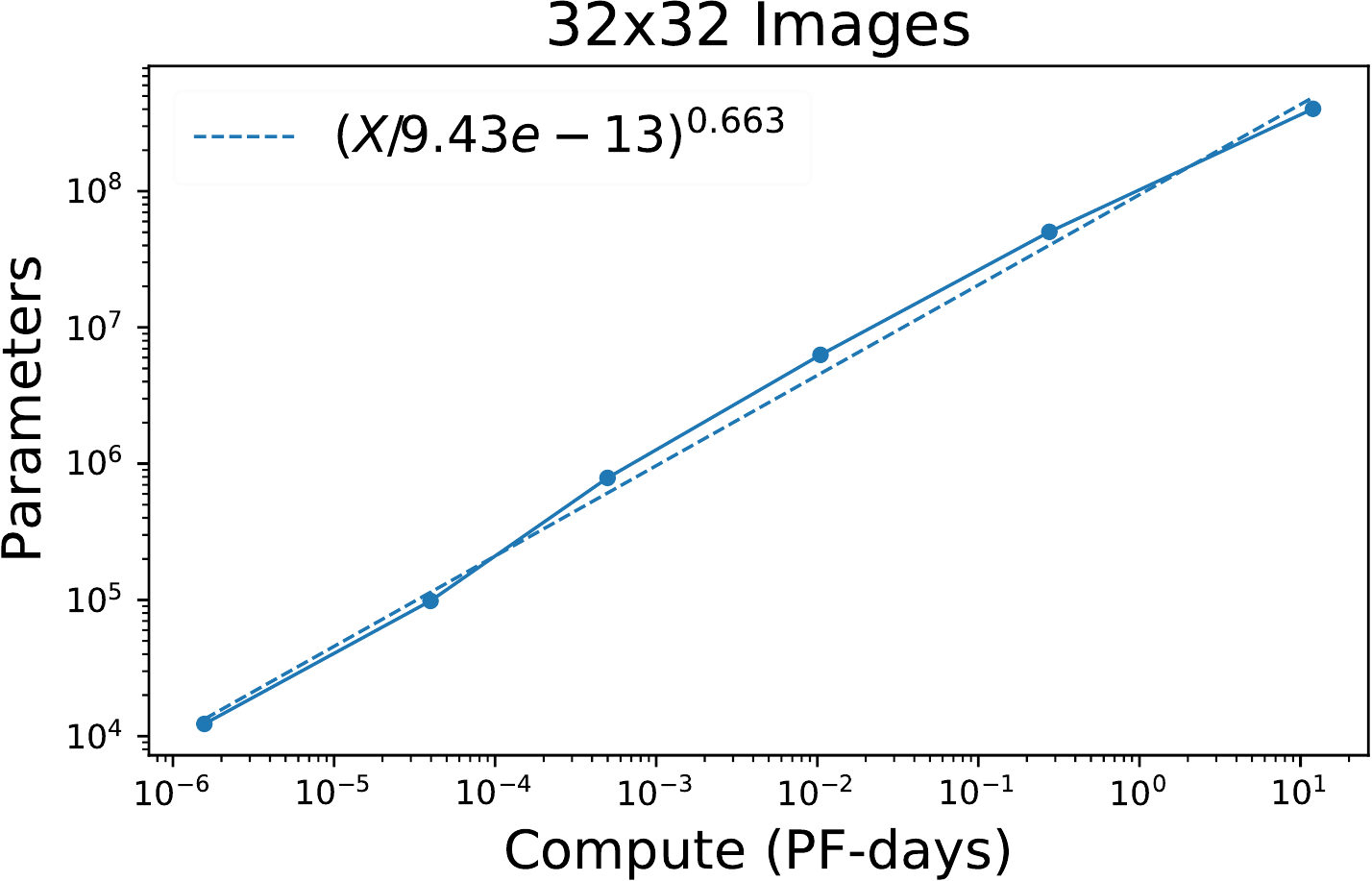} \\
\caption[Image Resolution Compute Scaling]{\textbf{Compute trends for varied image resolution (pixel-level)---} Scaling laws with compute for various image resolutions in pixels, along with power-law plus constant fits (dashed) to equation (\ref{eq:PowerLawPlusConstant}).  The fits for pixel-level image modeling are shown in table \ref{tab:ImageResolutionTrends}.  \label{fig:ComputeScalingImageResolution}}
\end{figure}

\begin{figure}
\noindent \centering{} 
\includegraphics[width=0.32\textwidth]{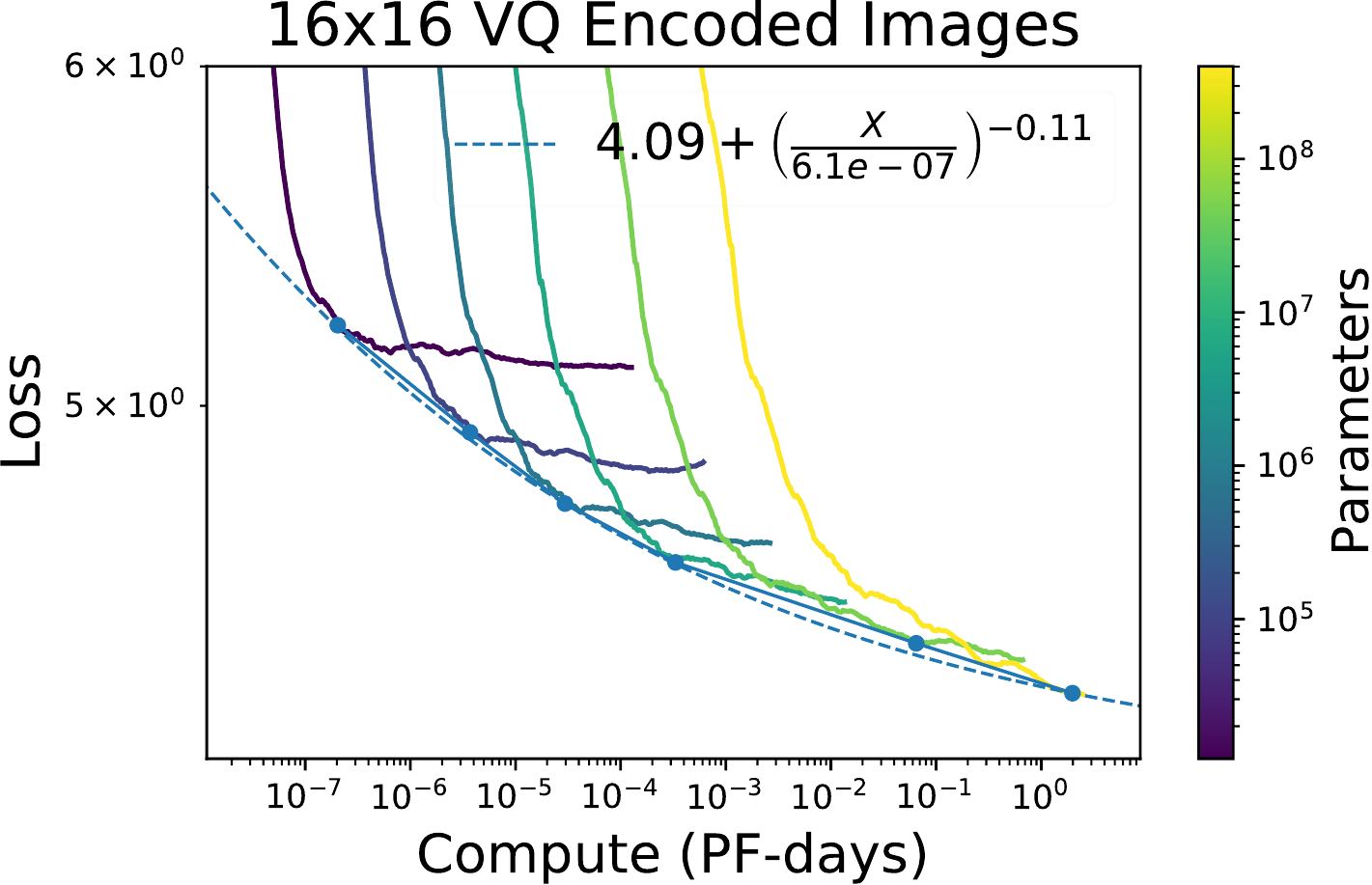}\hfill
\includegraphics[width=0.32\textwidth]{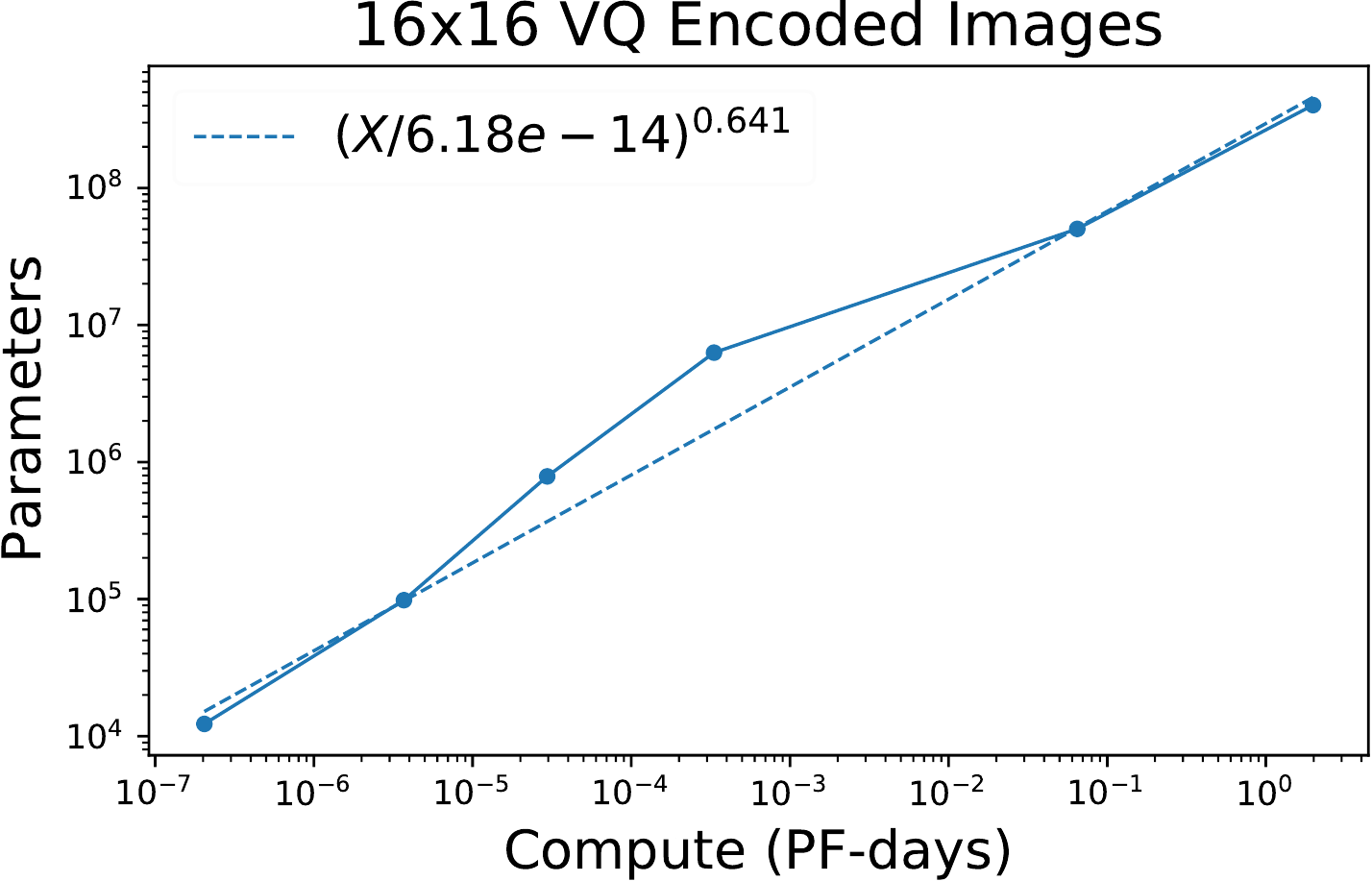}\hfill
\includegraphics[width=0.32\textwidth]{ImageVQ16ComputeReducibleLoss}
\\\vspace{1em}
\includegraphics[width=0.32\textwidth]{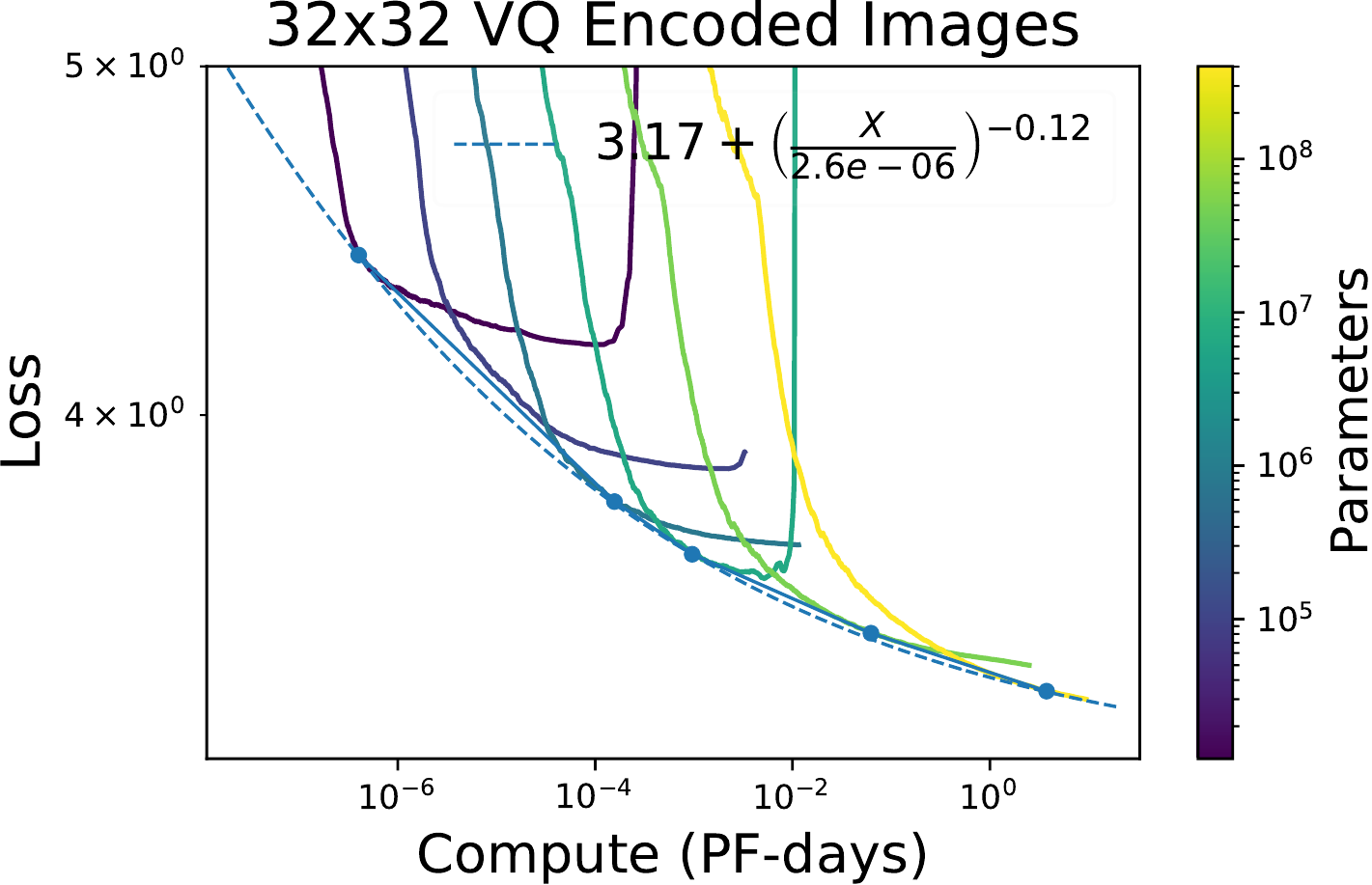}\hfill
 \includegraphics[width=0.32\textwidth]{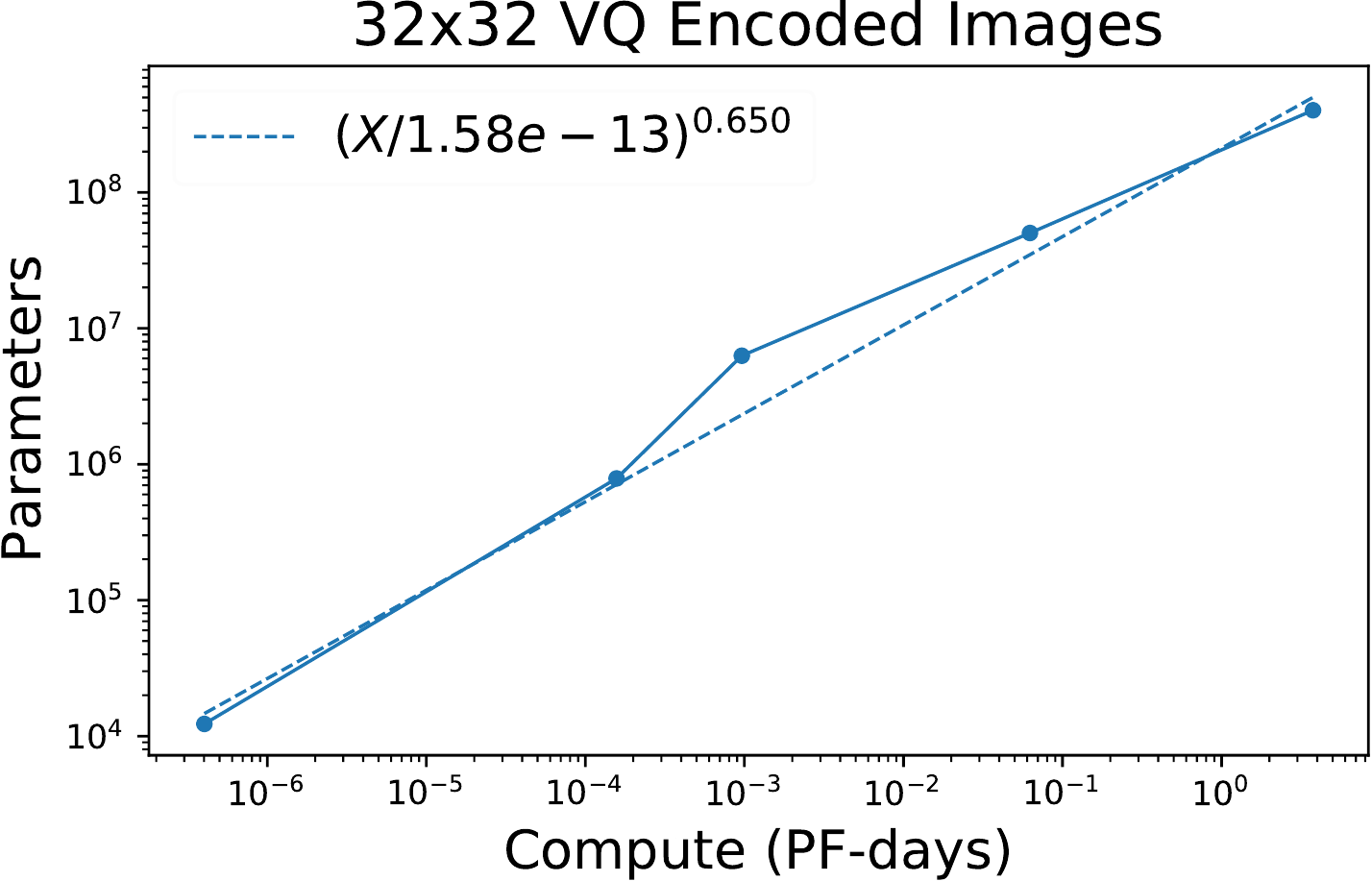} \hfill
\includegraphics[width=0.32\textwidth]{ImageVQ32ComputeReducibleLoss}
\caption[VQ Compute Scaling]{\textbf{Compute trends for various image resolutions (VQVAE-encoded)---} We display scaling laws with compute for 64x64 images encoded with two different VQ code resolutions, along with power-law plus constant fits (dashed) to equation (\ref{eq:PowerLawPlusConstant}).  A few of these runs diverged beyond the compute frontier; in the worst case this led to a visible deviation from the model size trend in figure \ref{fig:ModelSizeScalingImageResolution}. \label{fig:ComputeScalingImageVQ}}
\end{figure}

\begin{figure}
\noindent \centering{} 
\includegraphics[width=0.7\textwidth]{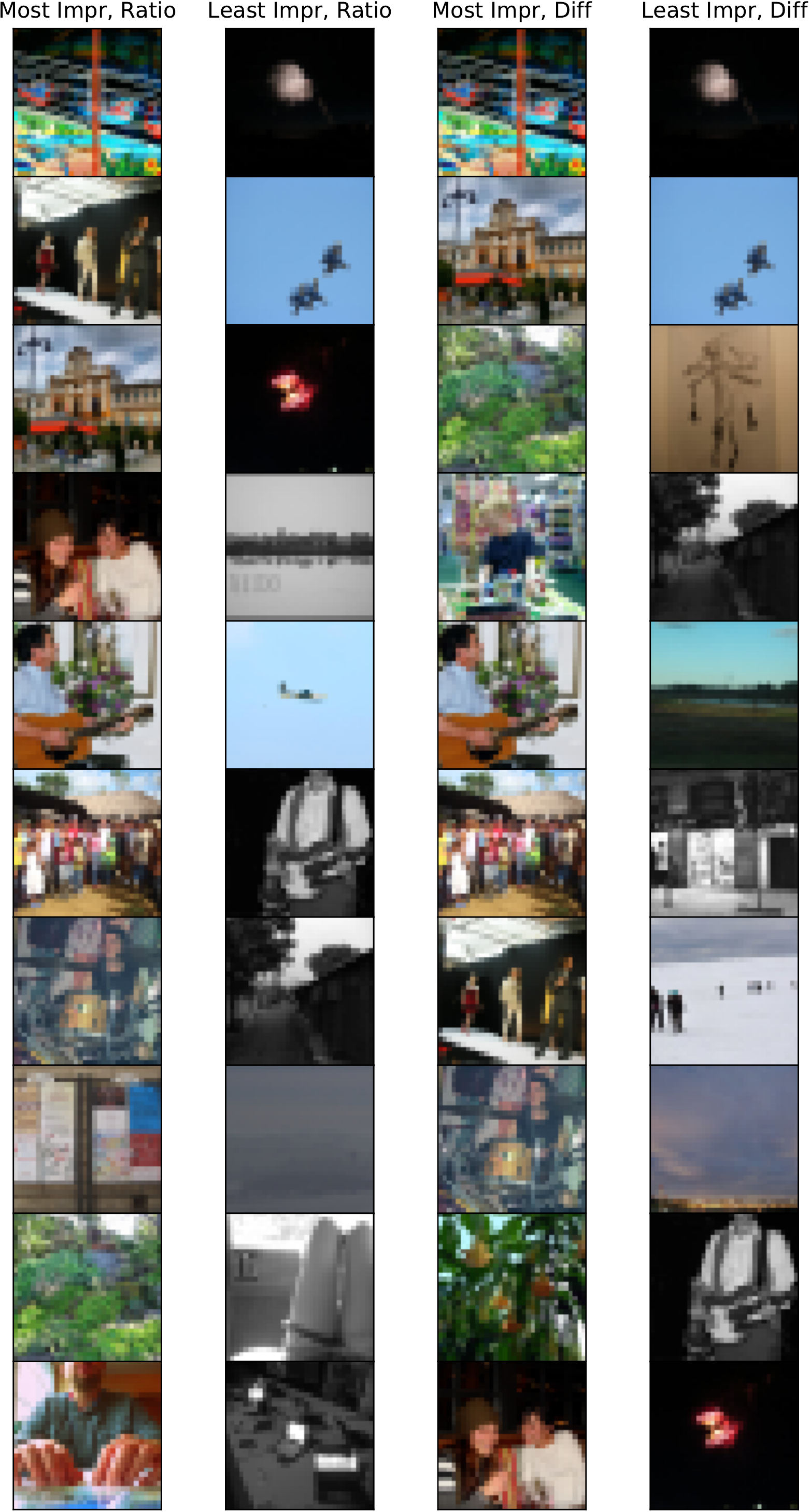}
\caption[Image Improved]{\textbf{Most and least improved images---} Here we show the images where the loss improved most or least between models with 400M parameters and 100k parameters.  These were the ten most or least improved images from a random sample of one thousand images in the test set, as measured by loss ratio and loss difference.  Images with complex colorful scenes involving people or crowds are typically most improved, while black and white images and those dominated by a simple background tend to be the least improved. \label{fig:MostLeastImprovedImages}}
\end{figure}

In figures \ref{fig:ComputeScalingImageResolution} and \ref{fig:ComputeScalingImageVQ} we provide some additional information documenting compute scaling trends for images with different resolutions and encodings.  In figure \ref{fig:MostLeastImprovedImages} we show images where the loss improved most or least as we pass from a 100k parameter model to a 400M parameter model.  In figure \ref{fig:TrendsIndividualImages} we also show trends for randomly selected individual images from the test set.

\begin{figure}
\noindent \centering{} 
\includegraphics[width=0.6\textwidth]{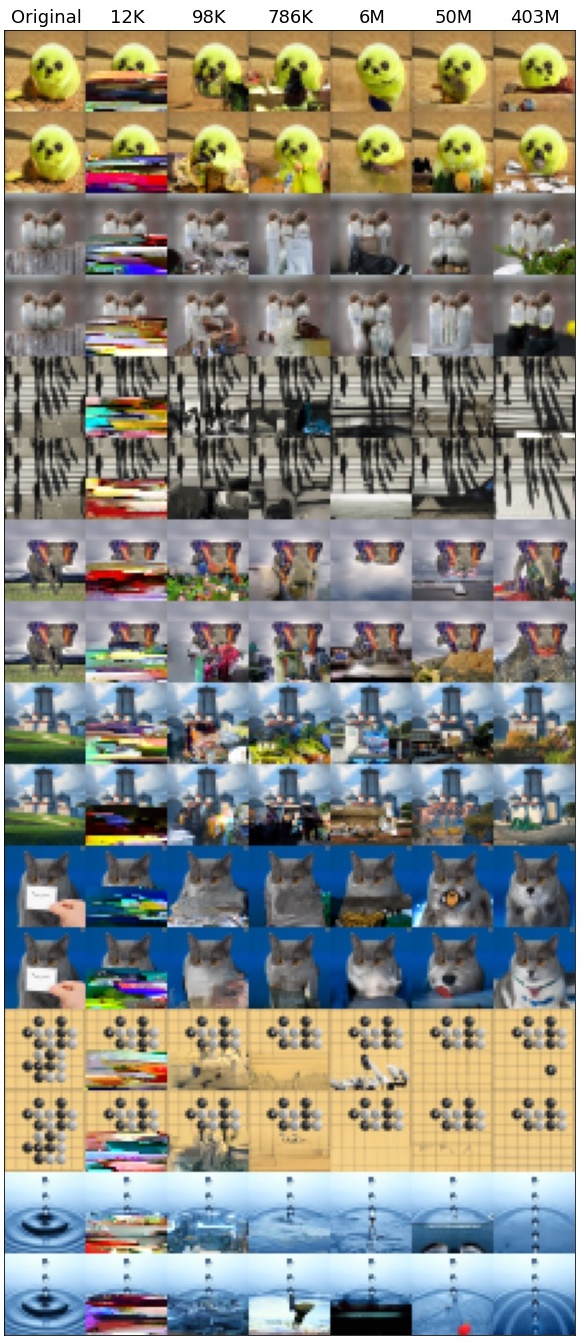}
\caption[32x32 Pixel Image Completions by Model Size]{\textbf{Trends in image completion quality---} Here we show conditional-completions of 32x32 pixel models of various sizes, where the leftmost column is the original image, and each of the other columns shows completions from a model with a non-embedding parameter count labeled at the top. Models are provided the top half of the image as conditional context and the bottom half is sampled with temperature 1.0. There is a clear trend of increasing photorealism with larger models. \label{fig:CompletionsBySize}}
\end{figure}

\clearpage
\section{Details of Math Experiments and Additional Results}
\label{app:MathDetails}

\subsection{Procedurally Generated Training Data}

We generated all training data procedurally using the code provided by \cite{DBLP:journals/corr/abs-1904-01557}.  Problems were generated by randomly sampling modules from the training distribution, with an `entropy' setting sampled uniformly from the integers $s \in [3,10]$.  The number of problems with entropy $s$ is approximately $10^s$, meaning that easy problems with low-entropy would likely be seen by the model many, many times, while some problems with $s \geq 9$ may not be seen at all.  This means that the easy components of the training distribution may be memorized.  Furthermore,  our procedurally generated data was not deduplicated from the `interpolate' test distribution \cite{DBLP:journals/corr/abs-1904-01557}, but it is completely disjoint from the `extrapolate' test distribution.  

The official extrapolate distribution only provides one difficulty level, and it also does not include all eight module-types.  So we also generated distributions of problems with smoothly increasing difficulty level by setting the entropy $s=1,2,\cdots 19$.  For most modules we simply used the interpolate settings, though for modules where other parameters were needed we generally used the extrapolation settings. Importantly, we did not include the  \textbf{probability\_\_swr\_p\_level\_set\_more\_samples} and \textbf{probability\_\_swr\_p\_sequence\_more\_samples generators}, as we found our models always performed  poorly on these problems, and quickly overfit on the loss for these generators (this can be seen in figure \ref{fig:MathPerformancebyModule}, where `probability' represents the mean of these two generators).  

Performance as a function of difficulty level and model size can be seen in figure \ref{fig:MathPerformancebyLevel}.  We note that performance degrades smoothly as we extrapolate away from the training distribution.

As an additional note, because these experiments were conducted much earlier, our dataset size scaling and aspect ratio scans use models with the fairly standard setting $m_{\rm mlp} = 4$ and $m_{\rm attn} = 1$, as with language and multimodal models, but different from the math models we used for compute and model size trends, where these parameters were smaller by a factor of 4, as with our image and video models.  We made this change to smaller $m_{\rm mlp}, m_{\rm attn}$ as we found it helped to improve the training stability of very deep math models.

It is also worth noting that we evaluated extrapolation performance both using the training data files provided with \cite{DBLP:journals/corr/abs-1904-01557} and by sampling with procedurally generated data (leaving out the two probability modules previously discussed).  For trend plots we have used the procedurally generated data, but for reporting final accuracies in figure \ref{fig:MathExtrapolatebyModule} we use the `official' files.

\subsection{Dataset Size Scaling}

\begin{figure}
\noindent \centering{} 
\includegraphics[width=0.48\textwidth]{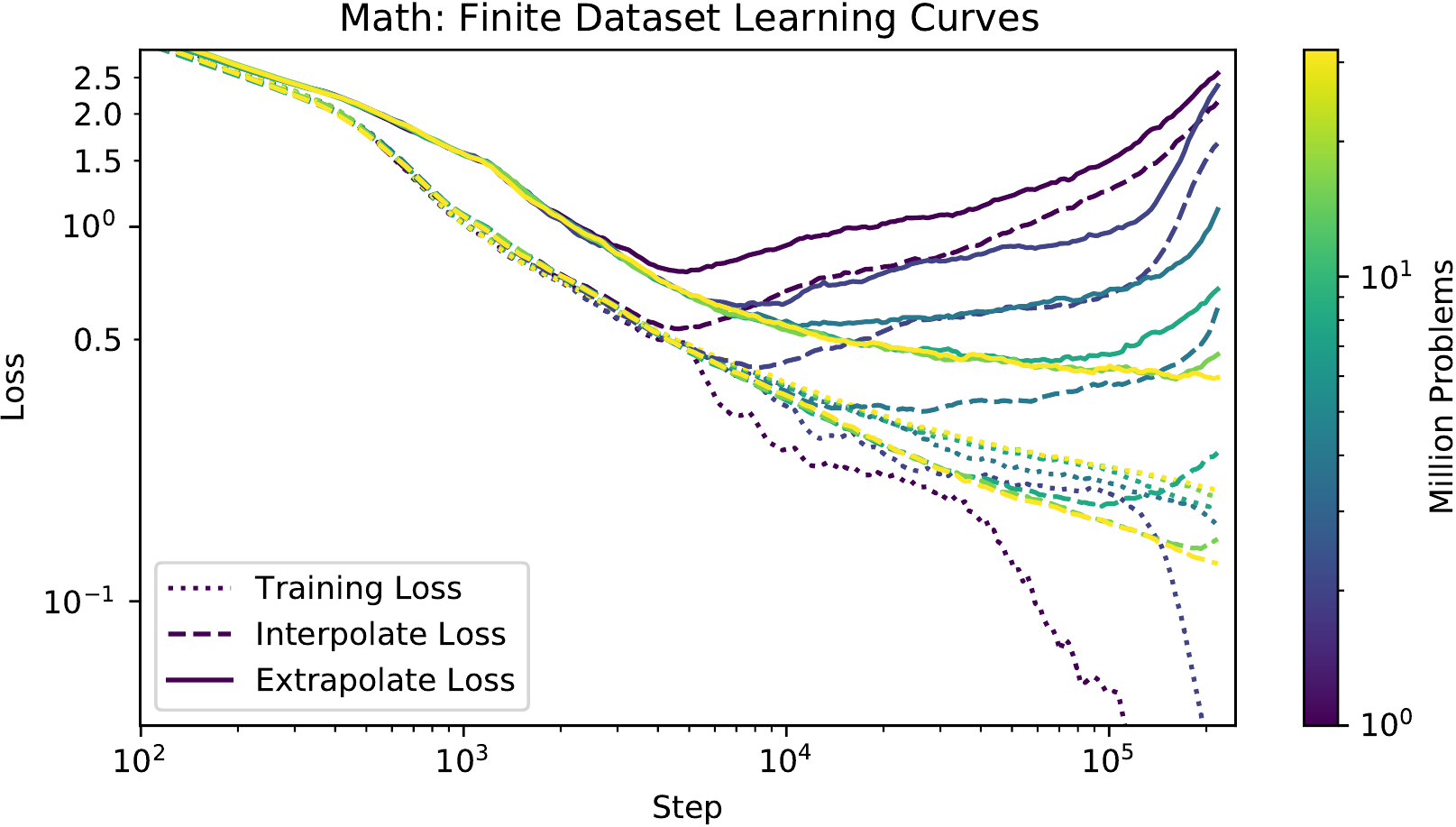}\hfill
\includegraphics[width=0.48\textwidth]{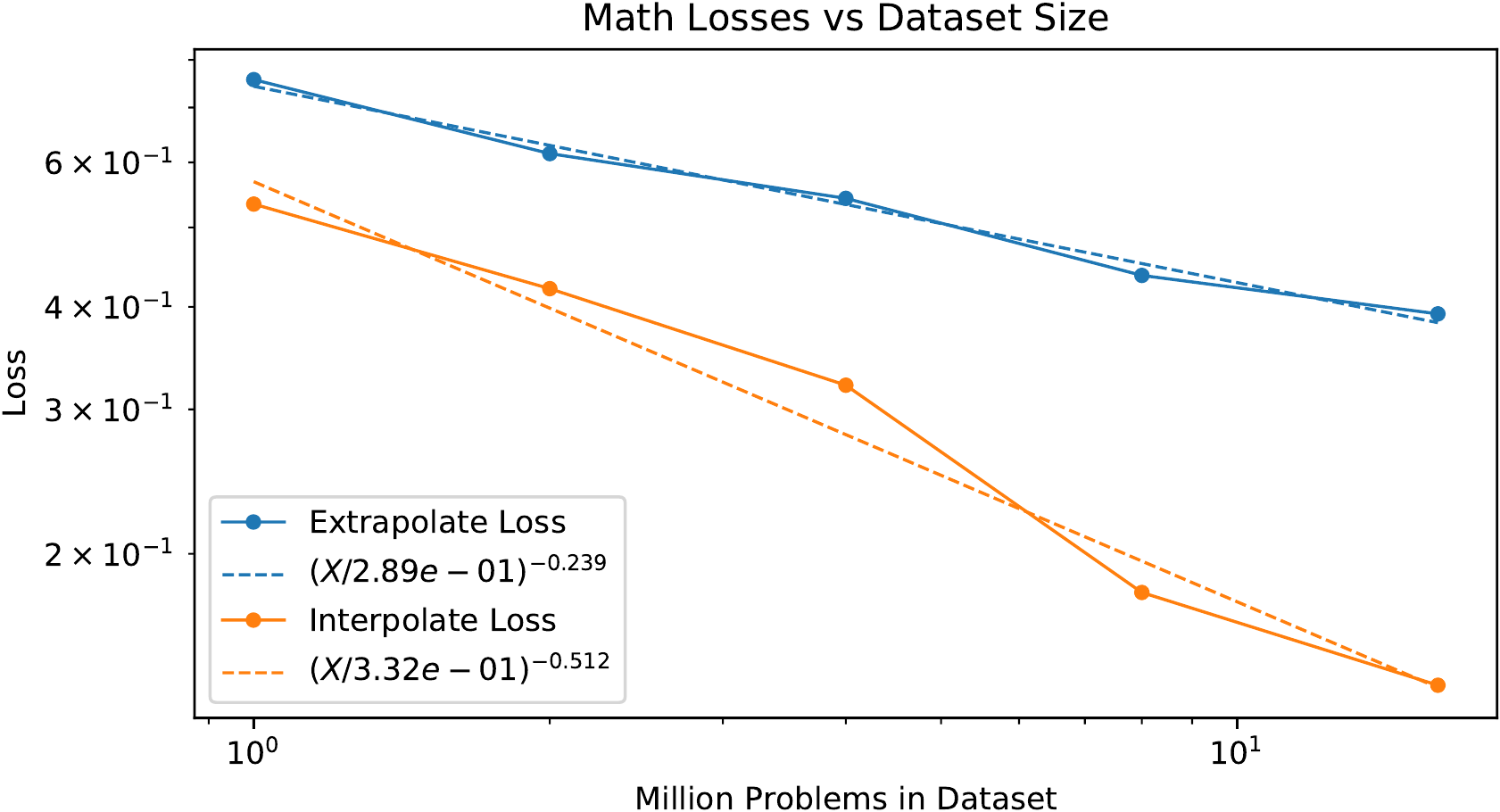}
\caption[Dataset Size]{\textbf{Math dataset size dependence---} We show learning curves and trends in early-stopped loss as a function of dataset size.  For the case of mathematical problem solving, we use a model with $n_{\rm layer} = 64$ and $d_{\rm model} = 512$ for all dataset sizes.  \label{fig:DatasizeScaling}}
\end{figure}

For the math dataset we studied optimal performance as a function of dataset size $D$, in the limit where $N \gg D$ so that performance is constrained by overfitting rather than by model size or compute budget.  For each dataset size and problem distribution, we define $L(D)$ by taking the minimum loss during training (this differs slightly from early stopping, since we may evaluate at different steps if there are several metrics, ie losses on different test distributions, as is the case for math).  For these experiments we used models with $n_{\rm layer} = 64$ and $d_{\rm model} = 512$ for all dataset sizes.  We obtain power-law fits for $L(D)$, as shown in figure \ref{fig:DatasizeScaling}.

\subsection{Additional Math Results}

\begin{figure}
\noindent \centering{} 
\includegraphics[width=0.48\textwidth]{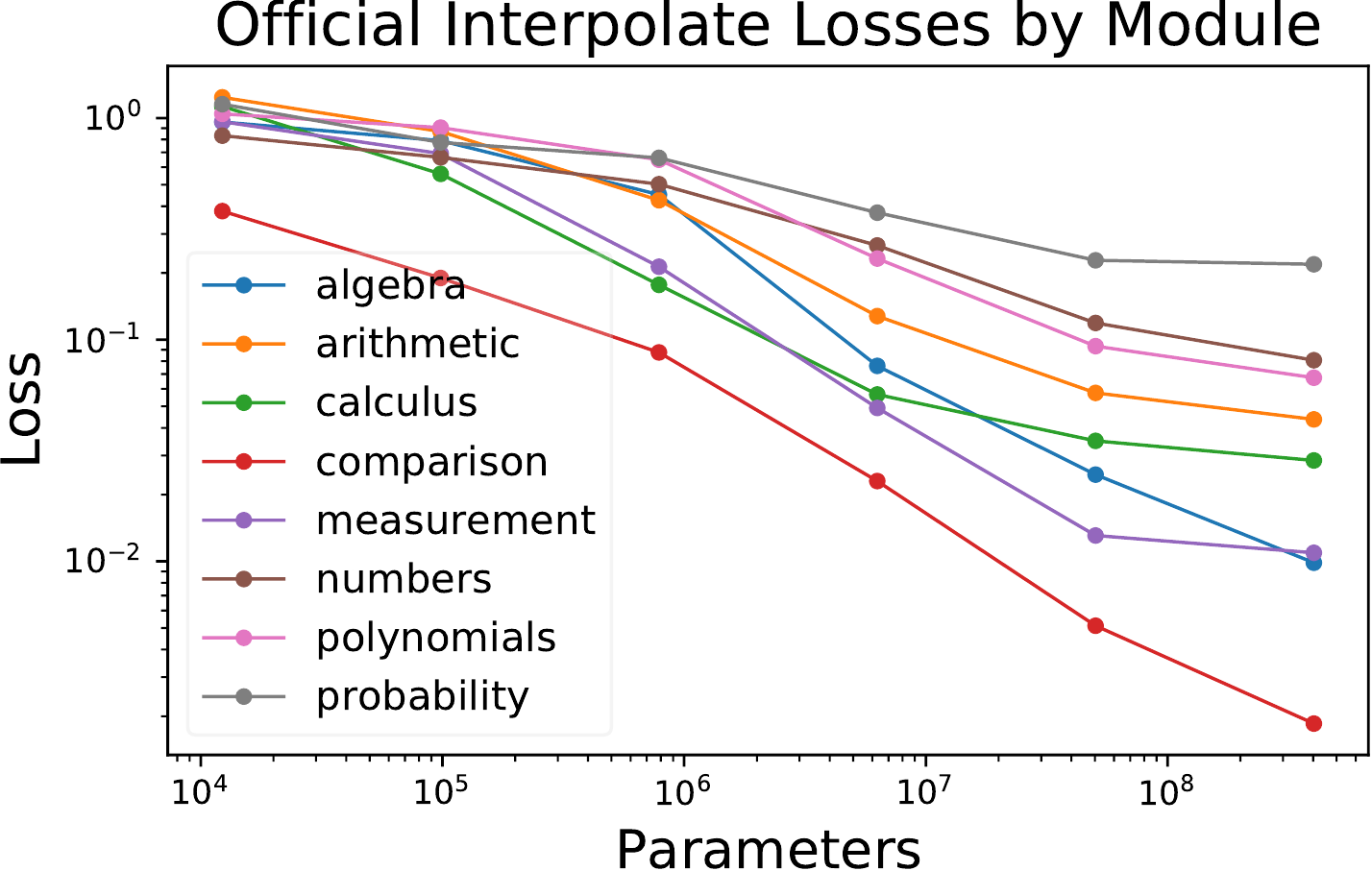}\hfill
\includegraphics[width=0.48\textwidth]{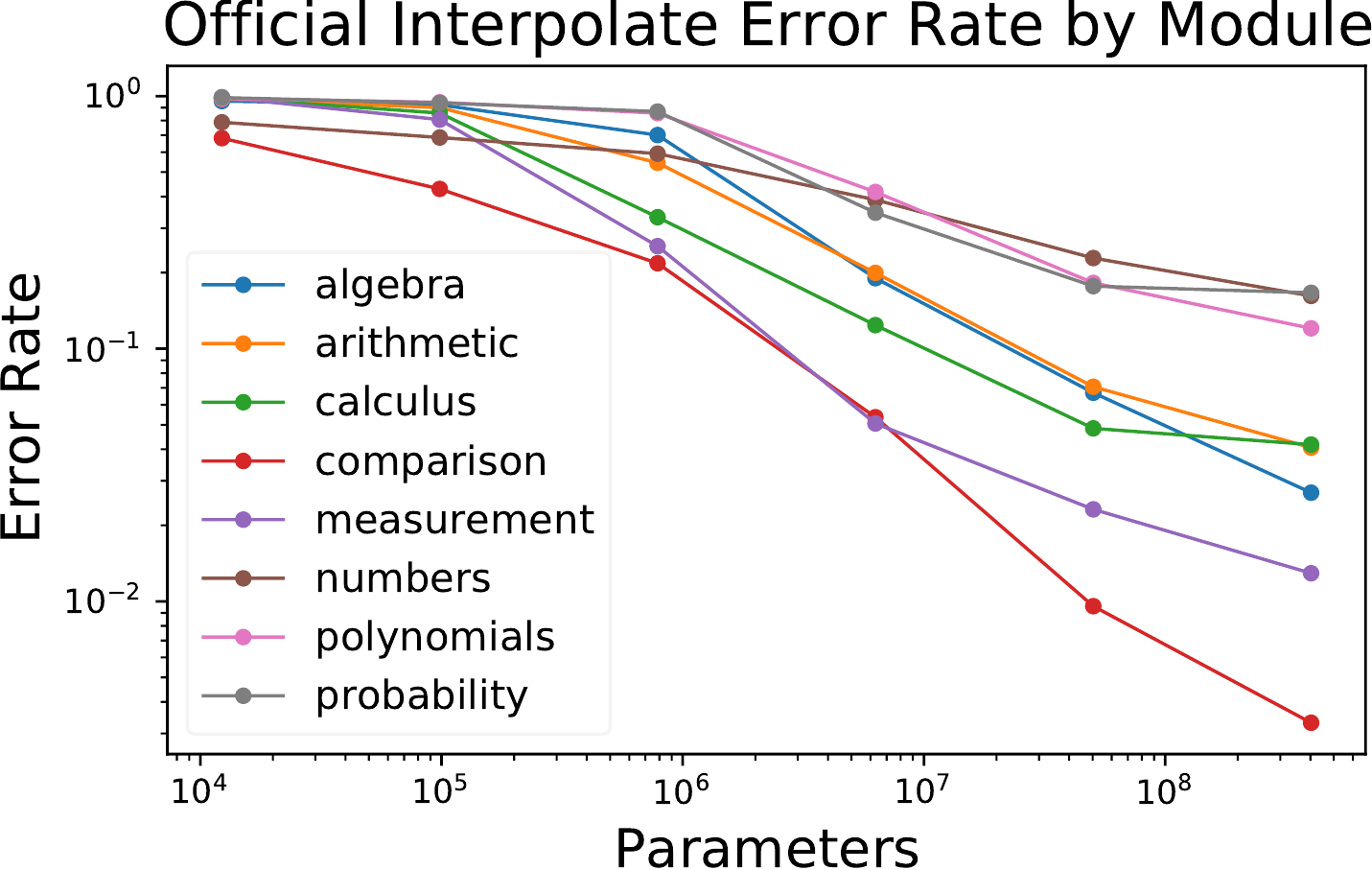}
\\\vspace{1em}
\includegraphics[width=0.48\textwidth]{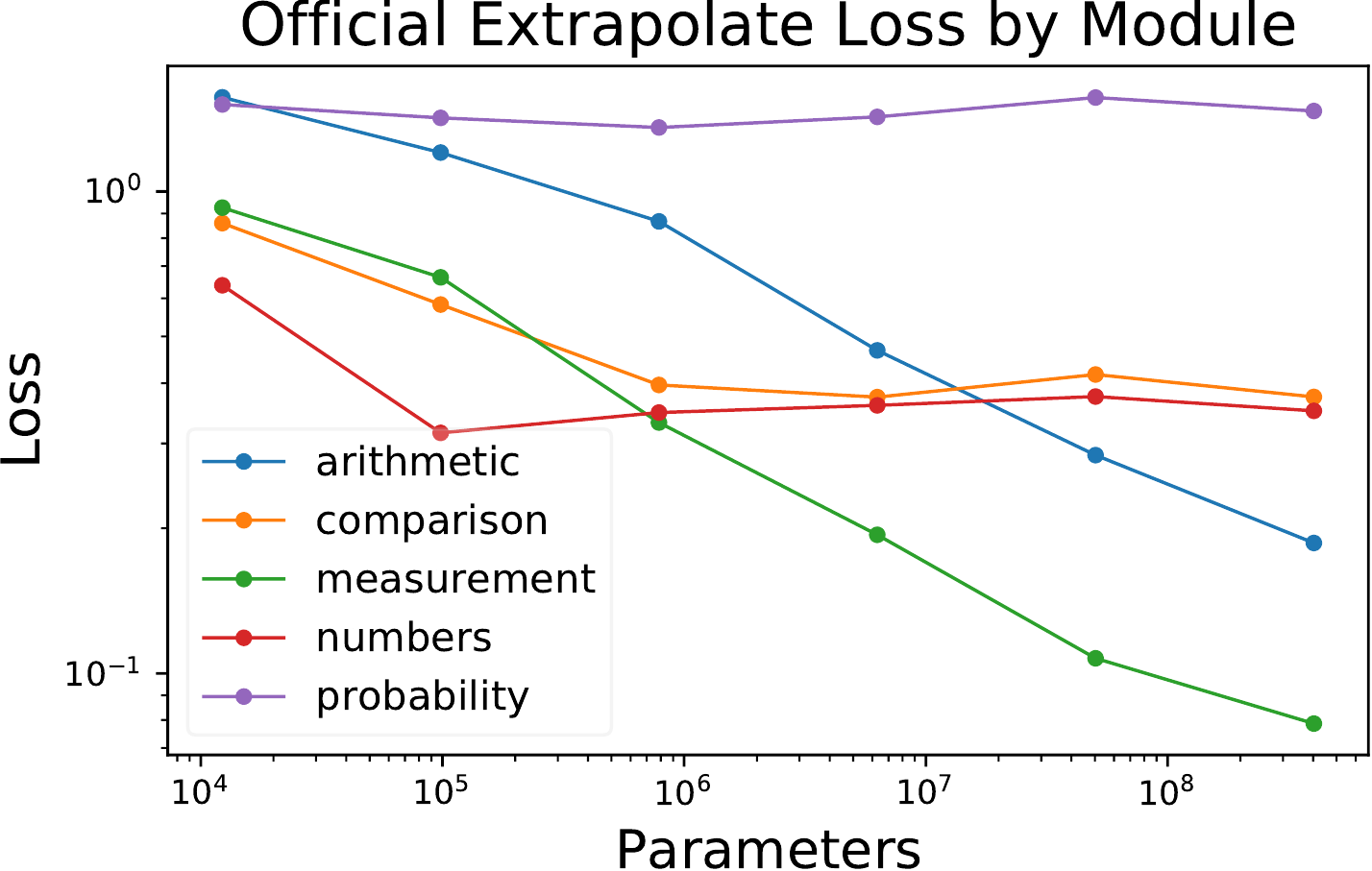}\hfill
\includegraphics[width=0.48\textwidth]{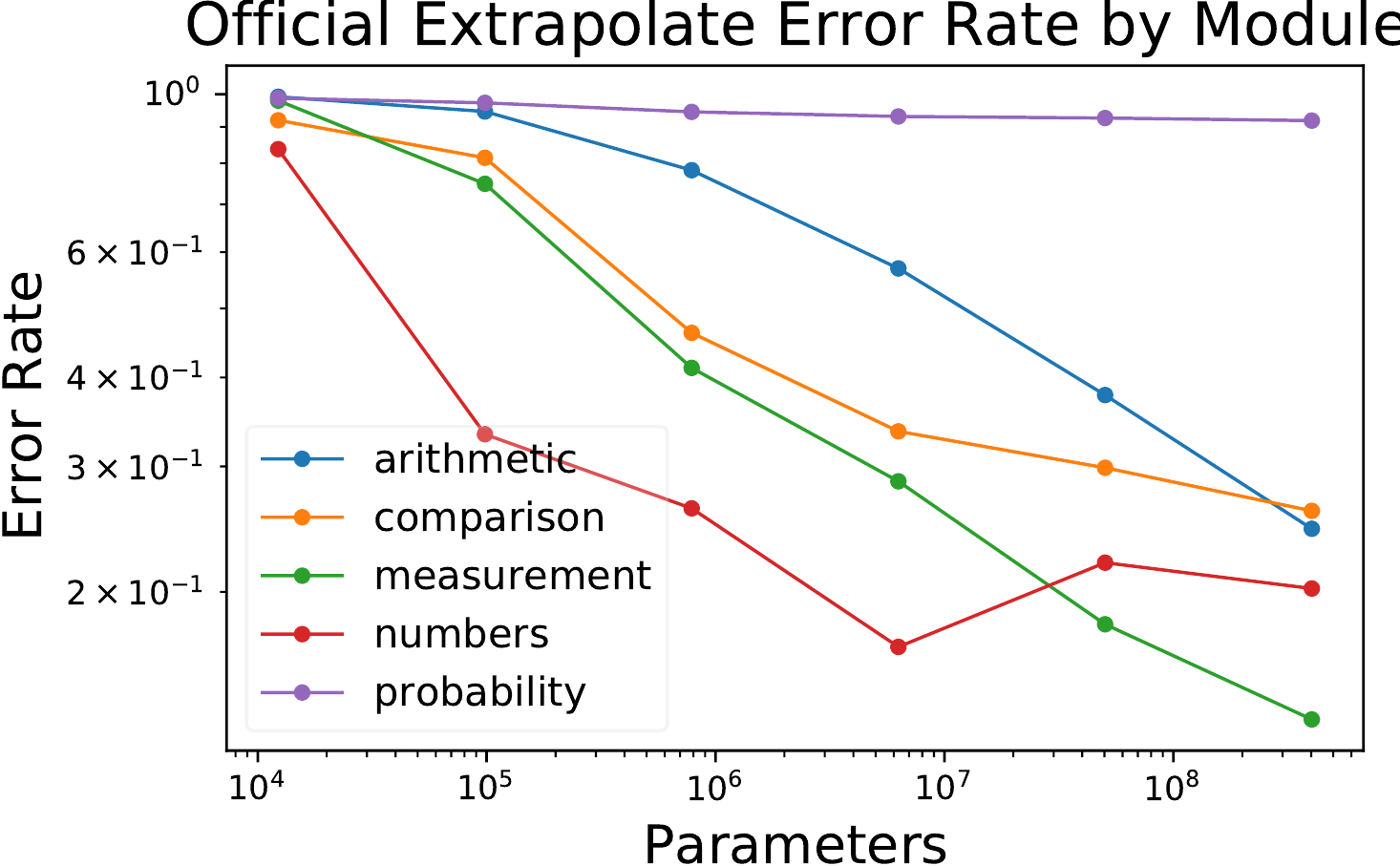}
\caption[Math by Module]{\textbf{Math problem types---} Here we show the performance of the math models on various modules of the math dataset, using the `official' files of problems provided by \cite{DBLP:journals/corr/abs-1904-01557}.  The interpolate problems may have been seen by the models during training, as our training set was procedurally generated.   We note that the losses on individual modules are approximate power-laws with model size on most of the interpolate modules, and on two the extrapolate modules.  \label{fig:MathPerformancebyModule}}
\end{figure}

\begin{figure}
\noindent \centering{} 
\includegraphics[width=0.48\textwidth]{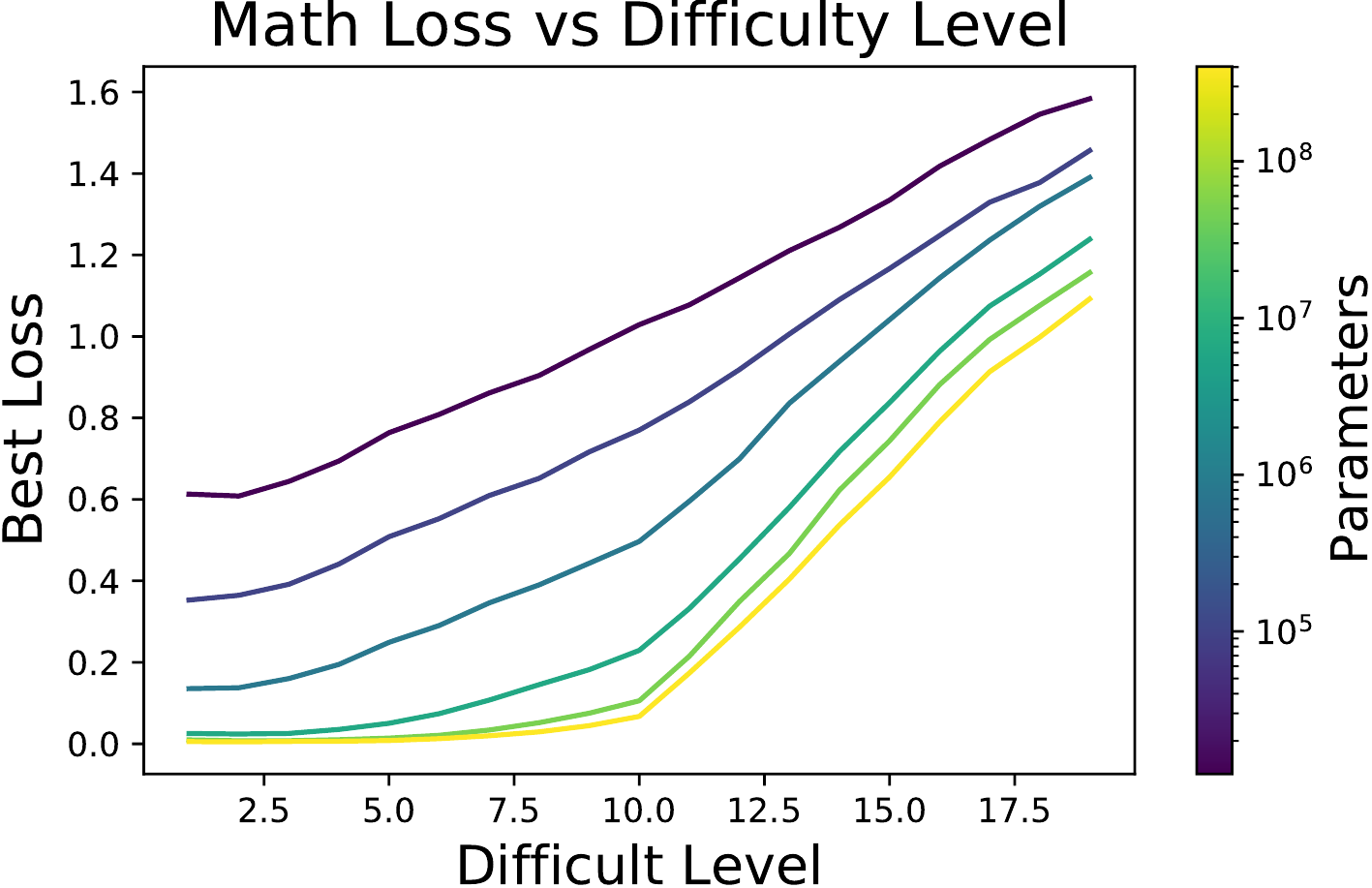}\hfill
\includegraphics[width=0.48\textwidth]{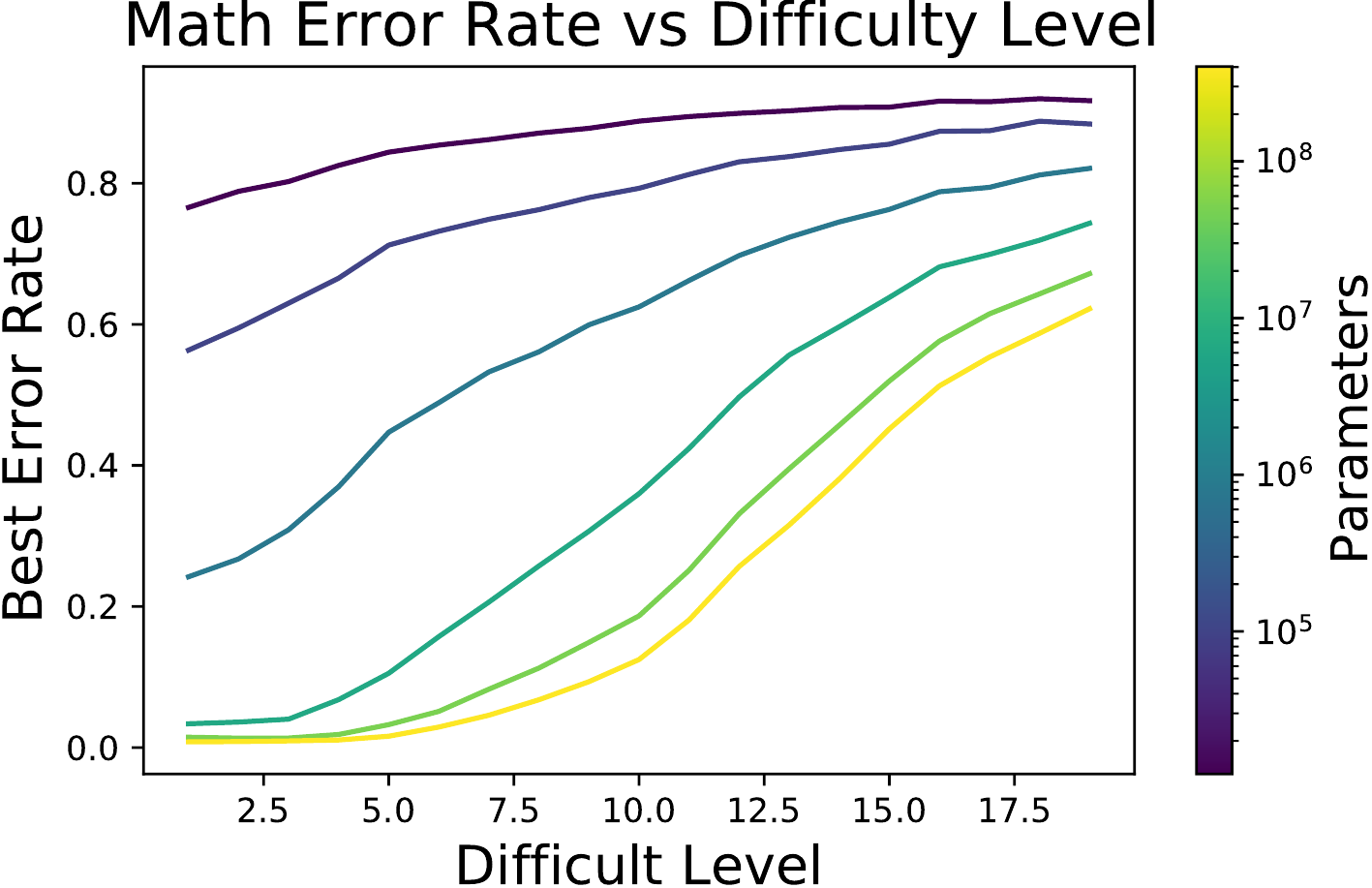}
\caption[Math by Level]{\textbf{Math difficulty levels---}  Here we show how performance of math models varies as a function of the difficulty level or `entropy' of the problem distribution, with levels $\leq 10$  represented in the training distribution.  We note an observable kink at level $10$, suggesting some degree of overfitting, though as we extrapolate to more difficult problems the performance varies smoothly.  It is  clear that larger models perform better.  \label{fig:MathPerformancebyLevel}}
\end{figure}

\begin{figure}
\noindent \centering{} 
\includegraphics[width=0.85\textwidth]{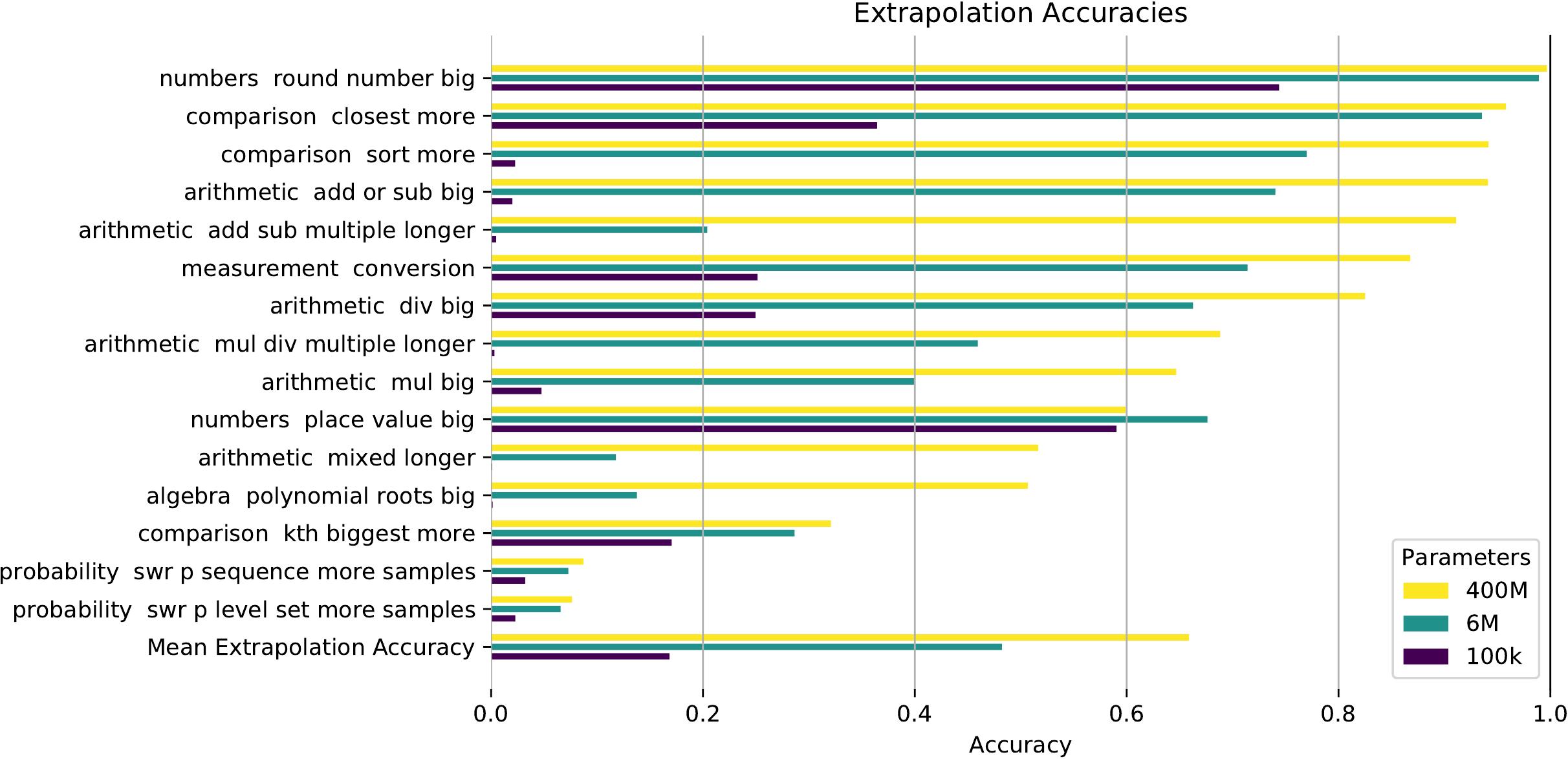}
\caption[Math Extrapolate by Module]{\textbf{Extrapolation results for all math problem types---} Here we show accuracies achieved by  models of three different sizes on the official extrapolation test set files from \cite{DBLP:journals/corr/abs-1904-01557}, grouped by problem generator. Performance almost always improves with model size, though as shown in figure \ref{fig:Extrapolation}, this is due  to the fact that larger models achieve better training loss. \label{fig:MathExtrapolatebyModule}}
\end{figure}

\begin{figure}
\noindent \centering{} 
\includegraphics[width=0.85\textwidth]{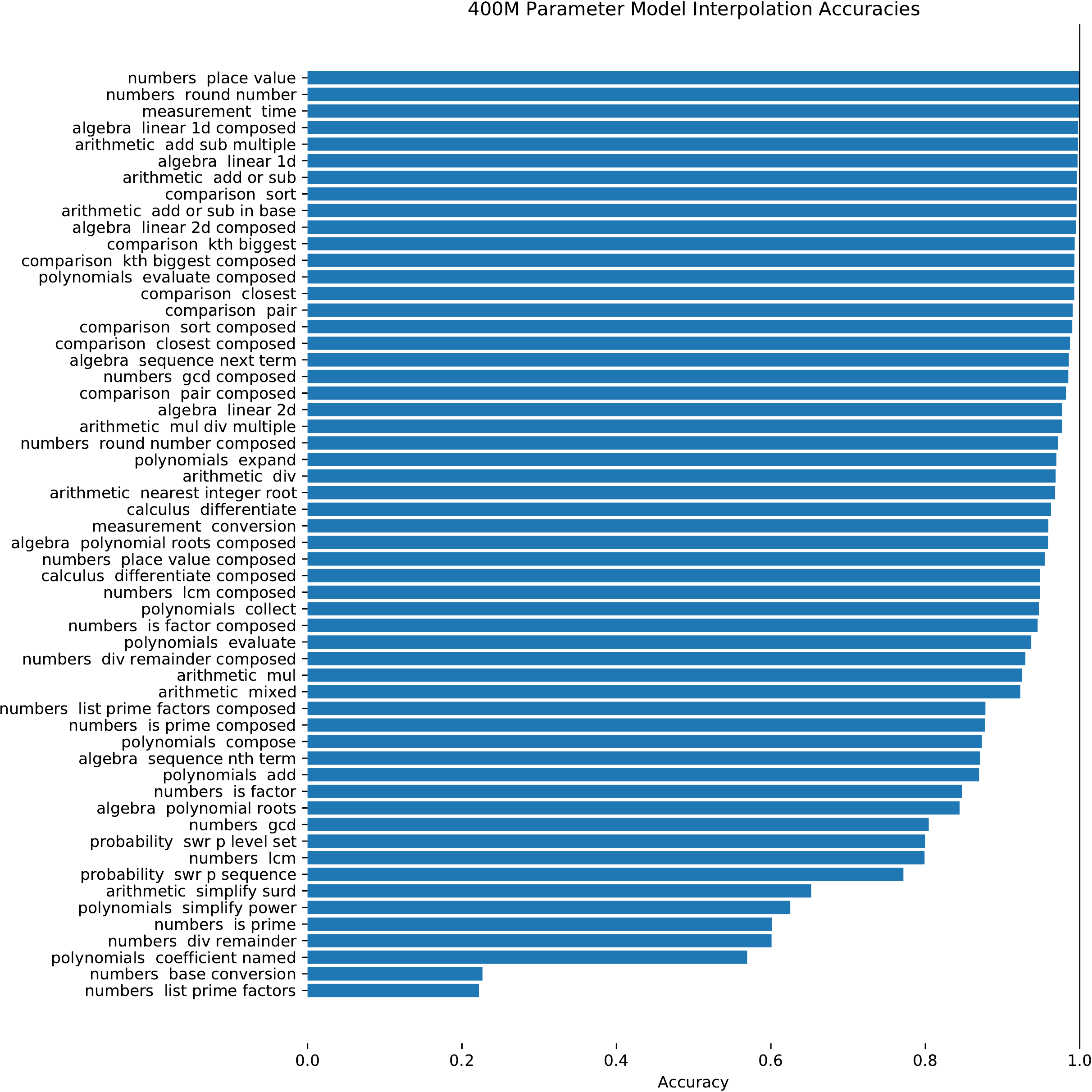}
\caption[Math Interpolate by Module]{\textbf{Interpolation results for all math problem types---} Here we show interpolation accuracies achieved by a 400M parameter model, by problem generator.  Note that these problems (files from \cite{DBLP:journals/corr/abs-1904-01557}) were not deduplicated from our procedurally generated training set, so they may be contaminated by memorization. \label{fig:MathInterpolatebyModule}}
\end{figure}

Here we provide several additional observations about math performance, which can be divided among different math modules and difficulty levels.  In figure \ref{fig:MathPerformancebyModule} we show performance on different modules (using the files provided in \cite{DBLP:journals/corr/abs-1904-01557}), while in figure \ref{fig:MathPerformancebyLevel} we show performance as a function of difficulty level for different model sizes.  We provide details of achieved accuracies on the official extrapolation and interpolation test sets in figures \ref{fig:MathExtrapolatebyModule} and \ref{fig:MathInterpolatebyModule}.

\begin{figure}
\noindent \centering{} 
\includegraphics[width=0.48\textwidth]{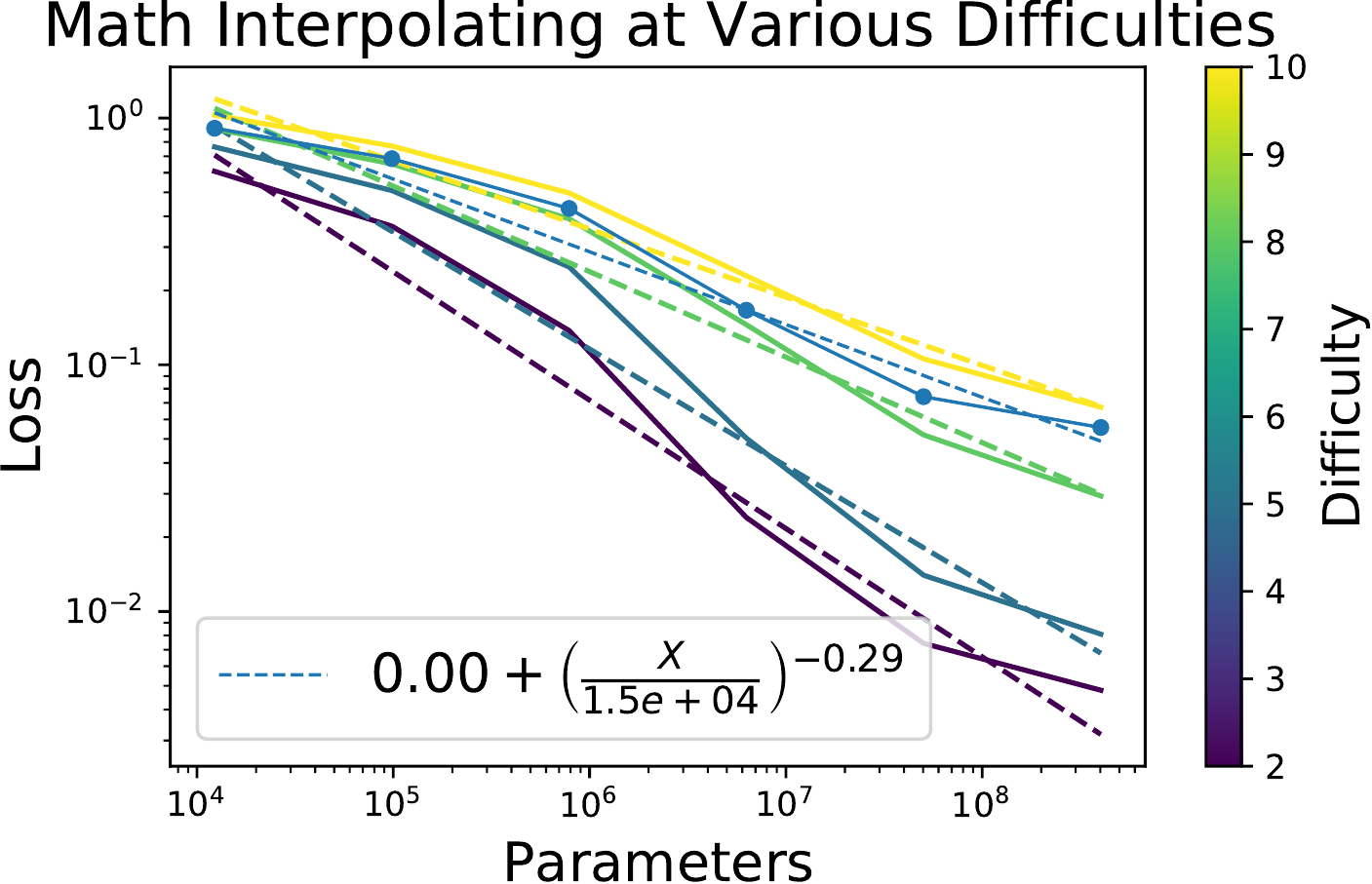}\hfill
\includegraphics[width=0.48\textwidth]{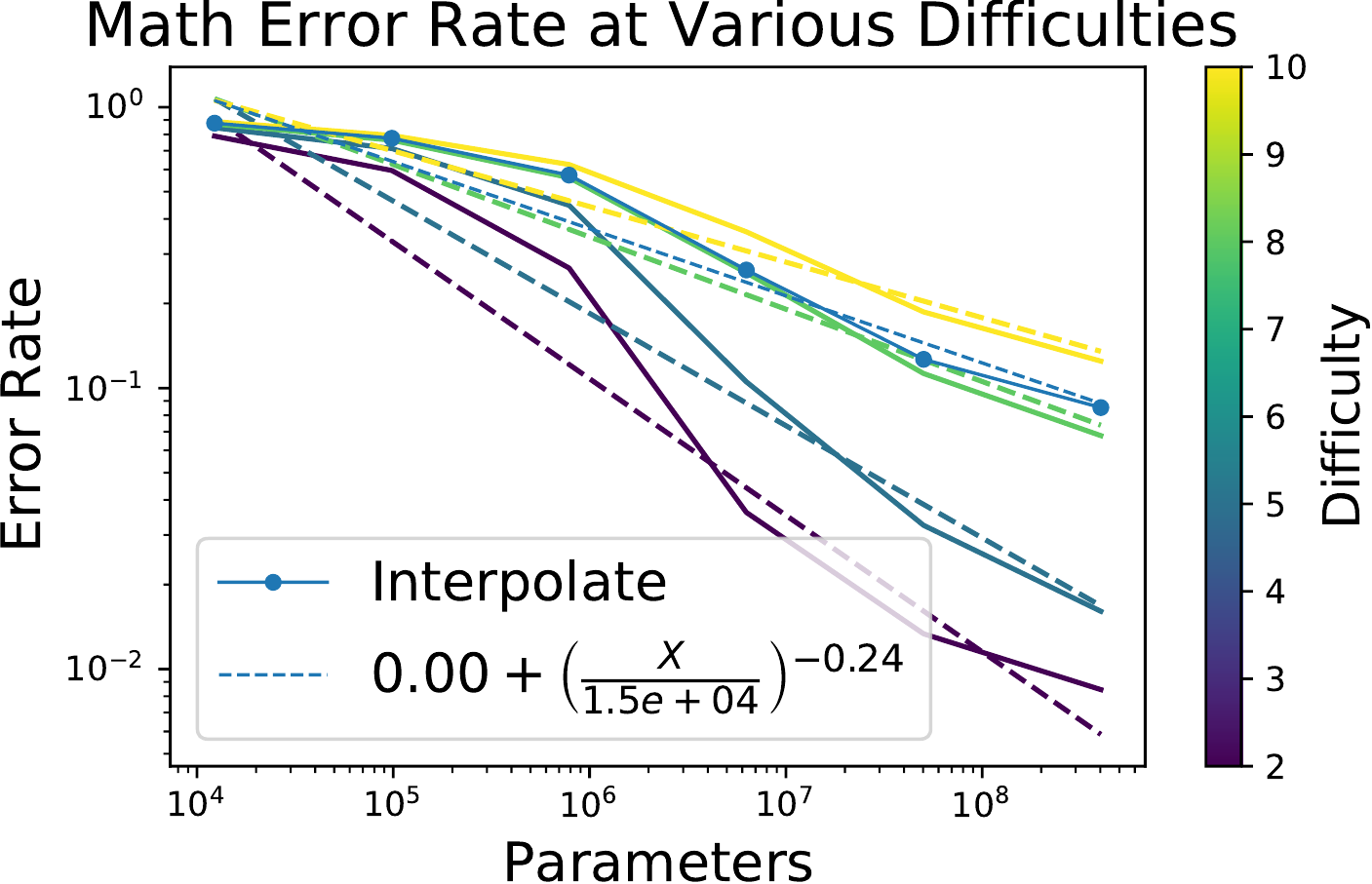}
\caption[Math Interpolate]{\textbf{Model size trends for math difficulty levels---} These plots show trends for the official interpolate dataset, as well as several difficulty levels that are within the training distribution.  We observe that the power-law trends are  distorted, perhaps as a consequence of memorization and the implicit curriculum in the data distribution. \label{fig:MathPerformancebyLevelInterpolate}}
\end{figure}

\clearpage
\section{Additional Multimodal Results}
\label{app:MoreMultimodal}

\begin{figure}
\noindent \centering{} 
\includegraphics[width=0.4\textwidth]{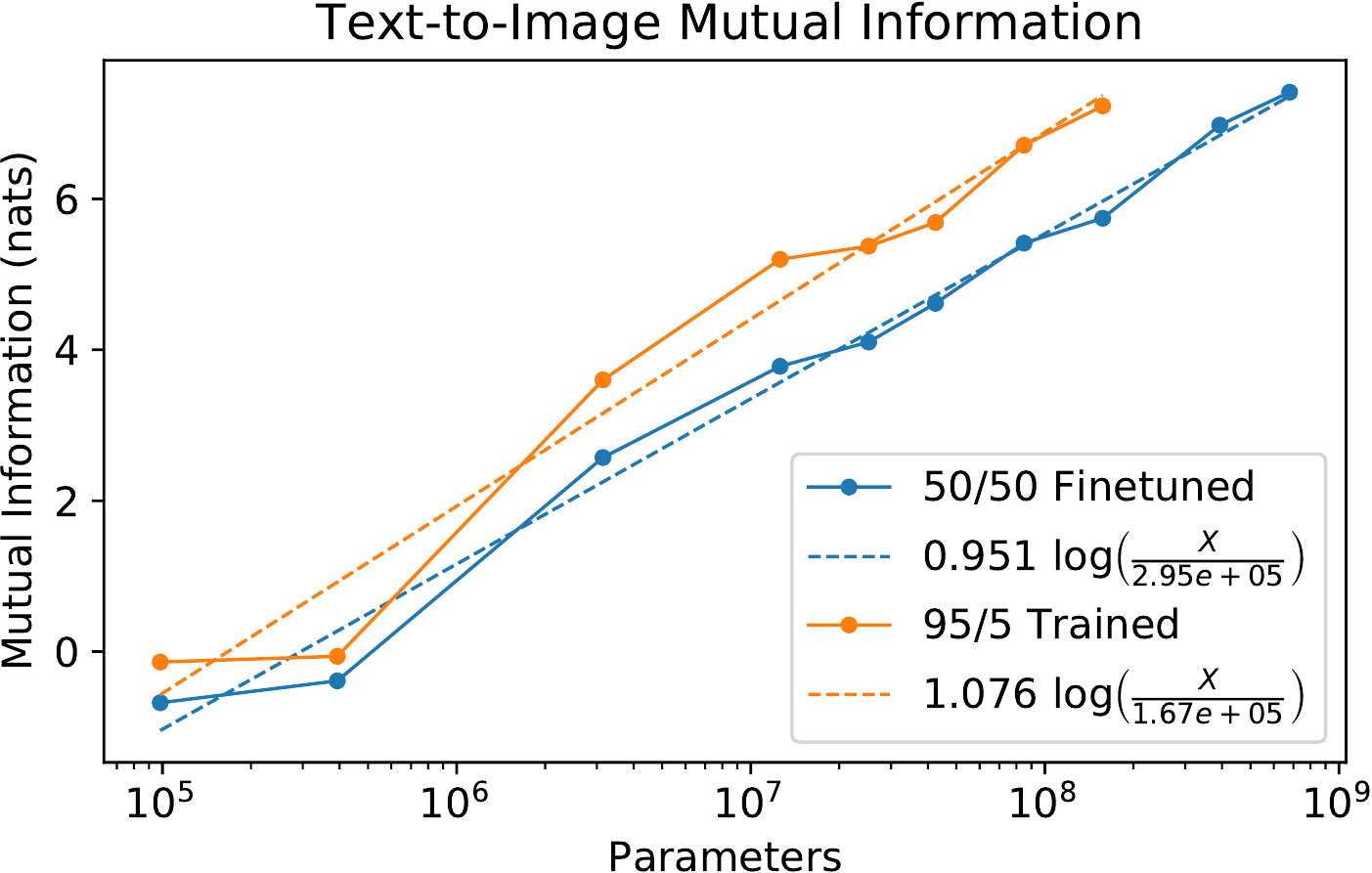}
\includegraphics[width=0.4\textwidth]{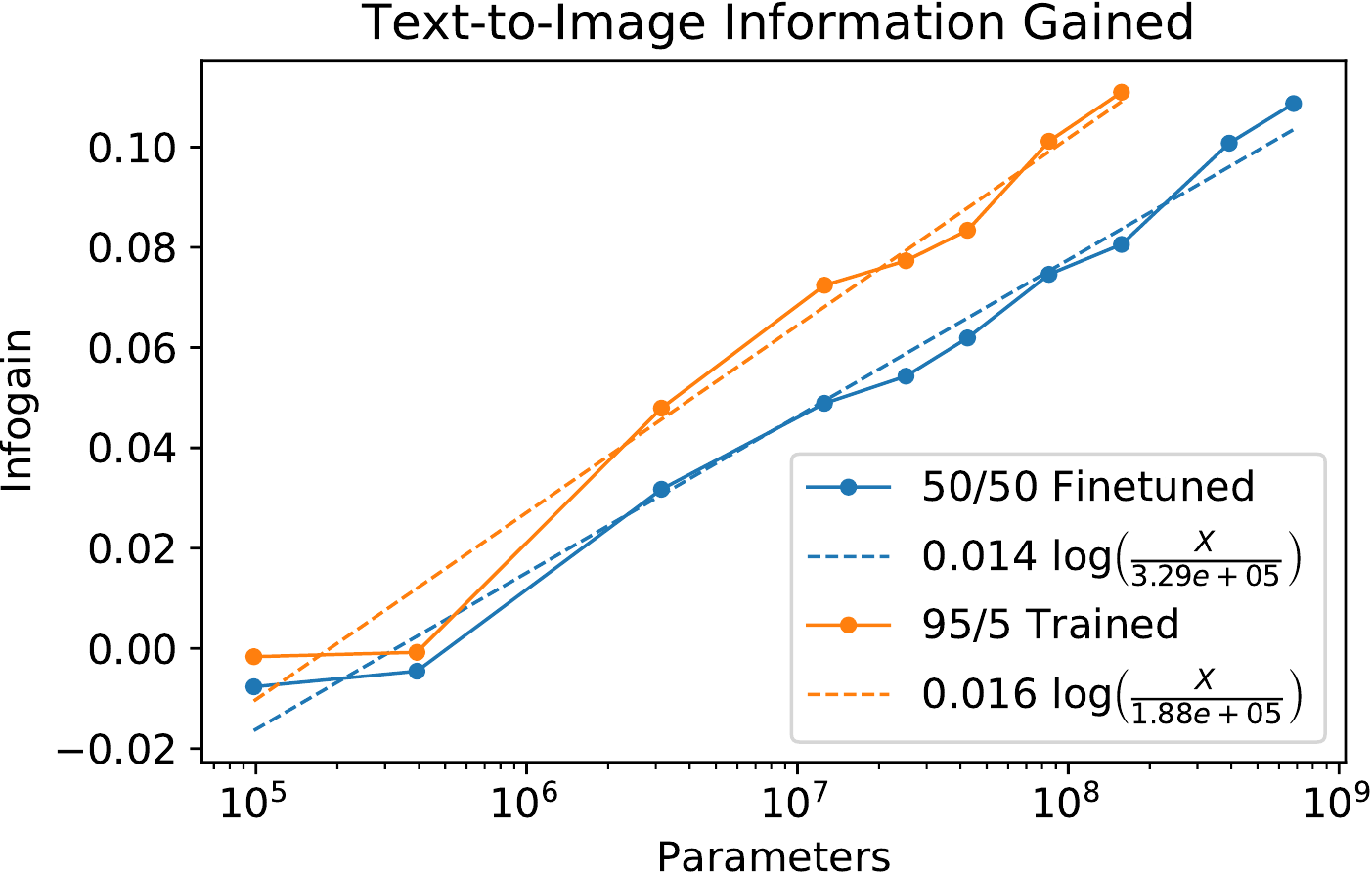}
\caption[InfoGain Trends]{\textbf{Mutual information } In this plot we show the empirical mutual information for text-to-image multimodal models, as well as the Infogain, or the mutual information divided by the empirical entropy of the text.  \label{fig:VerifyingMutualInfoValidity} }
\end{figure}

\begin{figure}
\noindent \centering{} 
\includegraphics[width=0.3\textwidth]{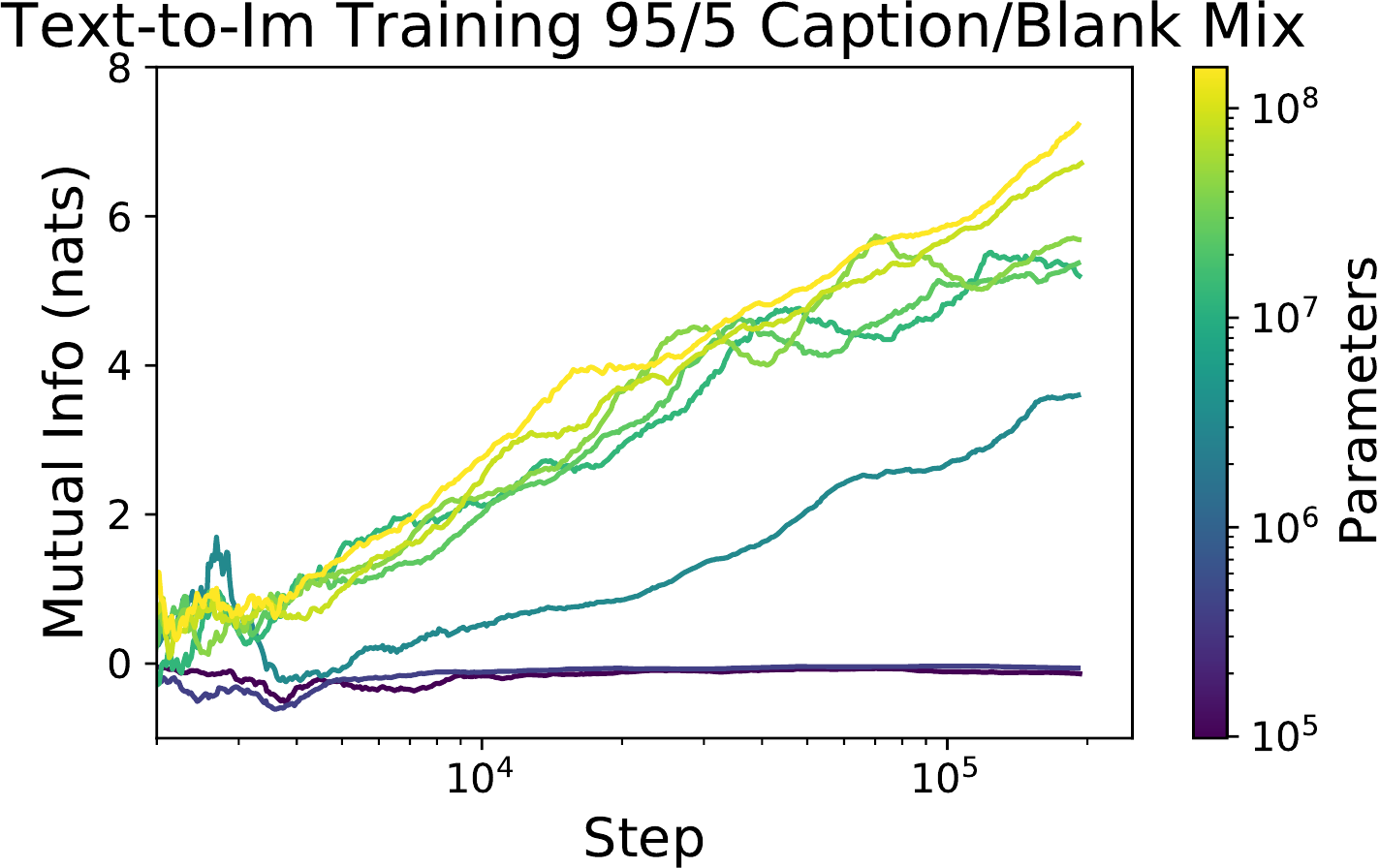}
\includegraphics[width=0.3\textwidth]{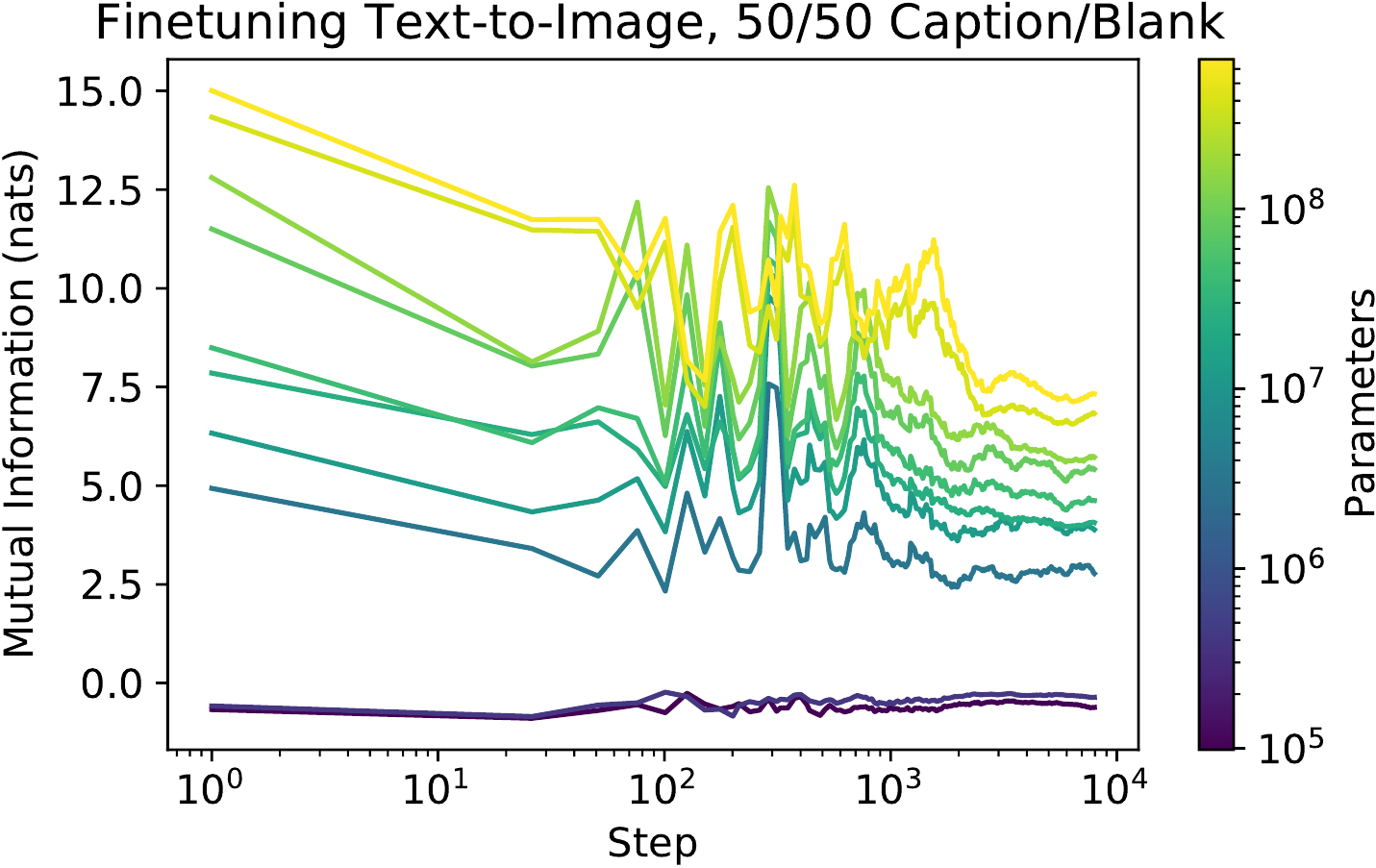}
\includegraphics[width=0.3\textwidth]{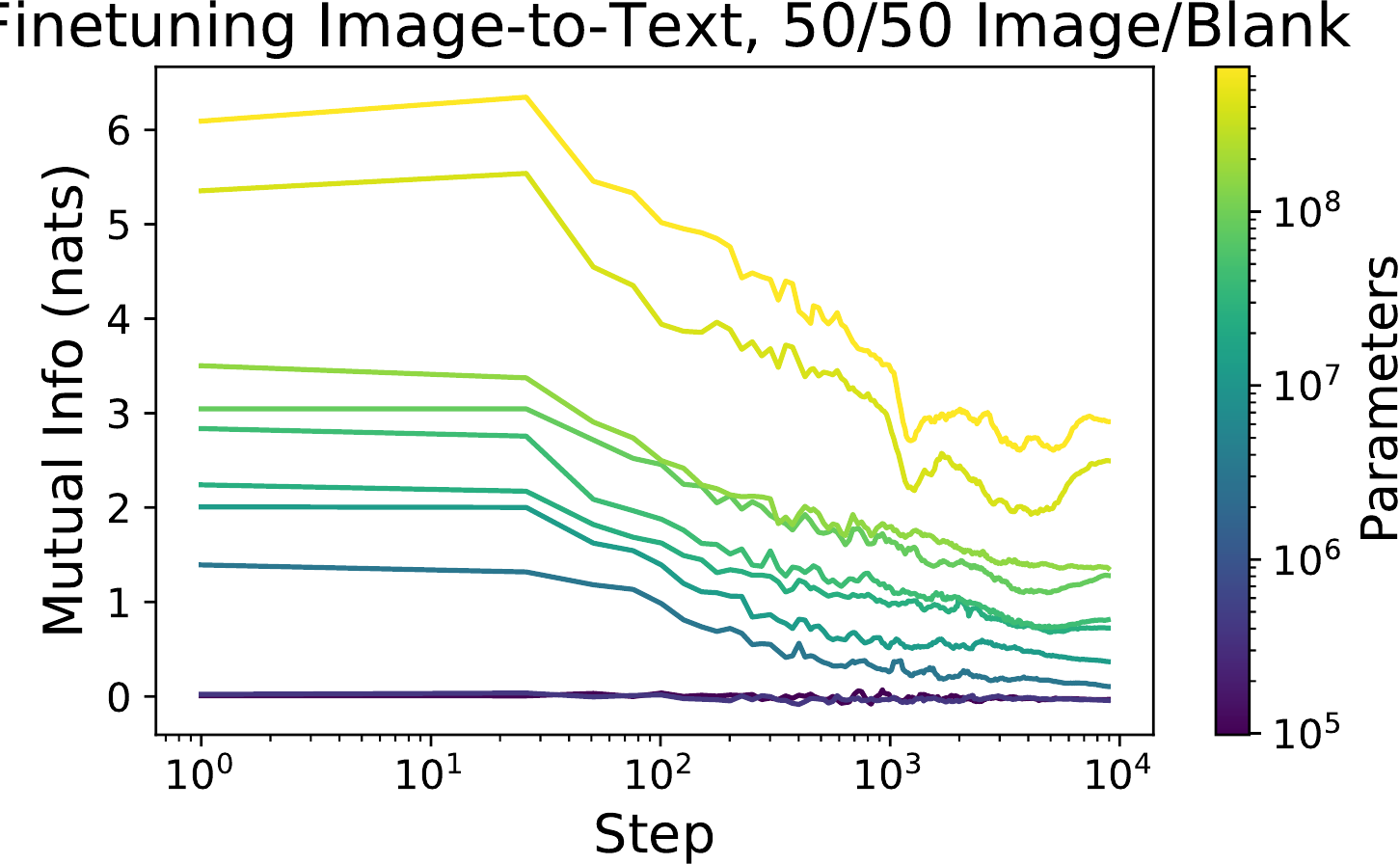}
\caption[Mutual Info Learning Curves]{\textbf{Mutual information learning curves---}  Here we show learning curves for the mutual information when either training or finetuning on a mixture of data with and without captions or images. We include training and finetuning on mixtures in order to ensure that our mutual information and Infogain estimates are not confounded by issues with blank captions or images being out of distribution. \label{fig:MutualInfoLearningCurves}}
\end{figure}

Here we show a few additional results on the multimodal experiments.  The learning curves for the mutual information are shown in figure \ref{fig:MutualInfoLearningCurves}.  This includes both training from scratch on a 95/5 mixture of captioned and blank-caption data for text-to-image, as well as finetuning for 10k steps on a 50/50 mixture for both multimodal directions.  We compare the final mutual information and infogain for the two strategies in figure \ref{fig:VerifyingMutualInfoValidity}; they are very similar.

\clearpage
\section{Additional Language Results}
\label{app:MoreLanguage}

\begin{figure}
\noindent \centering{}  
\includegraphics[width=0.9\textwidth]{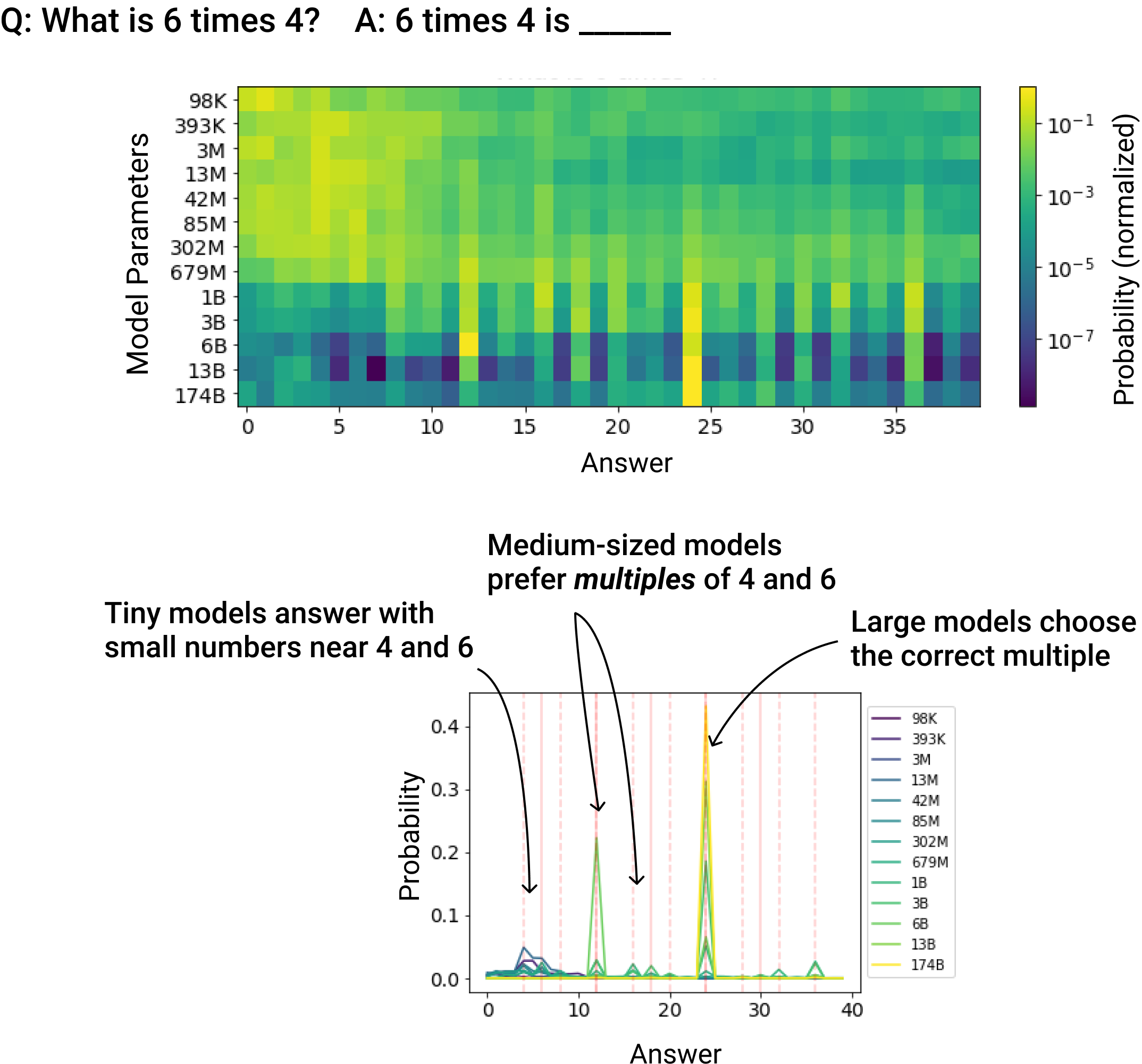}
\caption[Arithmetic Trends]{\textbf{Arithmetic---} We show the progression of arithmetic capabilities of GPT-3 family models as we increase the parameter count  \cite{brown2020language} \label{fig:Arithmetic}. We measure the probability of different numeric answers for a simple multiplication problem.  On the top we show a heat-map of normalized probabilities for each model size, and on the bottom we show a line chart of un-normalized probabilities.

The smallest models put some weight on small numbers near those in the question.  Somewhat larger models start to put some weight on multiples on 4 and 6 (visible as bright vertical streaks on the heat map, and marked as red lines on the line plot), suggesting that they've started to understand the meaning of the multiplication question. The largest models choose the correct answer confidently. }
\end{figure}

\begin{figure}
\noindent \centering{}  
\includegraphics[width=1.0\textwidth]{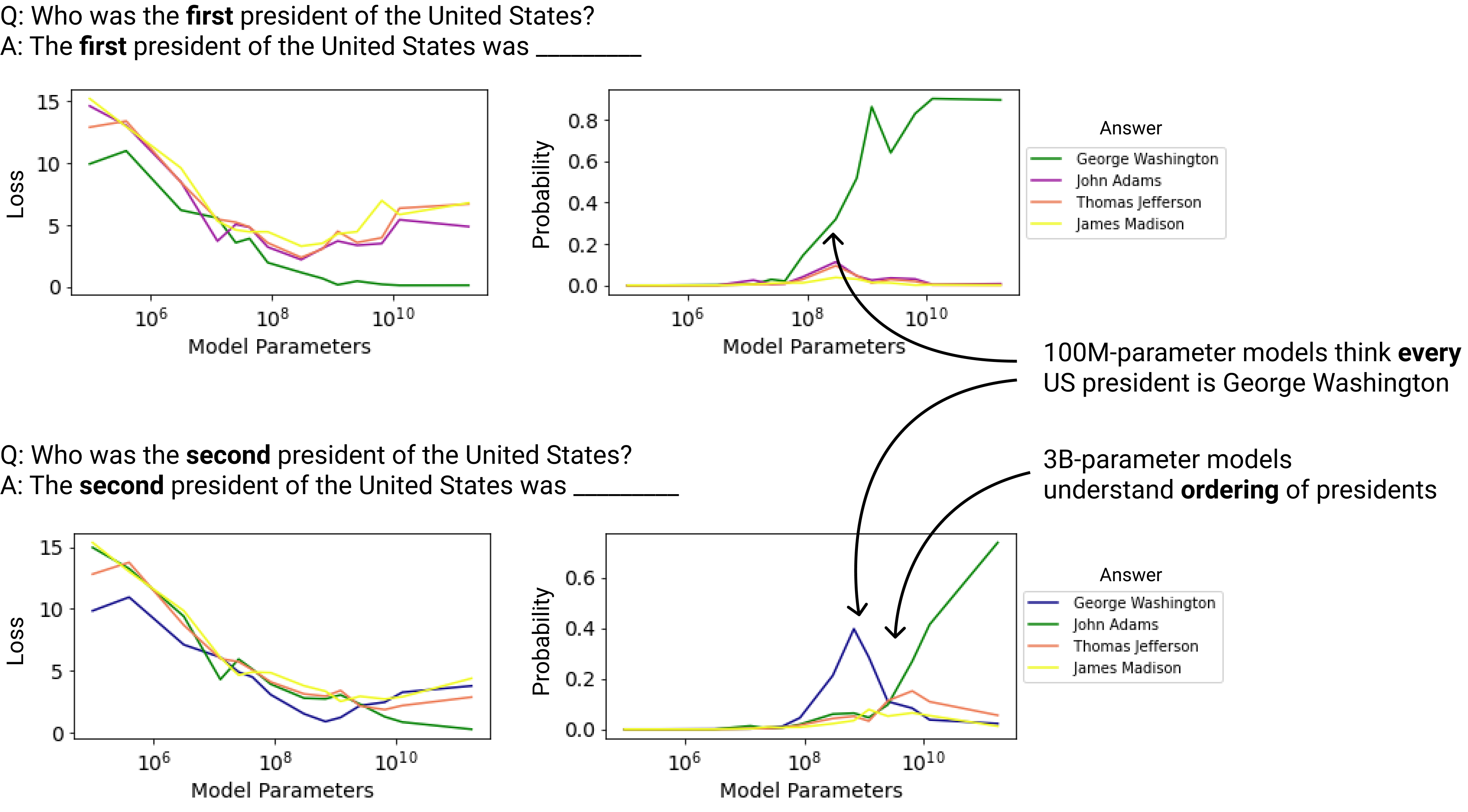}\hfill
\caption[Arithmetic Trends]{\textbf{Q\&A---} We show the progression of simple Q\&A capabilities of GPT-3 family models as we increase the parameter count  \cite{brown2020language}.  \label{fig:Presidents}  We ask the model who the first and second president of the United States was.

Tiny models appear to have trouble understanding the question, and don't  place any significant probability on the correct answer.  Larger models understand that we're requesting a US president, but fail to understand that the ``second president'' and ``first president'' are different requests, placing most of their weight for both questions on ``George Washington''. Only larger models understand both aspects of the questions, answering both correctly.}
`\end{figure}

Here we show a few additional results on the language experiments that measure how performance improves with parameter count.  In figure \ref{fig:Arithmetic}, we investigate the progression of arithmetic capabilities, and in figure \ref{fig:Presidents} we measure the ability to answer a simple factual question.  In both cases we find smooth improvement in the loss on the correct answer as the model size increases. However, we also observe some qualitative ``phases of learning'', with small models having difficulty understanding the question being asked of them, larger models showing some rudimentary understanding, and the largest models correctly answering the questions.

\clearpage
\section{Mutual Information, Infogain, and Scaling}
\label{app:MutualInfoInfoGainScaling}

We are studying the empirical mutual information 
\be
I(X, Y) = \mathbb{E}_{x,y \sim q }  \left[ \log \frac{p(x,y)}{p(x) p(y)} \right] 
\ee 
where $p$ is the model distribution and $q$ is the true distribution of the data.  This must be smaller than the cross-entropy loss of the model 
\be
L(X) = \mathbb{E}_{x \sim q }  \left[ \log \frac{1}{p(x)} \right] 
\ee
on either $X$ or $Y$, so that the empirical InfoGain in equation \ref{eq:InfogainDefinition} cannot be greater than one.   As with the usual mutual information, the empirical mutual information is maximized when $y = f(x)$ or vice versa, so that the relation between $X$ and $Y$ is deterministic, and minimized when $p(x,y) = p(x) p(y)$.

However, it's worth noting an interesting subtlety: in some cases it is possible for \emph{our evaluations} to cause an apparent violation of the bound InfoGain $< 1$.  This can occur in language models that are not precisely translation invariant when $x=$ the first $T$ tokens while $y=$ the following tokens.  For example, it's theoretically possible that a language model with limited computational resources would assign a higher probability to ``The MD5 hash of `powerlaw' is e9f7a4aafeda67a0dab579ba480c24d6'' than to the sequence ``e9f7a4aafeda67a0dab579ba480c24d6'' by itself.

\subsection{Approximate Derivation of Scaling Relations}

We do not know how to derive the relation \ref{eq:ScalingInfo} for multimodal models.  However, we can derive a similar relation for the mutual information and infogain in language models.  In this case, we study  the mutual information between the first $T$ tokens in a text sample, and the next $T$ tokens (it is easy to generalize to sequences of different lengths). 

We know that for a given model size $N$, the loss scales as a power-law with token position $t \geq 1$ \cite{kaplan2020scaling}.  In fact, we can roughly approximate
\be
L(t) \approx L (N) +  \frac{L_U - L(N)}{t^p}
\ee
where $p < 1$ is a power,  $L_U$ is the unigram entropy, and $p$ is roughly independent of $N$.  This model is not perfect, but it permits a straightforward estimate of the empirical mutual information, namely
\be
I([1,T], [T+1, 2T]) & \approx & (L_U - L(N) ) \sum_{t=1}^T \left[\frac{1}{t^p} - \frac{1}{(t+T)^p} \right]  
\nn \\
&=& (2H_T^{(p)} - H_{2T}^{(p)})  (L_U - L(N) )
\ee
where $H_T^{(p)}$ is the $T$th harmonic number with power $p$.
We can evaluate or approximate $H_T^{(p)}$ if desired, but the point is that it's identical for all $N$, and so the $N$-dependence of this expression comes only from $L(N)$.  Because the exponent $\alpha_N \ll 1$ for language models, we can approximate $N^{-\alpha_N} \approx 1 - \alpha_N \log (N)$ to obtain equation \ref{eq:ScalingInfo}. 

Similarly, to approximate the infogain we need to divide by the loss on the final $T$ tokens, so that
\be
{\rm Infogain} \approx   \frac{(2H_T^{(p)} - H_{2T}^{(p)})(L_U - L(N) )  }{T L(N) + (H_{2T}^{(p)} - H_{T}^{(p)}) (L_U - L(N) ) } 
\ee
Expanding this using $L(N) \propto N^{-\alpha_N} \approx 1 - \alpha_N \log (N)$ leads to the approximate formula from section  \ref{sec:Multimodal}.  But more generally we see that the InfoGain is bounded by a certain ratio depending only on $p$ and $T$, since $L(N)$ lies between $0$ and $L_U$.  So it will not actually approach 1.

\subsection{Estimating $D_{\mathrm{KL}}$ Between Real-World Distributions}

\begin{figure}
\noindent \centering{} 
\includegraphics[width=0.32\textwidth]{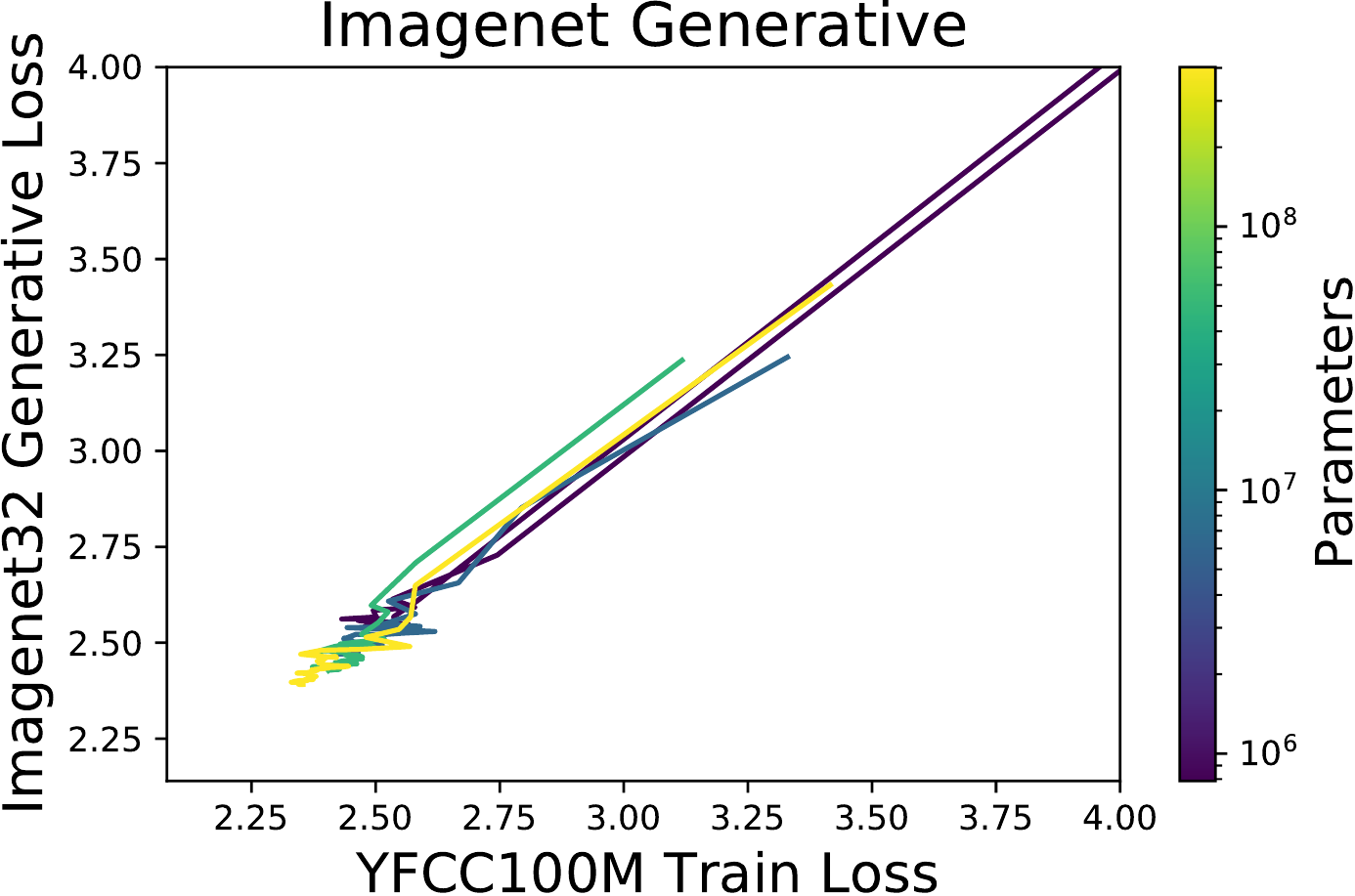}\hfill
\includegraphics[width=0.32\textwidth]{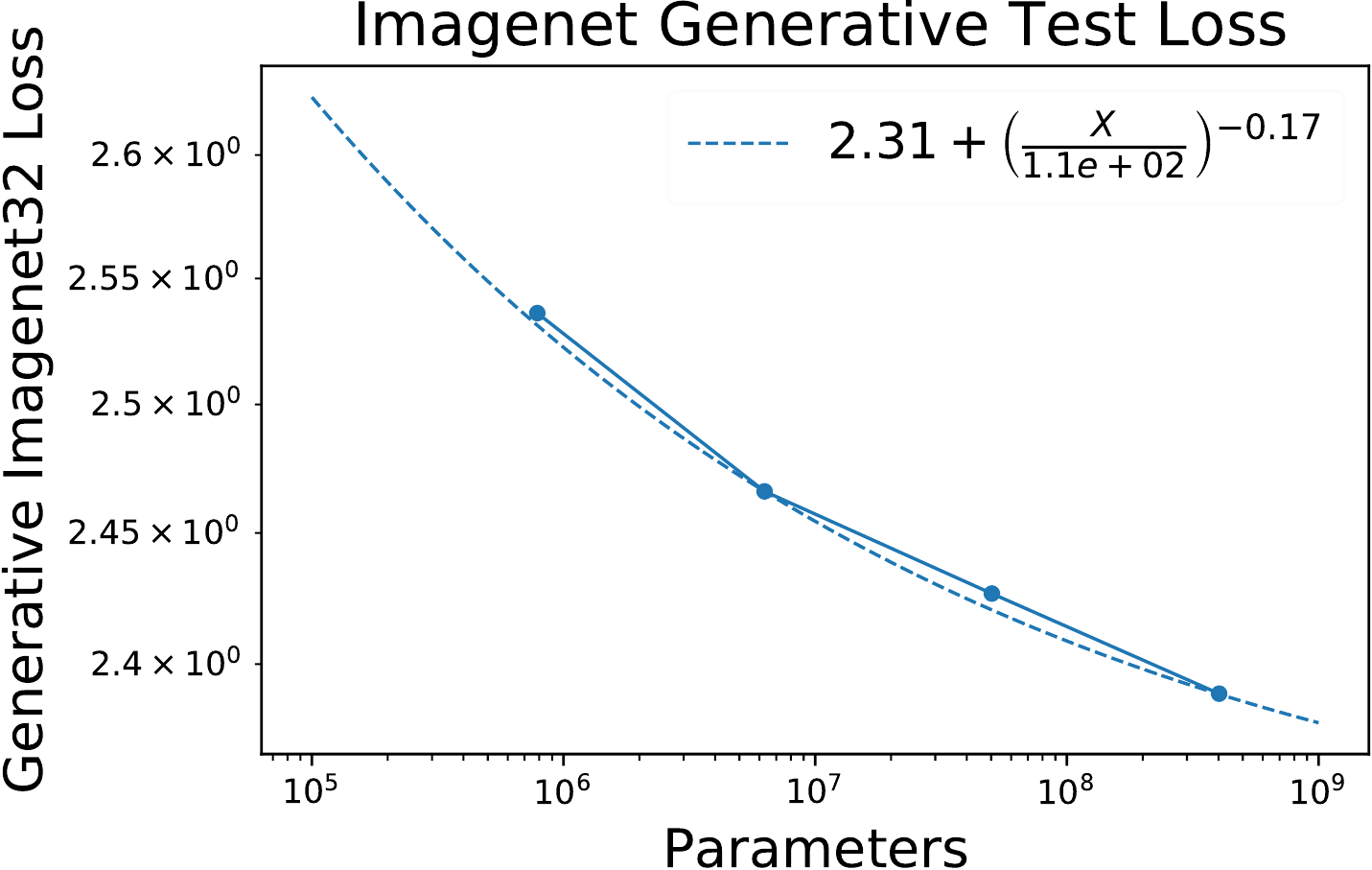}\hfill
\includegraphics[width=0.32\textwidth]{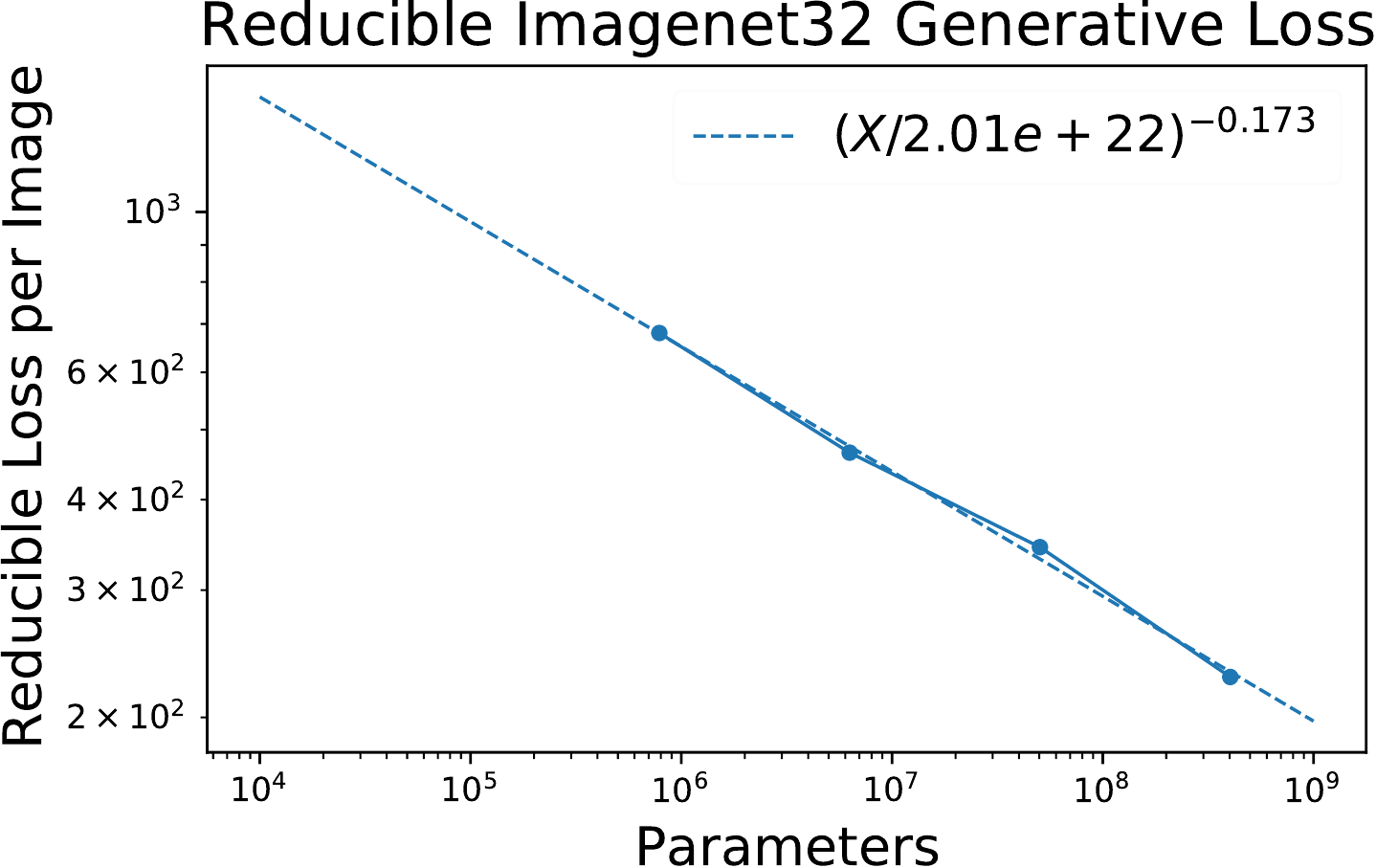}
\caption{\textbf{Generalization from YFC100M to ImageNet generation---} We show information about evaluating YFCC100M-trained models on ImageNet data distribution.  On the left we show that loss on ImageNet depends only on the loss on  YFCC100M, and does not otherwise depend on model size.  In the center we show ImageNet loss vs model size, and on the right we subtract the irreducible loss to compute the reducible loss.  These results strongly suggest that $L(N)$  follows a power-law plus constant form, even  somewhat off of the training distribution.  \label{fig:ImageNetGeneartiveResults}}
\end{figure}

We have interpreted scaling trends in terms of the intrinsic entropy of the data distribution and the KL divergence between the true distribution and our models.  This is based on the idea that with infinite data, model size, and compute we could model the data distribution exactly.  If the empirical loss of our models on a new data distribution also follows a predictable scaling trend, then this means we can estimate the fundamental KL divergence between the new distribution and the training distribution.  Since our models were trained on YFCC100M images \cite{DBLP:journals/corr/ThomeeSFENPBL15}, 
it's interesting to examine the trends for the loss on ImageNet, as we would expect in the infinite limit
\be
L({\rm ImageNet}) = D_{\mathrm{KL}}({\rm ImageNet} || {\rm YFCC100M}) + S({\rm ImageNet})
\ee
where on the left we have the cross-entropy loss on ImageNet for a model trained on YFCC100M.  We show the loss $L(N)$ when evaluating on ImageNet in figure \ref{fig:ImageNetGeneartiveResults}, where we see that it appears to follow a power-law plus constant trend.  Unfortunately this isn't enough to identify $D_{\mathrm{KL}}({\rm ImageNet} || {\rm YFCC100M})$ because we also need a separate estimate of $S({\rm ImageNet})$, but our techniques are not easily applied there due to overfitting.  But this quantity might be extracted by studying dataset size scaling in the future.

\section{Hyperparameter Settings}
\label{app:hyperparameters}

Here we include more details on the hyperparameter settings used to train the models.

All models used a learning rate schedule with a 3000 step linear warm-up followed by a linear decay to $1/10$ of the maximum learning rate.  Model hyperparmeters and learning rates are shown in tables \ref{tab:HPtext2im} and \ref{tab:HPmathimagevideo}. The number of attention heads was always chosen to be $\mathrm{max}(2, d_{\rm model} / 64)$.  Most models were trained with roughly $5 \times 10^5$ tokens per batch; differences from this are noted in the captions of the tables below.  `Parameters' always refers to the non-embedding parameter counts, and is approximate (we do not include biases for simplicity).  

All models were trained for at least 250k steps (parameter updates), but many models were trained for significantly longer, as we noted that they had not yet reached the compute-efficient frontier, or did not seem to have converged.  Trends in the loss as a function of model size were computed at the step minimizing the test loss.  We used very similar learning rates for all models of a given size; these were determined through an initial grid search.

\begin{table}
\centering
\begin{tabular}{rrrrr}
\hline
Parameters &  $d_{\rm model}$ &  $n_{\rm layer}$ &       Max LR & Image-to-Text? \\
\hline
            98304 &       64 &        2 &  0.00164 & \checkmark \\
           393216 &      128 &        2 &  0.00144 & \checkmark \\
          3145728 &      256 &        4 &  0.00115 & \checkmark \\
         12582912 &      512 &        4 & 0.000959 & \checkmark \\
         25165824 &      512 &        8 & 0.000862 & \checkmark \\
         42467328 &      768 &        6 & 0.000789 & \checkmark \\
         84934656 &      768 &       12 & 0.000692 & \checkmark\\
        157286400 &     1280 &        8 & 0.000606 & \checkmark \\
        393216000 &     1280 &       20 & 0.000479 & \\
        679477248 &     1536 &       24 & 0.000402 & \\
\hline
\end{tabular}
\caption{\label{tab:HPtext2im} \textbf{Multimodal hyperparameter settings---}  All Text-to-Image model settings are shown, the Image-to-Text models used identical settings, but the two largest models were not trained. `Parameters' refers to the non-embedding parameter counts, and is approximate (we do not include biases for simplicity). 
 These models were all trained with a batch size of $128$ text/image pairs, or $409600$ tokens per batch.}
\end{table}

\begin{table}
\centering
\begin{tabular}{rrrr}
\hline
 Parameters &   $d_{\rm model}$ &  $n_{\rm layer}$ &        Max LR \\
\hline
     1.23e+04 &       32 &        4 &  0.002686 \\
     9.83e+04 &       64 &        8 &  0.001597 \\
     7.86e+05 &      128 &       16 &  0.000950 \\
     6.29e+06 &      256 &       32 &  0.000565 \\
     5.03e+07 &      512 &       64 &  0.000336 \\
     4.03e+08 &     1024 &      128 &  0.000200 \\
     3.22e+09 &     2048 &      256 &  0.000119 \\
\hline
\end{tabular}
\caption{\label{tab:HPmathimagevideo} \textbf{Math, Image, and Video modeling hyperparameter settings---}  `Parameters' refers to the non-embedding parameter counts, and is approximate (we do not include biases for simplicity). The math models used $n_{\rm ctx} = 512$ and a batch size of 524,288 tokens per batch.  Video models used a batch size of $128$ video clips, for a total of 524,288 tokens per batch. All image models used a batch size of 128 images, so the batch sizes in tokens vary depending on image or VQ resolution. We did not train the largest model size in some domains.}
\end{table}

\newpage
\bibliographystyle{halpha}
\bibliography{bibliography}

\end{document}